Institute of
**Materials and Processes**

Matthias Feiner, Manuel Schöllhorn (Hrsg.)

# KI 4 Industry
## Praxisnah umgesetzt
Der Online Kongress zum Einstieg in KI für KMU

Dokumentation zur Tagung vom
12. & 13. November 2020

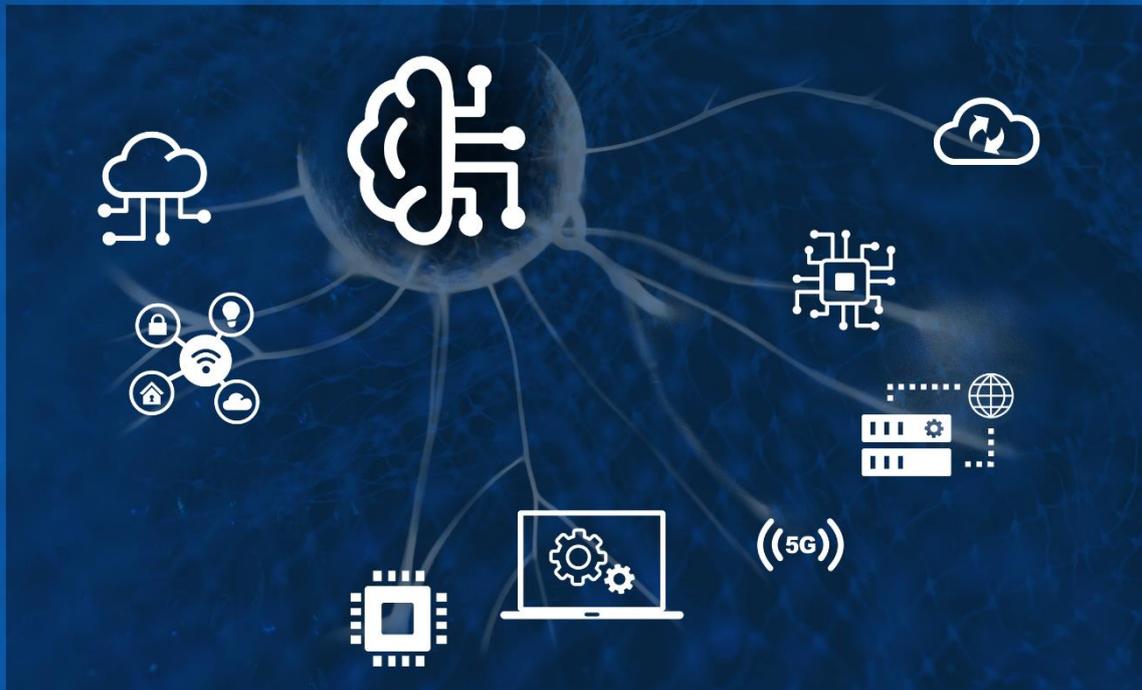

VDI
Karlsruher
Bezirksverein

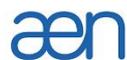

æn
automotive·engineering·network

**Hochschule Karlsruhe**
University of
Applied Sciences

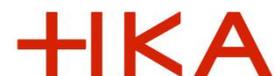

Der Online-Kongress zum praktischen Einstieg in die Künstliche Intelligenz für KMU

# KI4Industry

# KI für den Mittelstand

Matthias Feiner, Manuel Schöllhorn (Hrsg.)

Karlsruhe, 12. – 13. November 2020

Titelillustration von: Thomas Bertram

# KI4Industry
# KI für den Mittelstand

## Conference Chairs

Cordula Goj, Cordula Goj Connecting

Matthias Feiner, M. Sc., Institute of Materials and Processes (IMP) Hochschule Karlsruhe

## Program Committee

Prof. Francisco Javier Fernández García, Universidad de Oviedo

Prof. Dr. Ing. Martin Kipfmüller, Hochschule Karlsruhe

Prof. Dipl. Wirtsch.-Ing. Fritz J. Neff, Vorstand Engineering, AEN Automotive Engineering Network

Prof. Dr. Martin Simon, Verein Deutscher Ingenieure (VDI), Hochschule Karlsruhe

Prof. Dr. Michael C. Wilhelm, OST Ostschweizer Fachhochschule, Schweiz

## Organising Committee

Thomas Bertram M. Sc. Institute of Materials and Processes (IMP) Hochschule Karlsruhe

Jasmin Just, M.A. Hochschule Karlsruhe

Cäcilia Schallwig, M.A. Hochschule Karlsruhe

Prof. Jürgen Walter, Hochschule Karlsruhe

Manuel Schöllhorn, B. Eng., ATW-Ivensys GmbH

## Publication Chair

Lisa Kiefer, B. Sc., Hochschule Karlsruhe

Annette Knödler, BBA, Hochschule Karlsruhe

# Inhaltsverzeichnis

# Vorwort

## KI4Industry – was ist das?

In den letzten beiden Jahren, haben wir in unserer Funktion als Technologietransferdienstleister für die Industrie immer wieder die Beobachtung gemacht, dass viele mittelständische Unternehmen wahrnehmen, dass ein großer Veränderungsprozess begonnen hat: durch immer leistungsfähigere und günstigere Computertechnik sind heute viele Dinge möglich, die so vor einigen Jahren noch nicht umsetzbar waren. Diese „digitale Revolution" wurde zunächst unter dem Namen Industrie 4.0 diskutiert. In der jüngeren Vergangenheit, in der sich der Fokus auf die Analyse von Daten und die Ableitung von sinnvollen Handlungen verschoben hat, wird eher von KI gesprochen.

Bei beiden Themen blieben für den Mittelstand aber immer einige Fragen offen: Was bringt diese Entwicklung für meinen konkreten Arbeitsalltag? Was muss und kann ich tun, um mein Unternehmen für die Zukunft fit zu machen? Hier setzt KI4Industry an: wir möchten eine Plattform schaffen, die es gerade mittelständischen Unternehmen ermöglicht, Antworten auf diese Fragen für ihr Unternehmen zu finden. Startpunkt ist hierbei die erste Tagung KI4Industry, deren Vorträge auf den kommenden Seiten in Form von Artikeln nochmal nachzulesen sind. In der Zukunft werden weitere Veranstaltungen folgen und auch die ersten bilateralen Projekte zwischen Industrie und Hochschule, die aus der Plattform gewachsen sind, sind bereits in den Startlöchern.

Prof. Dr.-Ing Martin Kipfmüller,

Speaker Institute of Materials and Processes

# Mission Control: Mit schlanken und agilen Produktionsansätzen durch die Krise

Per Olof Beckemeier[1], Thomas Meibert[1], Bernd Langer[2], Jan Kotschenreuther[2]

[1]e&Co. Entrepreneurs & Consultants
Berlin, Deutschland
per.olof.beckemeier@eandco.com, +49 151 52 742 518
thomas.meibert@eandco.com, +49 174 3936001

[2]Hochschule Karlsruhe
Fakultät Maschinenbau Mechatronik
Karlsruhe, Deutschland
bernd.langer@h-ka.de, +49 721 925 1904
jan.kotschenreuther@h-ka.de, +49 721 925 1749

**Zusammenfassung.** Die Corona-bedingten Verwerfungen in der Lieferkette zwingen die Unternehmen schneller zu agieren und in kleineren Planungsintervallen zu handeln, basierend auf einer breiten Basis von Entscheidungswissen. Schon aktuell haben Unternehmen mit einem höheren Stand in der Digitalisierung klare Vorteile gegenüber ihren Wettbewerbern mit einem geringeren Stand. Die Digitalisierung der Prozesse ermöglicht durch Daten-Transparenz rascheres Handeln und ein beschleunigtes Heben von Effizienzen.

Datengestützte Produktion und eine intensive Vernetzung der Lieferkette wird einen weiteren Schub erfahren; die Digitalisierung wird ein nachhaltig entscheidender Überlebensfaktor.

Die aktuelle Lage wird über die Phase des Wiederanlaufs hinaus in vielen Betrieben ein zentral gesteuertes, cross-funktionales Krisenmanagement-Team erfordern – mit festen Routinen und klaren Zielsetzungen.

**Schlüsselworte.** Lean, Agile, Digitalisierung, Maschinendatenerfassung

## 1 Einleitung – Lieferkette im Krisenmodus

Die Weltwirtschaft ist durch die Corona Pandemie ins Stocken geraten. Die Lieferketten werden über viele Monate hinaus gestört sein. Die Einschätzungen des ifo Institutes gehen von einem Rückgang der Wirtschaftsleistung in Deutschland für 2020 um 7,2 bis 20,6 Prozent aus [1]. Die große Spreizung der Prognose zeigt die Unsicherheit, die aus der Vielschichtigkeit der Krisensituation resultiert. Die Beratungs- und Investmentfirma e&Co. AG geht von 18 Monaten bis zum Erreichen der Vorkrisenproduktion aus.

Die globale Wirtschaft wird nach der Pandemie anders aussehen als unsere gekannte Normalität. Der gleichzeitig ablaufende Wandel durch die Digitalisierung wird sich als kritischer Wettbewerbsvorteil entpuppen und verstärkt das neue Normal prägen. Welches Szenario sich auch abspielen wird: für die meisten Firmen wird es in den kommenden 18 Monaten primär um das Überleben gehen. Dies wird in Phasen passieren, die unterschiedlichen Handlungsschwerpunkte erfordern.

## 2 Phasen der Krisenbewältigung

Die Krisen-Forschung beschreibt Krisenverläufe in allgemeinen Phasen und stellt gute Vorgehen zur Bearbeitung von Ausnahmesituationen bereit. Einer ersten Phase der Krisenbewältigung folgt die Phase der Neuausrichtung (und Vorsorge). Der Phase hohen reaktiven Handelns folgt eine Verstetigung, die durch bewährte Methoden des Lean Managements gesteuert werden kann (vgl. **Abbildung 1**).

Aktuell befindet sich die deutsche Industrie in der akuten Krisen-Bewältigung. Der Wiederanlauf zielt auf die Vereinbarkeit von Mitarbeiterschutz und Produktion, gefolgt von der nachhaltigen Absicherung der Lieferkette. Zuerst ist die finanzielle Steuerung des Unternehmens auf Liquiditätssicherung ausgerichtet. Sobald die Stabilisierung erreicht ist, wird die Effizienz der Prozesse wieder in den Vordergrund rücken. Dies ist der Einsatzzeitpunkt von Mission Control zur Vorbereitung einer adaptiven Steuerung.

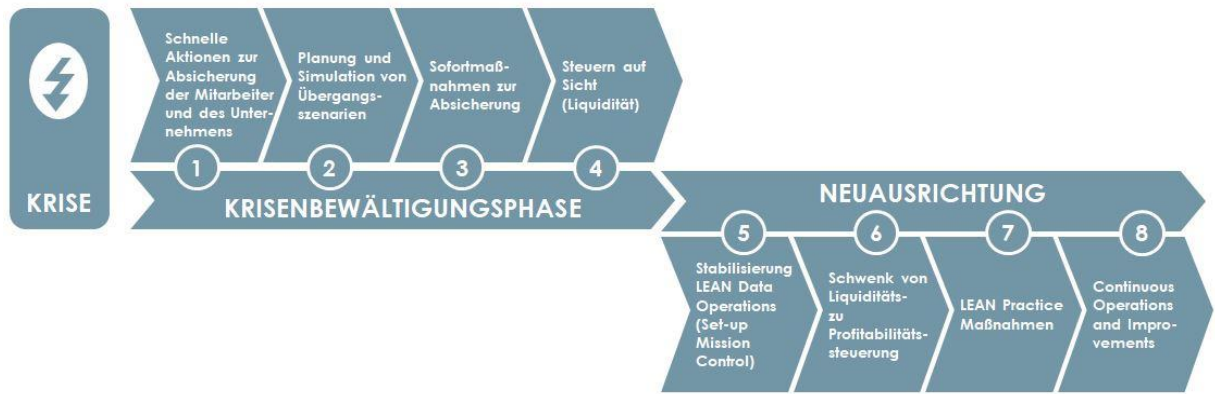

**Abbildung 1:** Von der Krisenbewältigung zur Neuausrichtung

## 3    Digitalisierung als Mega-Trend treibt

Die Digitalisierung als Treiber des Wandels in Industrie und Gesellschaft wird durch die Pandemie beschleunigt – im Bereich der digitalen Tools zum verteilten, kollaborativen Arbeiten hat es bereits einen starken Schub gegeben. Die vier Kernbereiche der Veränderung im digitalen Wandel können als CODE [2] beschrieben werden:

• **C**-reating Digital User Value – Wie aus Produkten digitale Erfahrungen werden

• **O**-perieren in Lernzyklen – Wie aus linearen Prozessen agile Zyklen werden

• **D**-esign bimodaler Organisation – Wie Skalierung und Innovation versöhnt werden

• **E**-volution der Führung – Wie postheroische Führung gelingen kann

Firmen, die bereits Komponenten der Smart Factory umgesetzt haben – insbesondere ein digitales Datenmanagement – können schon jetzt wesentlich besser mit der Corona-Situation umgehen. Das ist ein Wettbewerbsvorteil.

e&Co. und die Hochschule Karlsruhe arbeiten gemeinsam mit dem Softwarehersteller Freedom IoT zusammen. Im Rahmen dieser Zusammenarbeit wurde die Software „Freedom" zur Erfassung der Maschinendaten im Produktionsnetzwerk im Labor des Institute of Materials and Processes (IMP) installiert. Die erfassten Maschinendaten, die durch manuelle Eingaben ergänzt werden können, sind der unterste Baustein der Digitalisierungspyramide. Diese Daten lassen sich bereits auf diesem Niveau zur Kennzahl OEE (Overall Equipment Effectiveness / Gesamtanlageneffektivität) Daten aggregieren. Diese deckt Verluste in den Bereichen technische Verfügbarkeit, Leistung und Qualität auf. Dadurch werden faktenbasierte Entscheidungen vom Shopfloor bis zum Top-Floor möglich.

Darüber hinaus ermöglicht die Software die Erkennung von Ausfallgründen über die von der Maschinensteuerung generierten Alarmmeldungen, welche von Metadaten aus dem Produktionsumfeld ergänzt werden. So kann gezielt eine Priorisierung der eingesetzten Ressourcen vorgenommen werden um Verschwendungen in der Produktion zielgerichtet und nachhaltig zu reduzieren.

Ein weiterer Vorteil von Freedom ist die Benachrichtigung von Wartungspersonal bei Auftreten von bestimmten Fehlern. So kann die MTTR (Mean Time To Repair) wesentlich gesenkt und daraus resultierend die technische Verfügbarkeit signifikant gesteigert werden.

Digital Twins der Produktionsstätten (virtuelle Abbildungen) helfen in der Bewältigungsphase bei der schnellen Erstellung von Social-Distancing-Szenarien. In Kombination mit Maschinendaten können virtuelle GEMBA-Walks durchgeführt werden und die oft reiseaufwändigen Wartungsaktivitäten aus der Ferne erfolgen.

e&Co. empfiehlt für die Lieferketten und die Operations in der aktuellen Situation das Augenmerk auf das Operieren in Lernzyklen und die Evolution der Führung[2] zu legen. Dies führt zu Agilität und dann weiter zur Wandlungsfähigkeit. Die Führung in der Krisensituation muss diesen Prozess mit klaren Vorgaben anleiten und unterstützen.

## 4    Mit Agilität Wandlungsfähigkeit erreichen

Wandlungsfähigkeit beurteilt die Fähigkeit aller Assets eines Unternehmens (Menschen, Prozesse, Produkte) auf interne oder externe Veränderungsanstöße zu reagieren. Sie kann als Kombination von technischer Flexibilität und Agilität einer Unternehmung beschrieben werden.

Technischer Flexibilität meint die in Operations angelegte Variabilität des Leistungsspektrums. Agilität bezieht sich auf die Fähigkeit und vor allem Entscheidungstüchtigkeit, das Operations-Modell kostenoptimal an interne und externe Umfeld-Chancen anzupassen.

Die Flexibilisierung der Anlagen ist in der Regel mit Investitionskosten verbunden und kurzfristig nicht umsetzbar. Daher ist der Blick auf die Agilität zu richten, um eine größere Wandlungsfähigkeit in unsteten Lieferketten zu erreichen. Es verlangt einen agilen Umgang mit der Situation – dabei hilft eine agile Unternehmenskultur in der Gemeinschaft aller Mitarbeiter, Selbstverantwortung zu übernehmen, Probleme zu erkennen / zu lösen, Erfolge festzuhalten, die Fokussierung auf Aufgabenstellungen gemeinsam zu führen, Selbstmanagement zu leisten und Netzwerke aufzubauen.

## 5    Mission Control

Die Aufgabe des Managements ist es in den derzeitigen turbulenten Zeiten die Agilität zu erhöhen. Die Empfehlung an dieser Stelle lautet: Aufbau einer "Unternehmensentwicklungsfunktion" unter dem Begriff "Mission Control". Sie koordiniert den Übergang aus der Krisenbewältigungsphase in eine nachhaltige Neuausrichtung und stärkt die aktive Positionierung im Wettbewerb. Mission Control vereint ein tiefes Verständnis des Industry Shop Floors mit seinem erforderlichen Drill und einer smarten Digitalisierung der Operations, was erst eine wirkliche Operational Agility ermöglicht.

Um Reaktivität zu erhöhen muss die Koordination durch Planung und Hierarchiestrukturen der Koordination durch schnelle Kommunikation und Kollaboration weichen (vgl. **Abbildung 2**).

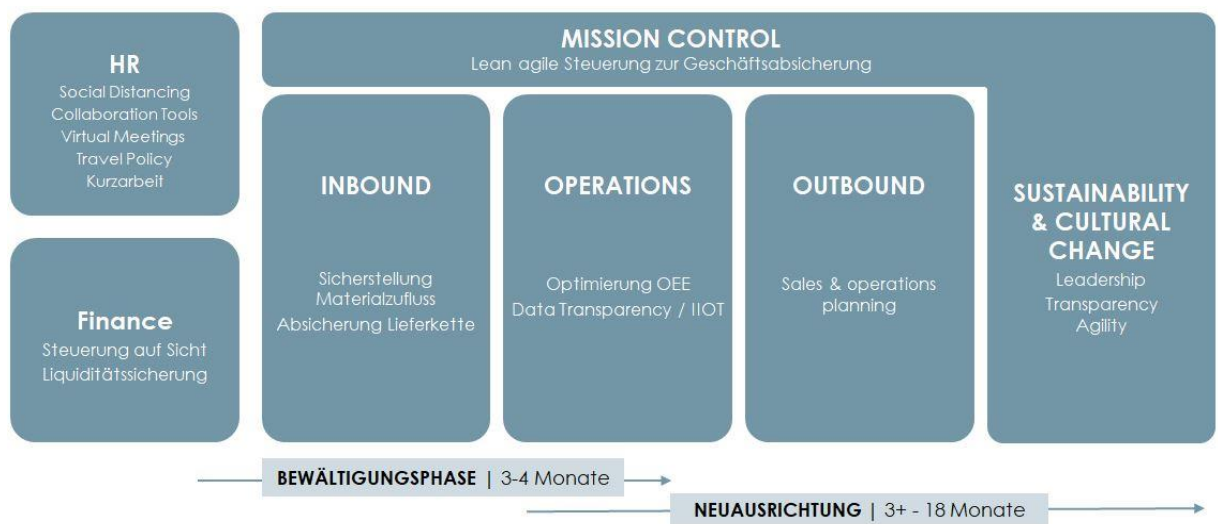

**Abbildung 2:** Strukturumbau von der Bewältigungsphase zur Neuausrichtung

Die klassischen Steuerungsprozesse koordinieren zu sehr auf einer Makroebene, als dass sich dieses Instrumentarium im Krisenmoment eignen würde. Zur Verfeinerung ist die Weiterentwicklung der Organisation durch den Einsatz agiler Prinzipien sinnvoll und somit ist die Mission Control (als Klammer der Veränderung) eine Chance für den Einstieg in den nachhaltigen Wandel. Dieser Einstieg wird durch eine Abwandlung eines Demming Circles (Plan-Do-Check-Act) als Kollaborationsprozess und durch Cross-Funktionale Teams flankiert.

Die Effektivität der Lean Practices hängt an der Transparenz und schnellen Datensammlung in der Operations und der Supply Chain. Ein IIoT-basiertes Lean Data Operations muss an die Lean Practices gekoppelt werden, es ermöglicht eine Zukunftsvoraussage und akkurate Planung, um Zeit, Geld und Assets zu sparen.

## Literaturverzeichnis

# Methoden der Künstlichen Intelligenz

*Thomas Bertram, Stefan Paschek*

*Hochschule Karlsruhe*
*Institute of Materials and Processes (IMP)*
*Karlsruhe, Deutschland*
*thomas.bertram@h-ka.de, (+49) 721 925 -2087*
*stefan.paschek@h-ka.de, (+49) 721 925 -1805*

**Zusammenfassung**. *Dieses Paper liefert einen Überblick über gängige Verfahren der Künstlichen Intelligenz. Dabei wird primär versucht ein Verständnis für grundlegende Verfahren zu vermitteln. Zuerst werden ausgewählte Verfahren des Machine Learning und anschließend ausgewählte Verfahren des Deep Learning beschrieben.*

**Schlüsselworte**. *KI, Künstliche Intelligenz, Machine Learning, Deep Learning, Data Science, Algorithmen*

## 1   Definition KI

Die Künstliche Intelligenz (KI) beschreibt ein Wissenschaftsfeld, das sich mit der Generierung, Verwaltung und Erzeugung von Wissen beschäftigt. Zwei wichtige Unterkategorien der KI sind Machine Learning sowie Deep Learning.

Machine Learning beschreibt Verfahren, die es ermöglichen anhand vorhandener Daten ein Problem nachzubilden. Die Verfahren können Daten einer Bedeutung zuordnen (Klassifikation), einen Wert vorhersagen (Regression) oder Muster in Daten finden beziehungsweise diese gruppieren (Clustern). Die Deep Learing Verfahren beschreiben ein Teilgebiet der Machine Learning Verfahren. Sie befassen sich mit tiefen Neuronalen Netzen. Diese Verfahren eignen sich besonders für komplexe Aufgaben wie die Klassifikation von Bilddaten.

Ausgewählte Methoden dieser Themengebiete werden in diesem Paper genauer erklärt.

## 2   Klassifikation vs. Regression vs. Clustering

Mit dem Begriff künstliche Intelligenz wird meist ein Verfahren gemeint, das eine Problemstellung im Bereich Klassifikation, Regression oder Clustering lösen soll.

Bei der Klassifikation wird versucht, einen Datenpunkt zu interpretieren beziehungsweise die Zugehörigkeit zu einer Klasse zu bestimmen (Bsp.: IO/NIO). Die Verfahren hierzu werden an einem vorhandenen Datensatz trainiert. Der Datensatz muss dabei die zu messende Größe und eine dazugehörende Klasse als Label aufweisen. Beim Training wird die Information in den Trainingsdaten verwendet, um den Algorithmus so zu parametrisieren, dass neue Daten in die entsprechenden Klassen des Trainingsdatensatzes eingeteilt werden können (Abbildung 1a).

Bei der Regression wird im Gegensatz zur Klassifikation nicht die Bedeutung der Daten ermittelt, sondern ein Wert vorhergesagt (Bsp.: Temperatur vorhersagen). Auch hier werden die Algorithmen an Datensätzen trainiert, die das Problem abbilden. (Abbildung 1b)

Das Clustering beschäftigt sich mit der Analyse von Daten. Hier werden ähnliche beziehungsweise zusammengehörige Daten zu Gruppen, sogenannten Clustern, zusammengefasst. Die Trainingsdaten müssen dabei nur Features, also die entscheidenden Merkmale, enthalten. Labels werden aufgrund des Clusterings in diesem Verfahren nicht benötigt. Der Algorithmus sucht selbstständig nach Grenzen, um die Daten in sinnvolle Cluster zu unterteilen. Zusätzlich können Zusammenhänge zwischen verschiedenen Parametern gefunden werden. Um die Ergebnisse richtig interpretieren zu können, ist Domänenwissen unumgänglich. (Abbildung 1c)

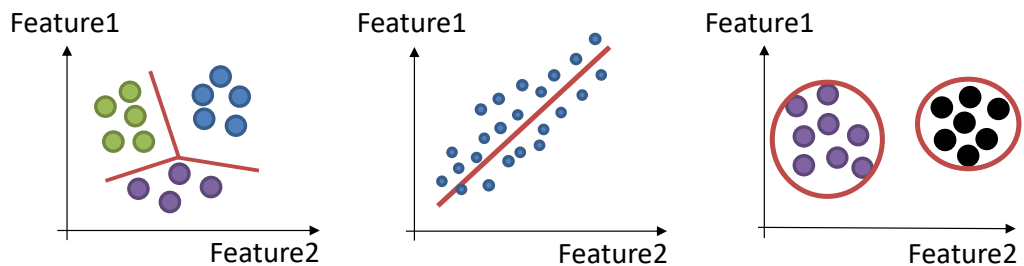

**Abbildung 1:** Von links nach rechts: Klassifikation (a) oder Grenzen finden, Regression (b) oder Werte ermitteln, Clustering (c) oder Daten zusammenfassen. Die roten Linien zeigen die Funktionalität der KI.

## 3 Datenaufbereitung

Da die Verfahren Wissen aus und mit Daten generieren, ist die Datenaufbereitung ein wichtiges Thema in der KI. Die Umsetzung und das Training der Algorithmen nehmen dabei meist nur $20 - 30\%$ des Gesamtaufwands eines KI Projektes ein. Datenakquise sowie die Aufbereitung der Daten entsprechen ca. 70% - 80% des Aufwands [1] (Abbildung 2).

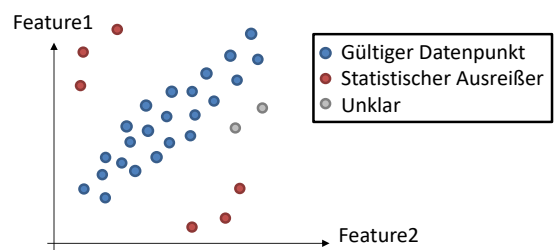

**Abbildung 2:** Messreihe mit Messfehler

Bei der Datenaufbereitung können mehrere Probleme auftreten. Es muss berücksichtigt werden wie mit statistischen Ausreisern umgegangen wird und ob es möglich ist, diese nicht zu berücksichtigen. Gerade im Randbereich des Gültigkeitsbereiches ist dies oft nicht klar. Hierbei ist wieder die Erfahrung von Spezialisten mit Domänenwissen gefordert. Außerdem können Daten ungleich verteilt vorliegen. Soll beispielsweise ein Prozess überwacht werden, gibt es deutlich mehr Daten für einen IO Prozess als einen NIO Prozess. Dieses Ungleichgewicht kann zu einer unterschiedlich starken Gewichtung beider Zustände führen (Abbildung 3). Dies ist vor allem bei Klassifizierungsalgorithmen ein Problem.

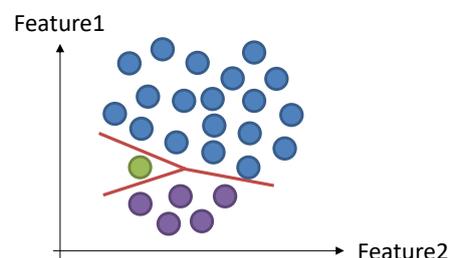

**Abbildung 3:** Ungleich verteilte Klassen können die größere Klasse präferieren

## 4 Ausgewählte Methoden Machine Learning

Im Folgenden werden gängige Methoden der künstlichen Intelligenz erklärt. Dabei liegt der Fokus auf dem Verständnis für die Verfahren und nicht den mathematischen Prinzipien dahinter.

### 4.1 Entscheidungsbäume

Entscheidungsbäume liefern eine definierte Vorgehensweise, um einen Wertebereich soweit zu unterteilen, bis eine Aussage über die kleinsten Teilstücke getroffen werden kann. Werden Daten mit dem

Entscheidungsbaum analysiert, wird bestimmt, zu welchem kleinsten Teilstück der Datenpunkt gehört. Der Aufbau dieses Entscheidungsbaums gibt dem Verfahren seinen Namen. Ausgehend von Wurzelknoten werden die Daten durch Bedingungen verzweigt. Ist keine weitere Verzweigung mehr möglich, erfolgt eine Zuweisung eines Wertes innerhalb eines Blattknotens. Dabei können Entscheidungsbäume sowohl für Regressions- als auch für Klassifikationsprobleme genutzt werden. (Abbildung 4).

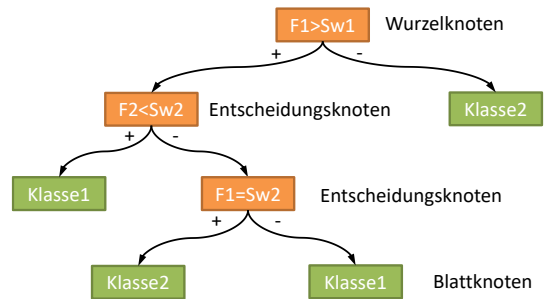

**Abbildung 4:** Aufbau eines Entscheidungsbaumes (F → Feature/ Sw → Schwellwert)

Der große Vorteil von Entscheidungsbäumen ist die die nachvollziehbare Struktur. Diese ist verhältnismäßig leicht zu verstehen und kann bei kleinen Bäumen sogar manuell angepasst werden. Außerdem benötigt das Verfahren nur wenig Rechenaufwand. Ein Nachteil der Entscheidungsbäume ist, dass durch die Entscheidungen eine Diskretisierung der Information in den Daten stattfindet (Abbildung 5).

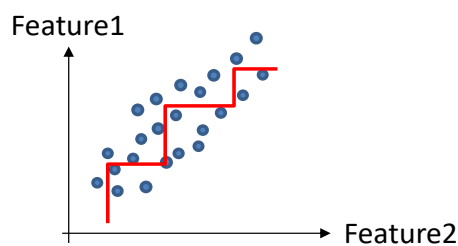

**Abbildung 5:** Diskretisierung durch Entscheidungsbäume

## 4.2 K-Nearest Neighbors

Der K-Nearest Neighbor Algortihmus versucht Daten anhand eines vorhandenen Datensatzes zu klassifizieren. Dabei wird von einem neuen, zu klassifizierenden Datenpunkt der Abstand zu allen vorhanden Datenpunkten berechnet. Die Zugehörigkeit der K nächsten Nachbarn zu einer Klasse, wobei K je nach Aufgabe gewählt werden kann, gibt Aufschluss darüber, zu welcher Klasse der neue Datenpunkt gehört.

Der Algorithmus muss nicht trainiert werden, um eine Aussage über neue Datenpunkte zu generieren. Außerdem ist das Ergebnis meist gut nachvollziehbar. Hinzu kommt, dass sich die Zuverlässigkeit mit der Größe des Datensatzes verbessert. Auf der anderen Seite nimmt die Geschwindigkeit, mit der der Algorithmus ausgeführt werden kann, mit steigender Größe des Datensatzes ab. Besonders schwierig sind Datenpunkte im Randbereich beziehungsweise Ausreißer zu klassifizieren, da diese zu einer Fehlklassifikation führen können. Ein weiterer Punkt ist die große Abhängigkeit der Vorhersage von der Anzahl der Nachbarn. So kann eine Veränderung des Parameters vor allem an den Randbereichen zu einer großen Varianz der Ergebnisse führen (Abbildung 6). Der Algorithmus reagiert außerdem empfindlich auf starke Ungleichgewichte zwischen den einzelnen Klassen im Datensatz.

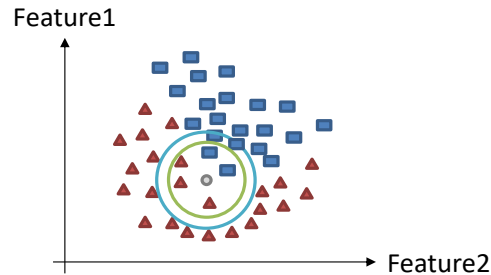

**Abbildung 6:** K-Nearest Neighbors mit unterschiedlicher Anzahl an Nachbarn. Im grünen Kreis (fünf Nachbarn) wird der neue graue Datenpunkt als Dreieck, im blauen Kreis (sieben Nachbarn) als Viereck interpretiert.

## 4.3 Support Vector Machine

Die Support Vector Machine ist ein Algorithmus zum Finden der optimalen Position der Trennlinie zwischen zweier Klassen eines Datensatzes. Im einfachsten Fall ist ein Datensatz durch eine Gerade in zwei Klassen aufzuteilen. Durch den Abstand der Datenpunkte der jeweiligen Klassen kann es unendlich viele Lösungen geben wie dieser Datensatz aufgeteilt werden kann. Deshalb wird die Separiergerade um eine Spanne erweitert. Der Algorithmus versucht die Position der Geraden so zu optimieren, sodass die Spanne um die Gerade möglichst groß wird. (Abbildung 7)

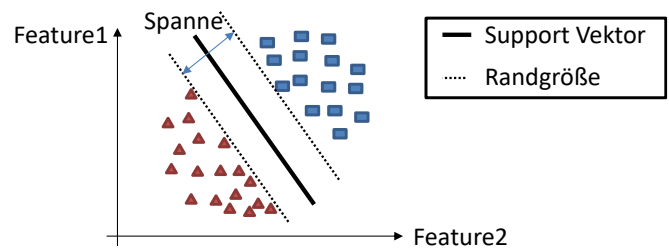

**Abbildung 7:** Veranschaulichung Support Vector Machine

Der Algorithmus kann Probleme gut generalisieren. Deshalb reichen auch oft kleinere Mengen an Datenpunkten, um den Algorithmus zu trainieren.

Da der Algorithmus versucht die Spannweite zu optimieren, ist er besonders empfindlich auf Rauschen und Ausreißer.

## 4.4 Gaussean Process Regression

Bei der Gaussean Process Regression werden verschiedene Funktionen durch vorhandene Datenpunkte gelegt. Der Mittelwert dieser Funktionen beschreibt eine Regressionskurve, die das Problem in den Daten nachbildet. Die äußersten Punkte dieser Funktionen beschreiben eine Einhüllende, welche die Schwankungsbreite der Funktionen angibt. (Abbildung 8)

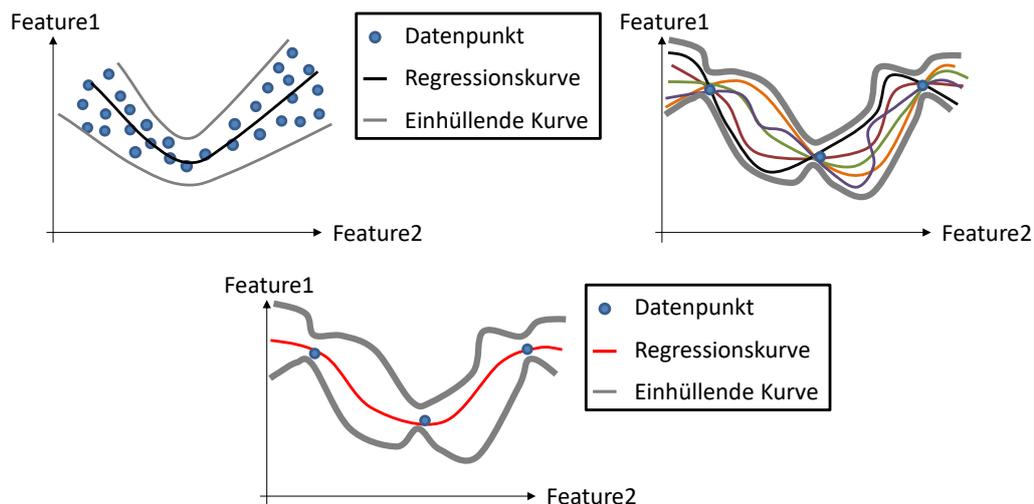

**Abbildung 8:** Veranschaulichung Support Vector Machine. Zu modellierendes Problem (oben links), Trainingsprozess an einzelnen Messpunkten (oben rechts), trainiertes Modell (unten)

Das Verfahren liefert ein Modell mit hoher Generalisierbarkeit. Außerdem wird die Schwankungsbreite zu jedem Messpunkt bestimmt (Einhüllende). Dadurch kann auch Rauschen modelliert werden. Beim Training ist jedoch eine hohe Rechenleistung nötig (steigt kubisch mit Anzahl an Datenpunkten). Außerdem sind die Modelle nur schwer interpretierbar.

## 4.5 Self Organizing Map

Die Self Organizing Map (SOM) dient dazu, Daten zu clustern. Dafür wird ein Netz aus Neuronen zufällig in einem Datensatz platziert. Von jedem Datenpunkt wird die Distanz zu allen Neuronen berechnet. Das Neuron mit geringster Entfernung zum Datenpunkt ist das Siegerneuron. Dieses wird anschließend in Richtung des Datenpunktes verschoben. Alle Nachbarneuronen werden ebenfalls in die Richtung verschoben - jedoch mit einer geringeren Distanz. Der Prozess wird so lange wiederholt, bis sich ein Gleichgewicht einstellt oder eine vordefinierte Anzahl an Durchgängen durchlaufen wurde. Bei diesem Trainingsprozess wandern die Neuronen über den Datensatz und bleiben bei Anhäufungen ähnlicher Datenpunkte hängen (Abbildung 9).

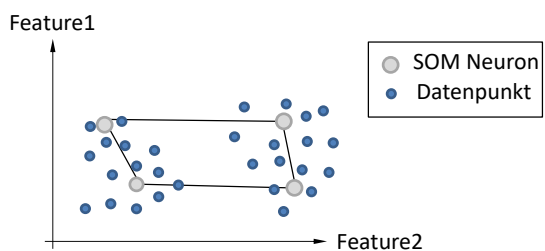

**Abbildung 9:** Clusterbildung mittels SOM

Das Verfahren ermöglicht die Dimensionalität von Problemen zu reduzieren. Außerdem ist das Ergebnis gut interpretierbar, da SOM zu jedem Feature eine Eingangsheatmap erzeugt. Die Clusterbildung kann sehr Aufschlussreich sein, hängt jedoch stark von der Anzahl der gewählten Neuronen ab. Durch die große Anzahl an Einstellparametern ist die Güte des Ergebnisses nur schlecht vorhersagbar und kann je nach Parameterwahl stark variieren.

## 4.6 Künstliche Neuronale Netze

Künstliche Neuronale Netze sind aus künstlichen Neuronen aufgebaut. Ein künstliches Neuron hat Inputs ($E_x$), welche gewichtet ($w_x$) aufsummiert werden. Anhand dieses Wertes wird über eine Aktivierungsfunktion ein output Wert erzeugt (Abbildung 10a). Durch Zusammenschalten dieser künstlichen Neuronen kann eine unterschiedliche Logik nachgestellt werden, die je nach Komplexität und Parametern verschiedene Aufgaben lösen kann (Abbildung 10b).

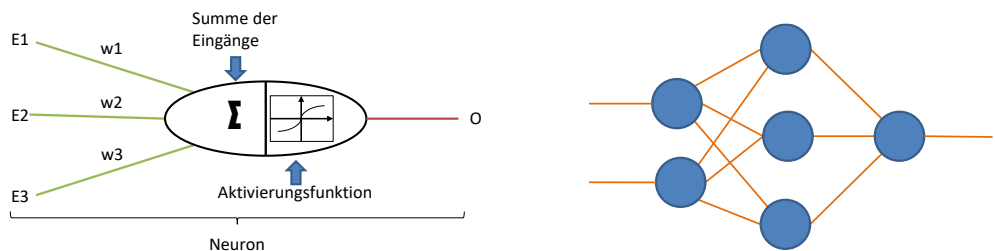

**Abbildung 10:** Künstliches Neuron (a) (links), künstliches Neuronales Netz (b) (rechts), blauer Kreis beschreibt ein Neuron

Die Künstlichen Neuronalen Netze eigenen sich sehr gut auch für nichtlineare Systeme. Ihre Approximationsfähigkeit ist mittels vorhandener Daten parametrierbar, wodurch viele unterschiedliche Aufgabenstellungen lösbar sind. Dafür müssen jedoch eine große Anzahl an Einstellparametern angepasst und eine passende Architektur gefunden werden. Durch die Komplexität sind schon kleinste Modelle meist nicht interpretierbar.

## 5 Ausgewählte Methoden Deep Learning

### 5.1 Was ist Deep Learning?

Deep Learning ist ein Teilgebiet des Machine Learning. Es befasst sich mit neuronalen Netzen. Wie bereits beschrieben, sind diese eine Kombination aus nichtlinearen Rechenoperationen, die miteinander verschaltet werden. Ein Netz ist meist in Ebenen aufgebaut. Es gibt eine Eingangs- und eine Ausgangsschicht. Dazwischen liegen die versteckten Schichten. Jede dieser Schichten ist aus parallel geschalteten Neuronen aufgebaut. Der Begriff Deep Learing kommt von der Anzahl der verstecken Schichten. Bei einfachen Problemen können hier zwei bis drei Schichten im Modell vorhanden sein. Sehr komplexe Modelle sind sogar mehrere hundert Schichten „tief" (Abbildung 11).

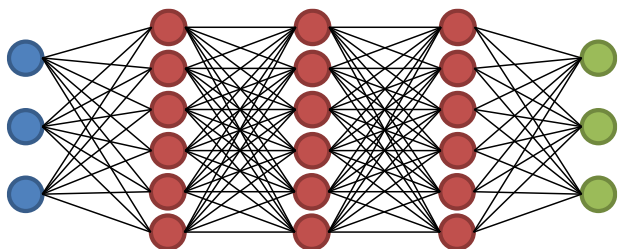

**Abbildung 11:** „Tiefes" Neuronales Netz

### 5.2 Funktionsweise von Deep Learning Netzen

Ein Neuronales Netz ist ein Modell, dass ein Problem nachbildet. Dabei wird zu einem Eingangsreiz an der Eingangsschicht ein Ausgangswert erzeugt. Die Verarbeitung des Eingangs passiert in den versteckten Schichten. Die Eingangsschicht regt die erste versteckte Schicht an. Diese erzeugt dann wiederum verschiede Ausgangswerte und regt die nächste Schicht an, bis am Ende Werte an der Ausgangsschicht erzeugt werden. Neuronale Netze können durch richtiges Training ihre Gewichtungsfaktoren parametrisieren, um generalisierte Lösungen für Problemstellungen zu finden. Dabei wird das Netz an die Informationen in den Trainingsdaten angepasst. Ein fertig trainiertes Netz kann dabei auch für neue, unbekannt Daten vorhersagen treffen (Abbildung 12).

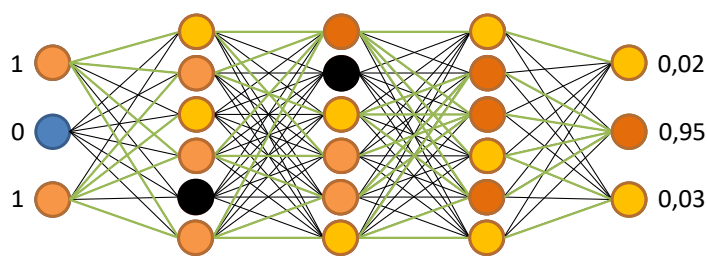

**Abbildung 12:** Angeregtes Neuronales Netz, Farbe der Neuronen und Verbindungen beschreibt deren unterschiedliche Aktivierung, Input erzeugt einen großen Output am zweiten Ausgangsneuron der Ausgangsschicht.

### 5.2.1 Training Neuronaler Netze

Beim Training werden die Parameter im Neuronalen Netz so angepasst, dass die Informationen in den Trainingsdaten approximiert werden. Dabei wird das Netz mit einzelnen Datenpunkten des Trainingsdatensatzes angeregt. Der Ausgang des Neuronalen Netzes wird anschließend mit einem Sollwert für diesen Datenpunkt, auch Label genannt, verglichen. Die Differenz zwischen Vorhersage des Netzes und dem Sollwert im Datensatz dient als Fehlergröße. Die Fehlergröße kann anschließend verwendet werden, um die Gewichtungsfaktoren des Neuronalen Netzes zu variieren und somit den Fehler zu minimieren. Dies geschieht durch Rückleiten des Fehlers vom Ausgangsneuron bis zu den Eingängen. Dabei wird für jedes Neuron der Einfluss auf den Fehler berechnet und dessen Parameter dementsprechend angepasst. Die Parameter mit dem größten Einflussfaktor werden so angepasst, dass der Fehler verringert wird. Dieses Verfahren wird auch Backpropagation genannt. Problematisch bei sehr tiefen Netzen ist Bestimmung des Einflusses eines Neurons auf den Ausgang über sehr viele Schichten hinweg, da es immer schwieriger wird den Einfluss eines Neurons auf das Ergebnis der Ausgangsschicht zu bestimmen (Vanashing Gradient). Damit das Netz möglichst gut ein Problem darstellen und generalisieren kann benötigt es einen möglichst großen Datensatz. Anderenfalls besteht die Gefahr, dass das Netz einzelne Punkte auswendig lernt. Außerdem muss die Architektur des Netzes zur Aufgabe passen. Ist das Neuronale Netz zu klein, kann keine richtige Generalisierung des Problems stattfinden, da die Komplexität des Problems zu hoch ist (Underfitting). Ist das Netz zu groß kommt es zum Auswendiglernen von Datenpunkten (Overfitting). Dabei kann das Netz mit den Trainingsdaten sehr gute Ergebnisse erzielen. Weichen die Daten jedoch von den Trainingsdaten ab werden meist nur sehr schlechte Ergebnisse erzielt. Im schlimmsten Fall können keine unbekannten Datenpunkte mehr erkannt werden (Abbildung 13). Der große Vorteil bei Neuronalen Netzen ist, dass diese Merkmale in Daten selbständig erkennen und gewichtet einem Ausgang zuordnen können. So werden auch Muster erkannt, die ein Mensch nicht erkennen kann.

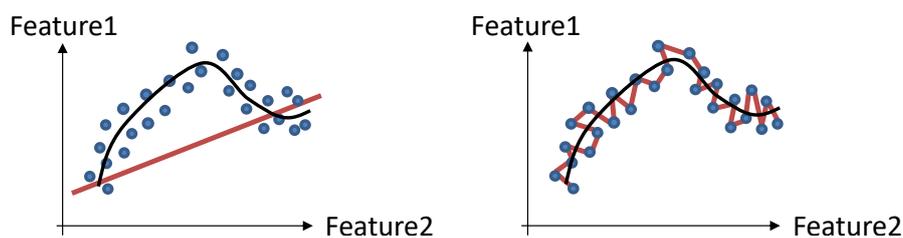

**Abbildung 13:** Underfitting (links), Neuronales Netz (rote Gerade) kann die Information in den Daten (schwarze Kurve) nicht nachbilden. Overfitting (rechts), Neuronales Netz lernt die Datenpunkte auswendig. Im schlimmsten Fall können nur noch die Datenpunkte selbst (blau) erkannt werden.

## 5.3 Convolutional Neural Networks

Die Convolutional Neural Networks (CNN) sind eine Erweiterung der Neuronalen Netze. Dabei wird dem Neuronalen Netz eine Musterextraktion mittels Faltungsoperationen vorausgeschaltet. Faltungsnetze (Convolutional Neural Networks) sind besonders in der Objekterkennung verbreitet, eigenen sich aber für jede Art der Mustererkennung. Dabei zeichnen Sie sich durch hohe Robustheit und gute Generalisierbarkeit aus. Auch hier ist der große Vorteil, dass die Netze selbstständig lernen Merkmale zu erkennen. Das CNN extrahiert Merkmale in den sogenannten Faltungsschichten und klassifiziert die Kombination der erkannten Merkmale im nächsten Schritt über ein Neuronales Netz.

### 5.3.1 Faltungsoperation und Pooling Operation

In der Bildverarbeitung ist es meist sehr aufwändig, bestimmte Features in Bilddaten zu extrahieren. Um den Aufwand zu verringern, werden sogenannte Filter Kernel verwendet (Bsp.: Sobel Operator). Diese bilden das Merkmal mit wenig Aufwand nach und überprüfen die Daten auf das Merkmal basierend auf der Faltung zweier Funktionen. Werden zwei Funktionen miteinander gefaltet und stimmen diese überein, entsteht ein großer Wert. Stimmen die Funktionen nicht überein wird ein kleiner Wert, idealerweise eine null berechnet (Abbildung 14).

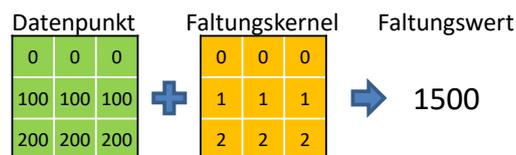

**Abbildung 14:** Faltungsoperation hohe Übereinstimmung (links), Faltungsoperation schlechte Übereinstimmung (rechts)

Bei der Faltungsoperation in Neuronalen Netzen wird ein Filterkernel verwendet, der ein Merkmal beschreibt. Dieser wird mit jedem Datenpunkt, bei Bilddaten mit jedem Pixel (und seinen Nachbarpixeln), multipliziert. Das Ergebnis wird aufsummiert. Ist das Ergebnis groß, ist an dieser Stelle das Merkmal, das der Filterkernel beschreibt. Der Vorgang wird für verschiedene Filterkernel wiederholt. Jeder Kernel erzeugt eine sogenannte Featuremap, eine Karte, in der einzelne Merkmale der Daten gespeichert sind.

Auf diese Weise können schnell große Datenmengen entstehen. Zur Reduktion der Daten werden die Faltungsoperationen meist mit Pooling Operationen verbunden. Beim Pooling wird ein Bereich der Featuremap betrachtet und nur der dominanteste beziehungsweise größte Wert gespeichert (Max Pooling). Anschließend kann in der reduzierten Featuremap wieder eine Faltung durchgeführt werden. Beim Aufbau eines CNN werden die Faltungsoperationen und die Pooling Operationen als Schichten analog zu den Schichten des Neuronalen Netzes beschrieben.

Besonders beim Training eines Faltungsnetzes ist, dass zusätzlich zu den Neuronen der Ausgangsschicht auch die Faltungskernel angepasst werden. Dies ermöglicht es dem Netz, selbständig zu lernen welche Merkmale wichtig und relevant sind.

### 5.3.2 Transfer Learning

Transfer Learning ist eine besondere Form des Trainings von Neuronalen Netzen. Hier wird ein Netz an einem großen vorhandenen Datensatz trainiert, der ein ähnliches Problem, wie das zu lösende Problem darstellt. Ein Beispiel ist das Training mit Bilddaten für die Objekterkennung.

Mit dem großen Datensatz werden die Parameter im Faltungsnetz angepasst. Anschließend wird ein Großteil der Werte im Netz eingefroren und nur die relevanten, letzten Schichten mit den eigenen Daten trainiert. Dadurch ist es möglich, mit relativ wenigen Daten komplexe Aufgaben zu lösen. Vortrainierte Netze werden oft zusammen mit öffentlichen Datensätzen zur Verfügung gestellt, können also direkt für Transfer Learning verwendet werden.

Bei Faltungsnetzen kann das Verfahren besonders gut gezeigt werden. Hier wird im ersten Schritt das gesamte Netz mit einem großen Datensatz trainiert. Danach werden alle Parameter zur Merkmalsextraktion (die Faltungsschichten) gespeichert. Das Netz behält also die bereits gelernte Musterextraktion. Im nächsten Schritt muss nur noch die Kombination der extrahierten Muster im Neuronalen Netz mit den eigenen Daten neu trainiert. Es werden nur die letzten Schichten des Netzes angepasst. So kann das Netz neue, unbekannte Objekte mit der bereits gelernten Musterextraktion erkennen.

## 5.4 Recurrent Neural Networks

Recurrente Neuronale Netze (RNN) dienen zur Analyse von sequenziellen Daten. Beispiele dafür sind Text- und Spracherkennung. Die Recurrenten Neuronalen Netze sind Neuronale Netze, die um eine interne Rückkopplung erweitert sind. Diese speichert den aktuellen Wert des Netzes und kann diesen bei der Verarbeitung neuer Daten berücksichtigen. Bei klassischen Neuronalen Netzen wird einem Input ein Ausgangswert zugeordnet (Feed Foreward Neural Network). Die Erweiterung ermöglicht es RNNs einem Input mit Bezug auf die Vergangenheit einen Ausgangswert zuzuordnen. (Abbildung 15)

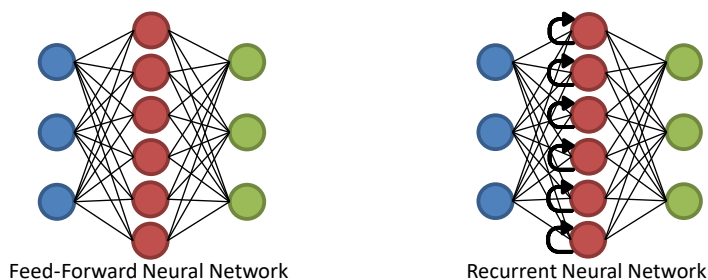

**Abbildung 15:** Vergleich eines Feed-Foreward Neuronalem Netz (links) mit einem Recurrent Neuronalem Netz (rechts)

Durch die Rückkopplung wird das Training deutlich komplexer, da nicht nur das neuronale Netz, sondern auch der Zustand des Neuronalen Netzes in der Vergangenheit berücksichtigt werden muss (Backpropagation Through Time). Durch diese Verschachtelung von neuronalen Netzen über die Zeit muss der Algorithmus schnell sehr tiefe Netze analysieren. Auch hier wird es mit zunehmender Tiefe immer schwieriger den Einfluss eines Neurons auf das Ergebnis zu bestimmen. Deshalb werden Informationen, die in der Vergangenheit liegen, auch mit jeder neuen Datenanalyse immer schwächer berücksichtigt.

Um das Problem zu adressieren wurden sogenannte Long Short-Term Memory (LSTM) Netze entwickelt. Diese erweitern die RNN um eine Schicht, die es ermöglicht Informationen zu speichern. Dadurch können auch sehr lange Datensequenzen verarbeitet werden, da die Informationen verfügbar bleiben.

## 6   Résumé

Dieses Paper beschreibt gängige Verfahren der Künstlichen Intelligenz. Dabei wurde speziell auf Machine Learning und Deep Learning Verfahren eingegangen. Die vorgestellten Verfahren eignen sich zum Lösen von Klassifikations, Regressions und Clustering Aufgaben. Je nach Anwendungsfall muss eine passende Methode ausgewählt werden. Für die Klassifikation können zum Beispiel Entscheidungsbäume oder tiefe Neuronale Netze verwendet werden. Bei der Regression bieten sich Verfahren wie die Gaussean Process Regression an. Beim Clustering können Self Organizing Maps zusammenhängende Datenpunkte und deren Abhängigkeiten voneinander erkennen. Für sehr komplexe Probleme bieten sich Deep Learing Verfahren an. Allerdings werden hier sehr viele Daten zum Training benötigt. Es bleibt anzumerken, dass viele der Methoden auch mehrere der Problemstellungen lösen können.

## Literaturverzeichnis

# Multidimensional Thermodynamic Attribute Clustering and Visualization using a Self-Organizing Map

*Matthias Feiner[1]\*, Stefan Paschek[1], Francisco Javier Fernández García[3],*
*Michael Arnemann[2], Martin Kipfmüller[1]*

*[1]Karlsruhe University of Applied Sciences, Institute of Materials and Processes (IMP),*
*Karlsruhe, Germany*
*matthias.feiner@h-ka.de, +49 721 925-2076*
*stefan.paschek@h-ka.de, +49 721 925-1805*
*martin.kipfmüller@h-ka.de, +49 721 925-1905*

*[2]Karlsruhe UAS, Institute of Refrigeration, Air Conditioning and Environmental Engineering (IKKU)*
*Karlsruhe, Germany*
*michael.arnemann@h-ka.de, +49 721 925-1842*

*[3]Polytechnic School of Engineering, University of Oviedo, C/Wifredo Ricart s/n,*
*Gijón, Spain*
*javierfernandez@uniovi.es, phone: +34 985 18 2112*

*\* Corresponding Author*

**Abstract**. *The Kohonen Self-Organizing Map (SOM) allows to visualize multi-dimensional data and to find correlation based on similar patterns. Here, various characteristics, measurements and control variables of a thermodynamic test stand (refrigeration cycle) with a swirl evaporator installed are displayed in a SOM in order to visualize correlations. The SOM was generated from a data set of experiments performed both, in steady-state conditions and during a dynamic test were the critical heat flux was reached.*

**Keywords**. *Neural network, swirl evaporator, high dimensional data analysis, Kohonen Self-Organizing Map*

## 1    Introduction

At the swirl evaporator test stand of the Institute of Materials and Processes at Karlsruhe University of Applied Sciences, 22 sensors are currently installed to log pressures, temperatures, mass flow and the performances of the main and auxiliary aggregates. A swirl evaporator enables high cooling capacity in a limited space. Its aim is to increase the critical heat flux by means of a turbulent flow in the evaporator. A detailed description of the swirl evaporator and its test stand can be found in (Feiner et al., 2018). The photo of the test stand is shown in Figure 1 and the parts of the experimental setup are shown in Table 1.

Multidimensional Thermodynamic Attribute Clustering and Visualization
using a Self-Organizing Map

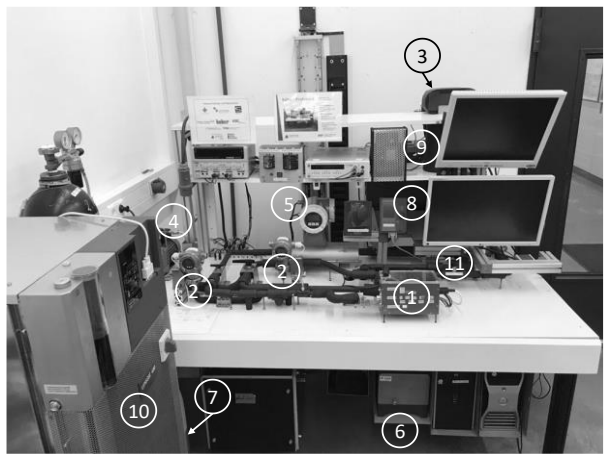

**Figure 1:** Photograph of the test stand (Feiner *et. al*, 2018)

The schematic drawing of the experimental setup is shown in Figure 2 and the parts of the experimental setup are also shown in Table 1. The system time constant is about $1 - 2$ seconds, thus the measured values are recorded every 500 ms, i.e. with a clock rate of 2 Hz, to account for the Nyquist-Shannon sampling theorem (Shannon, 1949). Since test series last several hours, a large amount of data is collected quickly (especially considering that some parameters can be tested independently like the length of the evaporator distance and heat load), but others influence each other (e.g. mass flow and pressure drop).

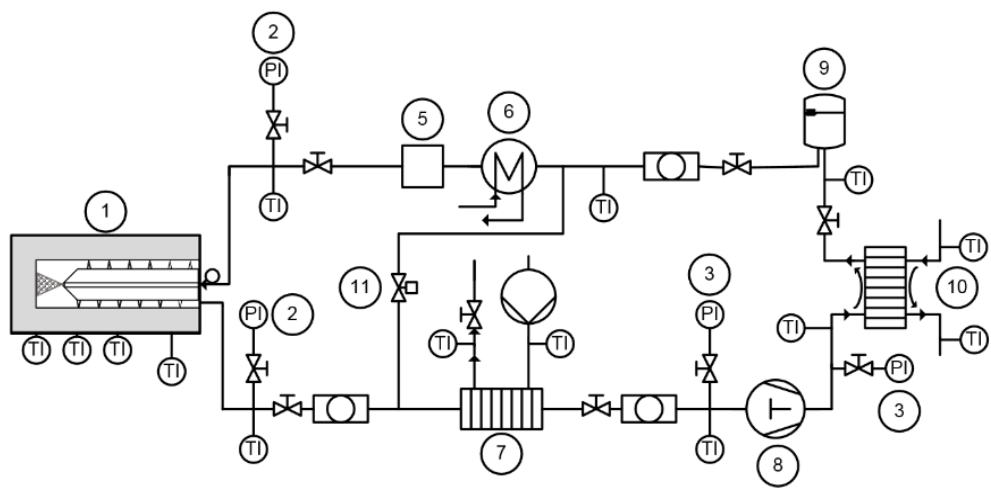

**Figure 2:** Experimental setup of the system (Feiner *et al*, 2018)

**Table 1:** Parts of the experimental setup (Feiner et al, 2018)

| | | | |
|---|---|---|---|
| 1 | Sample chamber with swirl evaporator | 7 | Post-evaporator |
| 2 | PI pressure sensor (F = 0.05 % o. m. v.) | 8 | Compressor (oil-free) |
| 3 | PI pressure sensor (F = 0.1 % o. m. v.) | 9 | Refrigerant assembler |
| 4 | Vacuum pump | 10 | Condenser |
| 5 | Mass flow meter | 11 | Step motor controlled expansion valve |
| 6 | Subcooler | TI | Thermocouples Type J |

A sketch of the swirl evaporator with the flow regimes and thermocouples are shown in Figure 3. The thermocouples measure the temperature in different locations of the housing to obtain a temperature field which later can be used for inverse thermal analysis and the estimation of local heat transfer coefficients.

Multidimensional Thermodynamic Attribute Clustering and Visualization
using a Self-Organizing Map

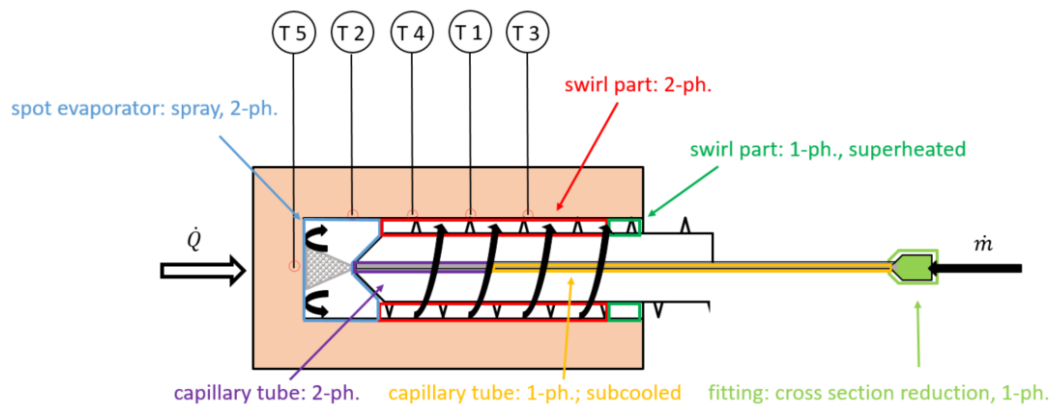

**Figure 3:** Sketch of the swirl evaporator with thermocouples and flow regimes (modified from *Feiner et al, 2021*)

For a better understanding of the interrelationships between the individual parameters, an AI-powered tool can assist in visualization and also detect correlations where the human observer would have missed it, due to the lack of comprehension skills of multidimensional data neuronal networks can be a fast and powerful technique that can be used to analyze this multi-dimensional data. They are able to deal with noisy data and to a certain point even with incomplete information. Neuronal networks have the ability to learn from experience as more and more data is presented and can be very effective especially in complex situations where it is not possible to define the rules or steps that lead to the solution of a problem. There are many kinds of artificial neural networks that can be classified by their learning methods, topology or applications. The learning methods can be subdivided into supervised and unsupervised neural networks. Supervised learning methods are preferred if the answer is already known and the aim is to find relationships between input and output variables, so they can be used to train the network (Russel & Norvig, 2010). For instance, Simple Recurrent Networks, Feed Forward Neural Network or Radial Basis Function, to give some examples. Their use is mainly for prediction. Their counterpart would be the so-called unsupervised learning methods. If the answer is not yet known to the researcher they can be used to find patterns or structures in the dataset (Hinton & Sejnowski, 1999). Because of their structure and operation mode they are called self-organized. There are different kinds of self-organized methodologies like unsupervised vector quantizer which is called a competitive algorithm or the Kohonen Self-Organizing Map (SOM) which is a competitive and cooperative algorithm (Kohonen, 1982). It can be used for both, classification and clustering, and therefore it provides information about the topology of the data (this method works even if there is only a small amount of data available). E.g., in Matlab colors like in a thermal picture of an infrared camera can be used to illustrate the classification of seismic attributes. E.g. Like in a thermal picture refers yellow to hot or a relatively high value, black refers to cold or a relatively low value. But because the colors are chosen because of their relative value to each other, here also lies a source of misinterpretation. To give an example: In data sets that fluctuate little, a small fluctuation can have a very drastic effect on the result even if the fluctuation is very small, say between 10.01 and 10.02 bar. Also, an actually constant value in the SOM can appear binary due to a "bit wiggle" (conversion from an analog to a digital signal). So in this case there are 3 attributes: black, red and yellow. The colors refer to the normalized values where black refers to the lowest and yellow to the highest value of the corresponding attribute. For example, if the minimum value of one attribute, e.g. pressure is 1 bar and the maximum is 12 bars, 1 bar would be displayed black, 12 yellow and 6 red. Values in between would be a mixture of these color- e.g. 9 would be orange etc.

## 2    Data generation and processing

With a sample rate of 2 Hz, the measured and controlled variables shown in Table 2 are written to a log file with the aid of a LabView program. The Self-Organizing Maps (SOM) shown in the two columns of Table 2 origin of the same experiment. Dataset A is the "big" data set where (after a 120-minute run-in period of the system) measurements were taken at 150 W for 1200 samples (10 minutes). After that the power was increased in 10 W increments until critical heat flux was reached at 210 W. The power increase until the critical heat flux density is reached was carried out with a control algorithm, which is why there is not necessarily the same time span between each power increase increment. The control algorithm always increases the power by 10 W as soon as the temperature in the housing of the evaporator has remained constant (i.e. fluctuations of less than 0.5 K) for 6 minutes. So it is not a fixed period but a waiting until a state.

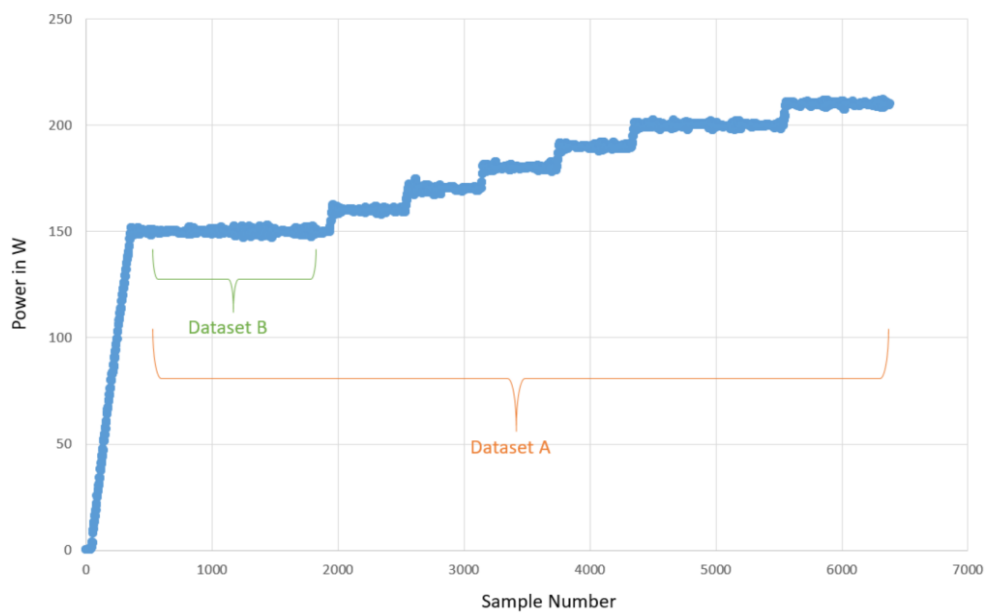

**Figure 4:** Power curve until the critical heat flux is reached

Data set B is the "small" data set where only the 60 minutes at 150 W steady state from data set A are considered. It can be stated that dataset A contains transient elements and data set B is a steady state. The SOM shows a grid of hexagons, which are called nodes, that organizes the measured and controlled variables which occur from the process described above. The user decides how many nodes to use. If there are not enough nodes, the data point refers to the nearest value. In this way, information can be lost and this should be avoided. On the other hand, too many nodes lead to a drastic increase in computing time and this might result in no pattern at all, even though, there is a correlation. Using hexagonal nodes instead of square ones has the advantage that only the 4 neighboring nodes are influenced whereas a hexagon has 6 neighbors. In this way, the SOM allows to visualize multi-dimensional data in fewer dimensions. In a self-organizing map, an N-dimensional state space is changed into N one-dimensional spaces that are represented in a 2-D map. A node is represented by 20×20 neurons arranged in a map. In this application, the SOM is placed in a 27-dimensional space, which are then visualized individually in 27 one-dimensional spaces with a 2-D SOM. This process is known as classification and can be used for vector quantization for data compression (Gray, 1984). Instead of two values represented by the colors: red and yellow; there is now only one information stored: orange. Each neuron of the SOM is a data object which takes a position in the 27-D state space. The weights of the neurons in the SOM then correspond to the coordinates of the measured values of the 27-D state space. A neuron in an artificial neural network is a data object that contains variables called weights and there are as many „weights" in a neuron as there are attributes; in this case 27. Because these weights are set randomly in the beginning, the patterns of the SOM may look different each time the process is performed. But even if the pattern looks differently, data which correlates will still be displayed in a similar pattern. The number of weights is called the dimensionality. Usually SOMs can't display time-dependent data but with a small tweak it is possible. In the first row there was a counter which was counting milliseconds. Through normalization and training, learning classification and clustering algorithms described below a certain pattern became visible. Low numbers are displayed black and high numbers yellow. Since this is a linear process, all colors are present to the same extent. Whenever a similar pattern of input 1 appears, there seems to be a correlation and the attribute could be time dependent.

Multidimensional Thermodynamic Attribute Clustering and Visualization using a Self-Organizing Map

## 3    Training and learning

Experimentally, it was found that a stable final value was reached after 200 runs, so every data point for each attribute is presented to every neuron repeatedly: in this case 200 times and with each repetition the neurons in a SOM network attempt to become like the input data presented. When the neuron received the data, it's weights are adjusted with the data values. This process is called training. Before training the neurons contain some random weighs. The orange data point has some component of red and yellow. During training the neuron is adjusted to the position of the input state space, based on the learning parameter. Thus, the neurons of the SOM are randomly positioned at the beginning of the training. Depending on how they are positioned, the visualization may change but the information remains the same. After training, the neuron now more closely resembles the trained data. The learning parameters determine how much to adjust. Adjusting too much results in a neuron to not learn about other data points. Adjusting to little will cause the neuron to not learn enough about the data point. All neurons are trained with all the data 200 times. Each time the data is used as input to the neurons is called an epoch. In the learning process the neuron learns (adjusts) a little more each epoch. The amount of movement is based on learning parameters. The user determines how many epochs are appropriate. 200 proved to be a good value here. The characteristic of a Kohonen SOM is that neurons not only adjust themselves to the data but also adjust the neighboring neurons. For this reason, it's also called cooperative learning. A Euclidean distance metric,

$$d_i^n = \sqrt{\sum_i^n (x_i - w_{i,s})^2} \qquad (1)$$

was chosen here to find the neuron $v$ closest to the data point (winning neuron). Hereby $d_i$ is the Euclidian distance, $w_i$ is the weight and $x_i$ the input data, for the index $i$. The neuron closest to the data point is the winning neuron and gets moved most in the direction of the data point using the update formula

$$W_v(s+1) = W_v(s) + \theta(u,v,s) \cdot \alpha(s) \cdot \big(D(t) - W_v(s)\big), \qquad (2)$$

with the weight vector $W_v(s)$ for the current step $s$. The weight vectors of the nodes in the neighborhood of the closest unit (including the closest unit itself) by pulling them closer to the input vector. The neighborhood function $\theta(u,v,s)$ gives the distance which lays in-between neuron $u$ and neuron $v$ in for the current step $s$ *(Kohonen and Honkela, 2011)*. The neighboring neurons of $u$ are also pulled towards the point but with a shorter distance. $D(t)$ is the Input vector for neuron $u$ and $\alpha(s)$ the learning rate which defines how quickly the neuron will adjust to the data presented. Cooperative learning explains why similar nodes in the SOM tend to be grouped together at the end of the process. Cooperative learning decreases with each epoch until it finally stops. At this point only competitive learning is performed during the final epoch. Competitive means it represents the data point which fits best but is not adjusting its neighbors anymore. When the data is highly dispersed the algorithm might not fully work. A sketch of the neural network basic structure is shown in Figure 5.

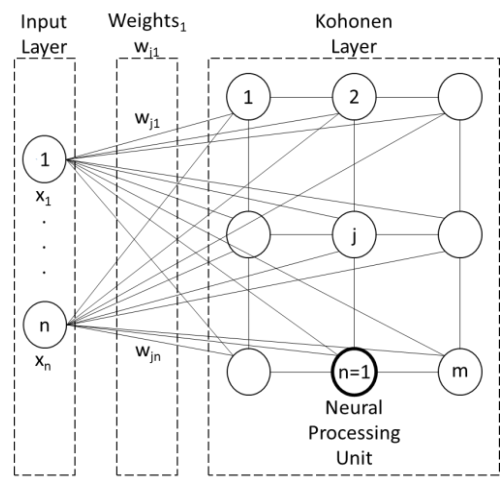

**Figure 5:** Kohonen Neural Network basic structure (Araujo et al. 2008)

Multidimensional Thermodynamic Attribute Clustering and Visualization
using a Self-Organizing Map

After all epochs are completed, every data point has some neurons which are within a close distance or are associated with it. Some neurons may not have data points which associate with them. Neurons with associated data points classify the data because all the data points have similar properties, in this case shown by its proximity to the found weights of the SOM. This process is called classification. Sometimes the data points cluster when cross-plotted in a diagram. After the classification the clustering takes place. Neurons will classify clusters in the data. They indicate similar signal responses (similar colors). Not all data has clusters which means that the attributes do not correlate well. The exact meaning of the clusters and outliners are left for interpretation. This is a tool that can be added to the more conventional seismic interpretation process and helps to reduce human bias in the analysis but can't be seen as substitution for such (Kohonen, 1984).

## 4    Analysis and discussion of the results

As shown in Table 2 in both data sets the correlation of the temperatures is clearly visible, i.e. Input 16, 17, 18, 19, (thermocouples which are placed radially around the swirl evaporator) 24 (the averaged radial temperature) and 26 (temperature front side). In data set A, a correlation with time is also visible as the power in data set A was gradually increased until the critical heat flux was reached. It is logical that temperatures correlate because a temperature field is established in the component via the heat source and sink. It is noticeable that in data set B the variables of the temperature still correlate but only a very weak correlation with time is visible. This is logical because dataset B was generated with a constant power of the heat source at 150 W. Furthermore, a correlation of the variables 22 and 23 (actual and target power) with the temperatures is visible in data set A, which is consequently not visible in data set B anymore. It is interesting to observe that the algorithm of the SOM still recognizes a correlation between actual and target power in data set A, but not in data set B. This is due to the fact that in data set B the target power is constant at 150 W and the actual power is only in the milliwatt range, but for the algorithm it oscillates arbitrarily around the target value. Input 5 and input 10, i.e. the low pressure after the swirl evaporator (suction line) and the temperature correlate inversely (only visible in data set A), since the pressure in the two-phase region and the temperature also correlates in opposite directions. Dataset B, input 12- the glycol input temperature is a good example for a binary dataset, where the actual variation was only 0.1 K (37.5 °C and 37.6 °C). A very neat, though not surprising, correlation is visible between input 10, the evaporating temperature, and inputs 16 –19 & 26, the temperatures of the thermocouples in the housing of the evaporator and also between Input 21 the temperature of the refrigerant after the compressor and Input 22 the heat load into the system.

**Table 2:** Input variables

| Input No. | Input Label | Dataset A transient | Dataset B steady state |
|---|---|---|---|
| 1 | Time | 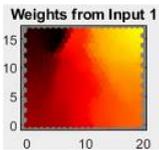 | 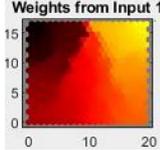 |
| 2 | Mass flux | 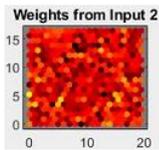 | 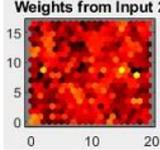 |
| 3 | Density of the liquid | 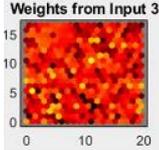 | 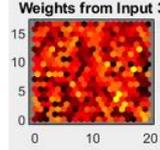 |

| Input No. | Input Label | Dataset A transient | Dataset B steady state |
|---|---|---|---|
| 4 | Temperature after subcooling | 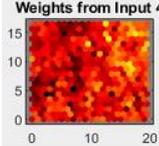 | 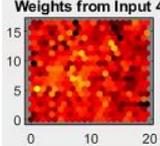 |
| 5 | Low pressure after swirl evaporator | 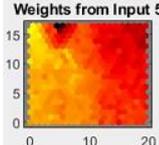 | 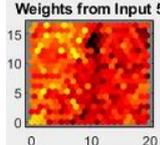 |
| 6 | Evaporation pressure | 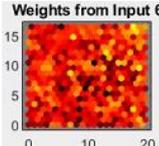 | 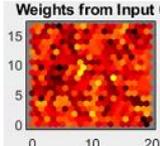 |
| 7 | Temperature before subcooling | 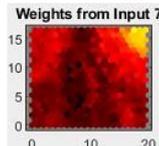 | 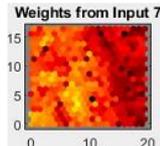 |
| 8 | Temperature after superheating | 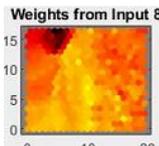 | 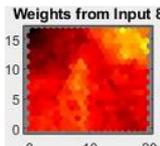 |
| 9 | Condensation temperature | 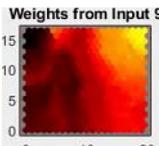 | 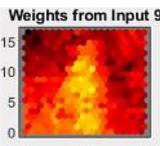 |
| 10 | Evaporation temperature | 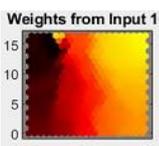 | 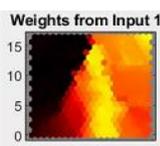 |
| 11 | Thermostat output temperature | 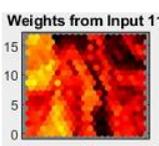 | 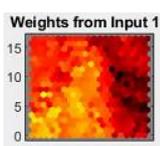 |
| 12 | Glycol input temperature | 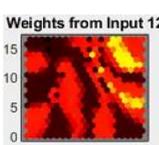 | 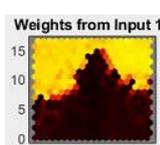 |
| 13 | Glycol output temperature | 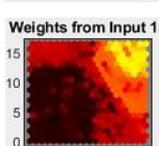 | 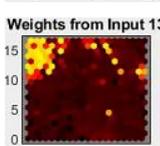 |

| Input No. | Input Label | Dataset A transient | Dataset B steady state |
|---|---|---|---|
| 14 | Condensation pressure | 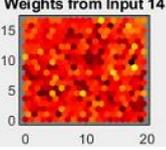 | 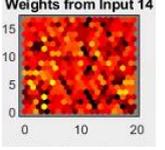 |
| 15 | Pressure after superheating | 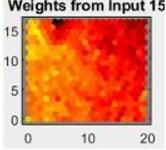 | 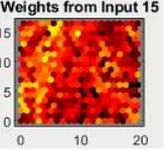 |
| 16 | Temperature thermocouple 1 | 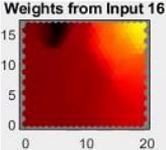 | 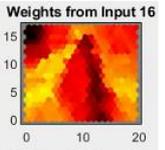 |
| 17 | Temperature thermocouple 2 | 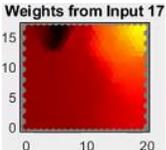 | 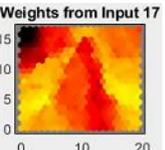 |
| 18 | Temperature thermocouple 3 | 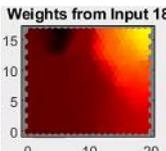 | 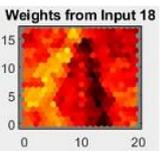 |
| 19 | Temperature thermocouple 4 | 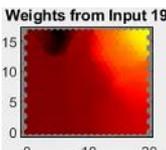 | 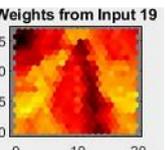 |
| 20 | Thermostat input | 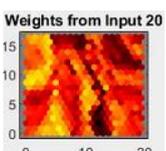 | 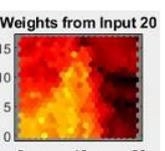 |
| 21 | Temperature after the compressor | 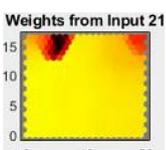 | 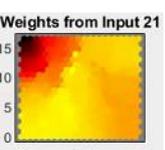 |
| 22 | Actual power | 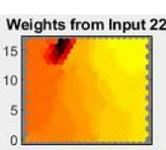 | 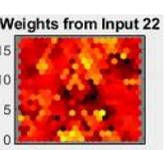 |
| 23 | Target power | 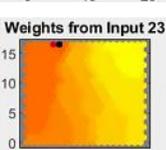 | 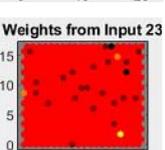 |

| Input No. | Input Label | Dataset A<br>transient | Dataset B<br>steady state |
|---|---|---|---|
| 24 | Mean value of radial temperature<br>(thermocouple 1 to 4) | 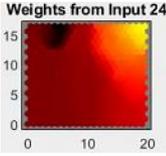 | 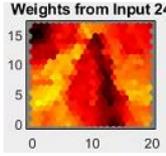 |
| 25 | Mass flow low pass filtered | 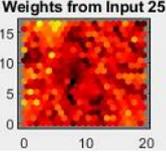 | 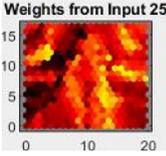 |
| 26 | Face temperature of the swirl evaporator<br>cavity (thermocouple 5) | 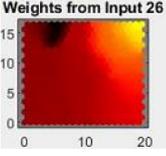 | 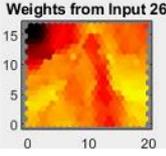 |
| 27 | Room temperature | 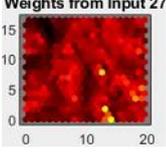 | 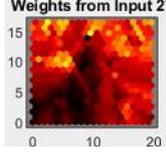 |

## 5 Summary

Kohonen SOM help classifying data, identifying clusters and outlaying data points. Some elementary but not surprising thermodynamic relationships could be represented, such as the correlation of the temperatures in the housing or the inverse correlation of evaporation temperature and pressure. So, it shows that this algorithm can be used for the analysis of a refrigeration cycle if enough data is collected. If the dataset is too small, information might not visible through the algorithm, as shown in the example of the small and large data set. Self-organizing maps can be used as a tool to reduce human bias because they assist in something the human brain is already very good at: pattern matching. But SOMs should in any case be seen as a supplement and not as a substitute for analysis. Last but not least, it should be stated, that correlation is not the same as causation and that if the amount of data collected is large enough, correlations will inevitably emerge somewhere (Granger, 1974). Perhaps the most prominent example of such a spurious correlation is the number of storks and the human birth rate in a country already stated by Simon in the 1950s (Simon, 1954).

## 6 Variables

| | | |
|---|---|---|
| $d$ | Euclidian distance | - |
| $D(t)$ | target input data vector | - |
| $s$ | current iteration | - |
| $t$ | index of the target input data vector | - |
| $u$ | index of the unit closest to the input data | - |
| $v$ | index of the node in the map | - |
| $w$ | neuron | - |
| $W_v$ | current weight vector of node $v$ | - |
| $x$ | Input data | - |
| $\alpha(s)$ | is a learning restraint due to iteration progress | - |
| $\theta(u, v, s)$ | neighborhood function | - |
| $\lambda$ | iteration limit | - |

# Potenziale für die Anwendung der Künstlichen Intelligenz (KI) in der mittelständischen produzierenden Industrie

*Martin Kipfmüller*

*Hochschule Karlsruhe*
*Institute of Materials and Processes*
*Karlsruhe, Deutschland*
*martin.kipfmueller@h-ka.de, +49 721 9251905*

***Zusammenfassung***. Im vorliegenden Artikel werden die Einsatzmöglichkeiten von KI in der produzierenden Industrie diskutiert. Ausgehend von einer Einordnung des Begriffes KI in Bezug auf den Einsatz im Ingenieurwesen, wird diskutiert, wie vor allem aus der systematischen Analyse von Produktionsdaten Mehrwerte für Unternehmen geschaffen werden können.

***Schlüsselworte***. *KI, Überblick, Anwendungsszenarien, Produktion*

## 1 Einleitung

Das Toyota Produktionssystem, Werkzeugmaschinen mit Parallelkinematik, globale Produktionsnetzwerke, Industrie 4.0 und jetzt künstliche Intelligenz (KI). Manchmal kann man sich nicht des Eindrucks erwehren, dass die Produktion genauso dem Zeitgeist unterworfen ist wie die Mode. Auch hier gibt es Trends, über die jeder spricht. Manchmal verändert sich unsere Organisation danach nachhaltig, wie zum Beispiel beim Einzug der Produktionssysteme in die Firmen. Andere Themen verschwinden wieder in einer Nische, weil sich das Erschließen der Potenziale doch schwieriger erwies als in der ersten Phase der Euphorie gedacht – dieses Schicksal hat zum Beispiel die Werkzeugmaschinen mit Parallelkinematik ereilt. Wieder, andere sind als Begriff schon abgegriffen, bevor die entwickelten Szenarien in der großen Breite umgesetzt wurden – das passiert gerade bei der Industrie 4.0

Und nun soll also die KI ihren Siegeszug in der Produktion antreten. Was ist davon zu halten? Wird das passieren? Wann? Und in welchem Umfang? Die Tagung KI4Industry kann diese Frage sicherlich nicht zur Gänze beantworten. Sie kann Ihnen aber eine Hilfestellung geben, sich eigene Meinung zu bilden. Wir zeigen Ihnen nach einem kurzen fachlichen Überblick an einigen Beispielen wie KI in Unternehmen gewinnbringend eingesetzt werden kann und welche Modelle es für den Einstieg gibt.

## 2 KI für die Industrie

Was ist das überhaupt, KI? Viele assoziieren damit im ersten vielleicht die sogenannte „starke KI": ein selbstständig denkendes elektronisches Gehirn, das unterschiedlichste Anwendungen mittels seiner Intelligenz lösen kann: Eine eigenständige künstliche Intelligenz, die – so die Vision einiger Denker und Visionäre – noch leistungsfähigere „künstliche Intelligenzen" hervorbringen kann, die ein Mensch nie zu entwickeln im Stande gewesen wäre. Die in einer endlosen Spirale der Optimierung den Menschen mit seiner bescheidenen „natürlichen Intelligenz" gnadenlos abhängt.

Dieses Gebiet wollen wir Visionären und Science-Fiction Autoren überlassen und uns mit der sogenannten schwachen künstlichen Intelligenz beschäftigen. Dieses, übrigens schon in den fünfziger Jahren entstandene Forschungsfeld, befasst sich mit „intelligenten Agenten", die ihre Umwelt beobachten und aus dieser Beobachtung Schlüsse ziehen, wie ein vordefiniertes Ziel am besten erreicht werden kann. Das heißt, über Sensoren erhält der Agent Wissen über die ihn umgebende Welt - die durchaus auch nur aus Algorithmen bestehen kann -, er verarbeitet dieses Wissen und führt dann eine Aktion aus (zum Beispiel über Aktuatoren oder auch wieder Algorithmen), um die Umgebung entsprechend seiner Zielvorstellung zu verändern.

Potenziale für die Anwendung der Künstlichen Intelligenz (KI) in der mittelständischen produzierenden Industrie

So allgemein diese Definition ist, so vielfältig sind auch die dabei angewendeten Methoden: Sie reichen von Entscheidungsbäumen und logische Operationen über Statistische Methoden (wie Bayes Theorem) bis hin zu natürlichen Strukturen nachempfundenen Systemen – den neuronalen Netzen.

Es gibt sie also gar nicht „die" KI – sondern KI ist eine Disziplin der Informatik, ein Werkzeugkasten aus dem sich auch ein Ingenieur bedienen kann, um seine Problemstellungen zu lösen. In den letzten Jahren haben Anwendungen, bei der sich die Entwickler aus diesem Werkzeugkasten bedient haben, immer beeindruckendere Dinge geschafft, so dass die Annahme, dass KI-Methoden auch in der Produktion viele Potenziale heben können, ganz und gar nicht als zu optimistisch einzustufen ist.

Auch wenn Industrie 4.0 als Begriff ein wenig überstrapaziert wurde, so ist doch der Einzug der Informationstechnologie in die Fabrikhallen ein ganz eindeutiger Trend. Nie war es zu so geringen Kosten und in so großen Mengen möglich Daten über die Produktion aufzunehmen und abzuspeichern. Und diese Daten bieten letztendlich die Basis für die Anwendung der Methoden der künstlichen Intelligenz.

Am Anfang steht zunächst meistens der Datenaufbereitung: Die aus teilweise unterschiedlichen Quellen (zum Beispiel Sensoren oder Steuerungen) stammenden Werte müssen in ein gemeinsames Format gebracht werden. Außerdem müssen die Daten synchronisiert werden, was meist durch unterschiedliche Aufzeichnungsfrequenzen erschwert wird. Trotz der rasanten Entwicklung des Speicherangebots müssen, gerade bei Produktionsanwendungen, oft weiterhin große Datenmengen reduziert und wichtige Features extrahiert werden: bei Untersuchungen zum Erodieren am IMP entstanden pro Minute mehrere Gigabyte Daten – das ist selbst mit modernen Systemen zu viel, wenn ein Prozess über mehrere Tage hinweg beobachtet werden soll.

Nachdem die Daten aufgenommen und geeignet strukturiert abgespeichert sind, können die eigentlichen KI Anwendungen gestartet werden. Es kann zum Beispiel nach Regressionen/Korrelationen in den Daten gesucht werden: wenn das Merkmal y sich verändert, dann verändert sich auch das Merkmal z. Daten können sortiert werden: Einige Bereiche eines mit einer Kamera aufgenommenen Bildes weisen zum Beispiel deutlich andere Graustufenwerte auf, als die Umgebung; sie gehören wohl zu einem Objekt, dass mit einer künstlichen Intelligenz erkannt werden soll. Es können die sogenannten maschinellen Lernprozesse stattfinden: wenn Sensor A den Wert x anzeigt, Sensor B den Wert Y und Sensor C den Wert Z, dann hat das in der Vergangenheit auf eine schlechte Produktqualität hingedeutet – wenn die Werte also wieder in die Richtung laufen, sollten die Prozessparameter angepasst werden. Soweit die Beispiele aus der Theorie – die in den Vorträgen von Thomas Bertram und Stefan Paschek noch detaillierter erläutert werden.

Was resultieren daraus jetzt aber für praktische Anwendungen in der Produktion: Als erstes großes Anwendungsgebiet sei die Bildverarbeitung genannt. Kameras, die zum Beispiel zur Qualitätskontrolle genutzt werden oder, um die Lage und Orientierung (Pose) von zu greifenden Teilen zu bestimmen, nehmen zunächst mal nur Informationen über die Lichtintensität und verschiedenen Bereichen des Bildes auf. Um daraus dann Gegenstände zu erkennen und deren Pose im Raum zu berechnen, ist eine Analyse der Daten notwendig bei der sich KI als ein mächtiges Werkzeug erwiesen hat.

Ein weiteres, noch viel direkter für die Produktion relevantes Gebiet ist der Betrieb nicht vollständig analytisch beschreibbarer Prozesse: Es existieren zwar für einige Produktionsprozesse bereits Formeln (zum Beispiel die Victor Kienzle Formel für die Zerspanung), aber selbst mit modernen Simulationstools sind Effekte wie Verschleiß in Wechselwirkung mit Maschine, Schmierstoff Temperatur usw. nur schwer zu beschreiben. Viele andere Prozesse – wie zum Beispiel das im Vortrag von Herrn Dr. Förster beschriebene Auftragen von Fetten – sind überhaupt nicht geschlossen analytisch beschreibbar. Dennoch ist es für die erfolgreichen Betrieb der Anlagen notwendig zu wissen, wann die Anlage gute Prozessergebnisse liefert, wann die Qualitätsanforderungen nicht mehr eingehalten werden können und vor allem was dagegen getan werden kann. Hier kann zum Beispiel das maschinelle Lernen ein sehr nützliches Werkzeug sein. Durch die Analysen großer Datenmengen, wird es möglich – zum Beispiel in Form der Parametrierung eines neuronalen Netzes – abzuspeichern, welche Prozess-, Material- und Umweltparameter zu einem guten und welche zu einem schlechten Prozessergebnis geführt haben. Nach dem Training eines solchen Netzes – das heißt, der Festlegung der Parametrierung des Netzes durch die Analyse eines großen Datensatzes – kann es also eine Prognose dazu abgeben, wie die Qualität der Prozessergebnisse/der Produkte bei den aktuellen Anlagenbedingungen sein wird. Hier liegt zunächst mal ein großes Einsparpotenzial in der Qualitätskontrolle, weil diese nicht mehr durch nachgelagerte Messungen erfolgen muss, sondern, direkt durch die Abfrage von Sensorwerten erfolgen kann. Es geht aber noch mehr: Wenn die Algorithmen entdecken, dass einige Umwelt- und/oder Materialparameter sich verändern, so können sie das im neuronalen Netz gespeicherte Wissen auch nutzen, um die aktiv beeinflussbaren Prozessparameter so anzupassen, dass der Prozess die Qualitätsanforderungen wieder erfüllt. Es ergibt sich also zusätzlich die Möglichkeit, den Ausschuss einer Anlage zu reduzieren/die Produktqualität zu verbessern, weil Prozessparameter individuell an die vorliegenden Material und Umweltbedingungen angepasst werden können. Hier liegt nach Meinung des Autors der größte Schatz in den Produktionsdaten verborgen, aber nicht der einzige:

Datenanalysen können außerdem helfen Verschleiß von Werkzeugen und Maschinen zu detektieren und diese zum wirtschaftlich besten Zeitpunkt, zu wechseln: Condition Monitoring. Sie können helfen die Anzahl von Fertigungsversuchen zu reduzieren, wenn bereits über den Prozess bekanntes Wissen intelligent eingesetzt wird: Bayes' Theorem – Dr. Salehi hat hierzu am IMP promoviert [1]. Und vieles mehr.

## 3   Fazit

Insgesamt ist die KI also eine Disziplin, die es ermöglicht, möglichst viel in Daten verstecktes Wissen zu extrahieren, zu analysieren und Entscheidungen daraus abzuleiten: in diesem Zusammenhang wird auch oft der Begriff des „Data Mining" verwendet. Zu diesem schon seit Jahrzehnten wachsenden Methodenwerkzeugkasten kommt ein weiterer Effekt hinzu, der die gerade sprunghafte Entwicklung des Feldes ermöglicht: Seit Jahrzehnten exponentiell wachsende Kapazitäten in der Computer- und Speichertechnik, sowie eine zunehmende Vernetzung machen heutzutage eine Aufnahme, Abspeicherung und Analyse von Produktionsdaten in Detaillierungsgraden und Volumen möglich, wie nie zuvor.

Nun gilt es diese Möglichkeiten zu nutzen. Wir Produktionstechniker müssen es wie Google, Facebook und Co. verstehen aus Daten Wert zu schöpfen. KI kann hierbei ein sehr mächtiges Werkzeug sein!

## Literaturverzeichnis

# KI- von der Wissenschaft in die industrielle Praxis

*Simon Ottenhaus, Michael Eder*

**KENBUN IT AG**
*Karlsruhe, Deutschland*
*simon.ottenhaus@kenbun.de, +49 721 781 503 02*
*michael.eder@kenbun.de, +49 721 781 503 02*

***Zusammenfassung.*** *Natürlichsprachliche Mensch-Maschinen-Interaktion im industriellen Umfeld ist eine Schlüsseltechnologie mit dem Potential Arbeitsabläufe zu vereinfachen und Prozesse intuitiver zu gestalten. Diese Veröffentlichung stellt KIDOU, den natürlichsprachlichen Assistent der KENBUN IT AG vor, der es ermöglicht Informationen einzugeben und abzurufen, ohne dabei die Hände zu benutzen. Die Spracherkennung und Sprachsynthese ist mittels Deep-Learning-Verfahren in einem Baukastensystem realisiert, das die Anpassung an spezifische Anwendungsfälle und die tiefe Integration in Kundensysteme ermöglicht. Da diese Anpassung sowohl Fachwissen im Anwendungsbereich als auch Kenntnisse der künstlichen Intelligenz (KI) in KIDOU benötigt, haben wir den KENBUN KI-Lifecycle entwickelt, der ein formalisiertes Verfahren zur Anpassung des Systems an Industrieanforderungen darstellt. Wir zeigen mögliche Anwendungsfälle von Sprachassistenten in der Industrie auf und beschreiben, wie der KI-Lifecycle die Risiken eines KI-Projekts frühzeitig abmildert und die KI-Potentiale systematisch zur Produktreife bringt.*

***Schlüsselworte.*** *Sprachassistenten, Künstliche Intelligenz, KI-Lifecycle*

## 1 Einleitung

Texte, Videos, Datenbanken, Tabellen, Zahlen, Audiodaten, Bilder und Co. - mit fortschreitender Digitalisierung ihrer Arbeitsabläufe generieren Industrieunternehmen eine ganze Flut an Daten. „Daten sind das neue Gold", so die derzeitige populäre Philosophie in der Geschäftswelt. Unternehmen setzen auf Data Science und erhoffen sich eine Effizienz und Qualitätssteigerung sowie eine erhöhte Mitarbeiter- und Kundenzufriedenheit und das möglichst entlang der gesamten Wertschöpfungskette. Die Potenziale sind riesig, können jedoch nur ausgeschöpft werden - und das ist die große Herausforderung - wenn die gesammelten Daten richtig, d.h. nutzerorientiert verarbeitet werden, ansonsten sind sie wertlos.

In der Praxis gestaltet sich die Aufbereitung als eher schwierig. Das liegt vor allem an drei Faktoren:

1. Die Daten liegen meist in verschiedenen Formaten und Datenbanken, in einer vorgegebenen Struktur und nicht nutzerorientiert vor.
2. Nutzer-Anfragen sind unstrukturierter Natur und müssen in strukturierte Daten übersetzt werden.
3. Es herrscht eine bidirektionale Diskrepanz zwischen natürlicher Sprache und maschineller Beschreibung.

Um diese Problemstellungen lösen zu können, müssen 1) alle Nutzeranfragen bzw. Eingaben in vorgegebene Strukturen überführt, 2) Strukturen in Texten bzw. in gesprochener Sprache erkannt, ggf. einem Kontext zugeordnet und anschließend extrahiert und 3) strukturierte Daten in natürliche Sprache umgewandelt werden. Als Lösungsansatz dienen wissenschaftliche Ergebnisse und Methoden, die in die Praxis transferiert und dem KI- Lifecycle (Abschnitt 3) zugeordnet werden.

KENBUN bietet mit seinem Produkt KIDOU (Abschnitt 2) eine funktionale Lösung der oben genannten drei Problemfelder und verschafft Unternehmen gleichzeitig eine natürlichsprachliche Schnittstelle zu internen Systemen. In der Industrie können Sprachassistenten vielseitig eingesetzt werden (Abschnitt 2.1), um den jeweiligen Kundenbedürfnissen gerecht zu werden, hat KENBUN das KIDOU-Baukastensystem entworfen. Das modulare System besteht aus mehreren „KI-Agenten", die in Abhängigkeit des Use Cases individuell kombiniert werden. KI-Projekte sind komplex und stellen Kunden und Anbieter gleichermaßen vor große Herausforderungen. Der KI-Lifecycle skizziert eine systematische Vorgehensweise in fünf Schritten und sensibilisiert hinsichtlich der Risiken sowie der strategischen Entscheidungen, die zu treffen sind (Abschnitt 3).

Neben der Produktentwicklung und den KI-Dienstleistungen bringt sich KENBUN unter anderem aktiv in das Forschungsprojekt FabOS[1] ein, in dem „die Grundlage eines Ökosystems für datengetriebene Dienste und KI-Anwendungen" geschaffen wird[2].

## 2   KIDOU: Natürlichsprachlicher Assistent

Mit KIDOU bietet KENBUN einen natürlichsprachlichen Assistenten für Kunden und Mitarbeiter an, der modernste Maschine-Learning-Technologien wie Text-to-Speech, Speech-to-Text, Intent Recognition, Sprechererkennung und Sentiment-Analyse kombiniert. Die Spracherkennung mit KIDOU ist äußerst präzise und lernfähig, da die zugrunde liegenden Deep-Neural-Network-Modelle flexibel an Kundenbedürfnisse und Wortschatz angepasst werden können. KIDOU kann On-Premises, in der Cloud, On-Edge und Hybrid betrieben werden. Nachfolgend werden mögliche Anwendungsfälle vorgestellt, die Architektur von KIDOU dargelegt, die Spracherkennung erklärt und ein Anwendungsfall aus der Praxis detailliert.

### 2.1   Anwendungsfälle und Einsatzmöglichkeiten

Von der Prozessoptimierung, bis hin zum Kundenservice und Marketing - die Einsatzmöglichkeiten von Sprachassistenten sind vielseitig. Insbesondere lohnt sich der Einsatz eines Sprachassistenten bei der Bearbeitung aller Informationen, obgleich strukturierter oder unstrukturierter Art. Das natürliche Interface ermöglicht eine intuitive Steuerung ganz ohne Maus und Tastatur (Voice-User-Interface). Selbst auf komplexe Systeme kann per Sprache zugegriffen werden.

Beispiele für Anwendungsfälle von KIDOU sind:

- **Dialogsystem für den Kundenservice (Lösung für Inbound Call Center).** Der Einsatz von KIDOU ersetzt das klassische „Drücken Sie eins für …"-System. Kund:innen können Ihre Anliegen frei, in natürlicher Sprache äußern, vergleichbar mit einem realen Dialog zwischen zwei Menschen. KIDOU versteht die vorliegende Fragestellung, kennt die vorausgehende Gesprächshistorie, kann flexibel auf das Gesagte des Kunden reagieren und Rückfragen stellen. Eine Verknüpfung des Back-End-Systems mit den kundeneigenen CRM- oder ERP-Systemen gewährleistet eine effiziente Dokumentation der Anrufe und ermöglichen eine fallabschließende Bearbeitung.
- **Zählerstanderfassung per Telefon.** KIDOU begrüßt die Anrufer und erfasst deren notwendige Stammdaten. Danach kann der Anrufer den Zählerstand einsprechen. Auch die Beantwortung von Fragen ist möglich.
- **Außendienst.** Mitarbeiter:innen im Außendienst verbringen viel Zeit im Auto, die effektiv genutzt werden kann. Vor dem Termin kann KIDOU den Mitarbeiter:innen Informationen zum Kunden geben. Nach dem Gespräch können die Mitarbeiter die Ergebnisse von KIDOU transkribieren lassen.
- **Wissensassistenten in der Industrie.** KIDOU kann natürlich sprachliche Fragen auf Basis von einer Wissensdatenbank beantworten. Das Wissen wird zunächst aus vorhanden Dokumenten ausgelesen und dem Mitarbeiter auf Anfrage als direkte Antwort präsentiert. Eine Anfrage „Was bedeutet Fehlercode 13 an Förderband X?" kann auf Basis der zugehörigen Betriebsanleitung direkt beantwortet werden: „Fehlercode 13 bedeutet ein verdecktes Lager."

- **Transkription von Konferenzen**. Besprechungen per Video-Konferenz werden immer häufiger. KIDOU kann automatisch ein Transkript der Konferenz erstellen, in dem die Sprecher unterschieden werden. Alle im Transkript festgehaltenen Beschlüsse und Aufgaben können automatisch extrahiert und an ein Backendsystem übergeben werden.

- **Sprachsteuerung von Apps.** KIDOU kann direkt in vorhandene Smartphone Apps integriert werden. So kann z.B. die Auswahl aus vorgegebenen Listen per Sprache erfolgen oder Zahlen eingesprochen werden. Ein Beispiel hierfür wird in Abschnitt 2.4 beschrieben.

- **Robotik.** KIDOU eignet sich ideal zur Sprach-Interaktion mit humanoiden Robotern und Industrierobotersystemen. KIDOU kann vordefinierte Anfragen und Anweisungen erkennen und dank der Sprachausgabe auch verständlich beantworten.

- **Extraktion von Informationen aus Gesprächen.** KIDOU extrahiert aus Gesprächsprotokollen vorab definierte Informationsobjekte, wie z.B. vereinbarter Termine, Beschlüsse, Zuweisung von Aufgaben, etc.

## 2.2 Architektur

Wie funktioniert das Kernstück der Sprach-KI, dem Teil der KI, der die Sprache des Benutzers entgegennimmt, verarbeitet, interpretiert und schließlich eine gesprochene Antwort ausgibt? Wie oft bei komplexen Systemen wird das Problem zunächst in Teilstücke zerlegt. Diese Teile lassen sich in drei Kategorien einordnen:

- Sprach Front-End: Anbindung von verschiedenen Sprachquellen, z.B. Mobilgeräte, Webseiten, Telefonie.

- Sprach-Kern: Schnittstelle zwischen Sprachsignalen und symbolischer Interpretation, z.B. als Text, als Intention oder als Dialog.

- Back-End: Anbindung von verschiedenen KI-Agenten und heterogenen Datenbanken an den KI-Kern.

Das *Front-End* übernimmt die Aufnahme der Sprache als Audiodaten, den Transport der Audiodaten zum Server, den Rücktransport der KI-Antwort als Audiodaten oder Text an die Ausgabe der KI-Antwort. Teil des Front-Ends ist die App oder GUI mit der der Nutzer mit KIDOU interagiert. Die Sprachsignale werden mit einem oder mehreren Mikrofonen quantisiert und diskretisiert und mit Puls-Code-Modulation (PCM) digitalisiert. Vor der Übertragung der Sprachsignale an den Server werden diese PCM Daten komprimiert, um Bandbreite zu sparen, z.B. in Mobilnetzen. Alle digitalisierten Sprachdaten werden auf dem Server zunächst von einer zentralen Instanz entgegengenommen: dem Audiorouter. Ähnlich wie ein IP-Router übernimmt der Audiorouter die Übermittlung von Audiodaten von verschiedenen Quellen zu mehreren Zielen. Der Audiorouter übernimmt auch die Dekodierung von komprimierten Audiosignalen und die Verwaltung von Verbindungen, sodass die Sprach-Antworten der KI immer zum richtigen Empfänger weitergeleitet werden, schließlich unterstützt KIDOU mehrere Benutzer gleichzeitig. Der Audiorouter führt auch eine Voice-Activity-Detection (VAD) durch, die Sprache von Stille und Hintergrundgeräuschen trennt. Die Kommunikation zwischen Audiorouter und den weiteren KI-Komponenten erfolgt nachrichtenbasiert über MQTT. Der Zwischenschritt über den Audiorouter ermöglicht eine hohe Modularität der einzelnen KI-Komponenten, da nicht jede KI-Komponente Verbindungsverwaltung, Dekompression, VAD und ähnliches übernehmen muss.

Der erste Teil des *Sprach-Kerns* ist die Umwandlung der Sprachsignale in symbolische Darstellungen. In der aktuellen Version von KIDOU werden die Sprachsignale des Nutzers analysiert und dabei drei Fragen beantwortet:

1. Was hat der Benutzer gesagt? ➔ Transkription der gesprochenen Sprach in Text.

2. Wer hat etwas gesagt? ➔ Zuordnung der Stimme zu einem bekannten oder unbekannten Sprecher.

3. Wie hat der Benutzer es gesagt? ➔ Wie ist die Stimmung des Benutzers?

Diese drei Fragen werden von je einer KI-Komponente beantwortet. Die gesprochene Sprache wird von der Speech-to-Text (STT) Komponente mittels State-of-the-Art Deep-Learning-Methoden in Text umgewandelt. Der nachfolgende Abschnitt „KIDOU-KI: Umwandlung von Sprache in Text (Speech-to-Text)" geht im Detail auf die Transkription der gesprochenen Sprach in Text ein. Die nachfolgenden KI-Module können mit diesen symbolischen Darstellungen einfacher trainiert werden, als direkt gesprochene Sprache zu verwenden. Ein Grund hierfür ist, dass Texte in großer Menge verfügbar sind und sich ein großer Teil der Sprachforschung auf die Analyse und Synthese von Texten konzentriert. Sprachmodelle, die in den letzten Monaten Schlagzeilen erzeugten (z.B. OpenAI GPT-Modelle[1], BERT[2], etc.) werden auf die Textverarbeitung trainiert. Die Ergebnisse und Fortschritte der textbasierten Sprachforschung macht sich KIDOU durch die Transkription der Sprache in Text zu Nutze. Die Sprechererkennung beantwortet die Frage „Wer hat etwas gesagt?" auch mittels Deep-Learning Methoden. Diese Sprechererkennung ist zum Beispiel für die Transkription von Gesprächen notwendig, wenn neben den gesprochenen Worten jeweils noch der Sprecher notiert werden soll. Die Erkennung der Benutzerstimmung kann in Call-Center Anwendungen hilfreich sein, so kann z.B. ein aufgebrachter Kunde sofort zu einem Mitarbeiter weitergeleitet werden.

Das *Back-End* von KIDOU erlaubt die Anbindung von verschiedenen KI- und klassischen Software-Komponenten als Informationsquelle oder Datenspeicher. Die Komponenten im Back-End können auf die symbolischen Informationen aus dem Sprach-Kern, also auf den transkribierten Text, die Sprecherzuordnung und die Stimmung des Benutzers zugreifen. KIDOUs Back-End ist modular und offen gestaltet, sodass verschiedene Anwendungen angeschlossen und untereinander integriert werden können. Beispiele für solche Anwendungen sind:

- Ein Inspektionsassistent, der Techniker:innen bei der Inspektion eines technischen Systems unterstützt. Der/Die Techniker:in führt die Inspektion durch und kann die Ergebnisse per natürlicher Sprache eingeben. Die Inspektionsergebnisse werden von KIDOU in symbolische, strukturierte oder tabellarische Strukturen übersetzt, die von nachgelagerten Systemen gespeichert und verarbeitet werden können.

- Ein Wissensassistent, der eine Wissensbasis aus einer Menge Dokumente extrahiert und dieses Wissen über gesprochene Fragen zur Verfügung stellt. Die Frage wird mit allen möglichen Dokumenten verglichen und eine oder mehrere Antworten werden aus den Quelldokumenten generiert. Das Ergebnis erhält der Benutzer in Form einer gesprochenen Antwort. Zusätzlich kann der Benutzer die zugrundeliegenden Quelldokumente abrufen.

- Ein Dialogsystem für ein Call-Center, das die klassischen „Drücken Sie eins für …"-Systeme ersetzt und einen Teil der Anfragen abschließend bearbeiten kann. Im Gegensatz zu den obigen zwei Beispielen ist KIDOU hier über mehrere Anfragen eines Benutzers aktiv und muss den aktuellen Zustand des Dialogs kennen. Dieser Zustand wird vom Dialogmanager gehalten und verwaltet. Der Dialog folgt dabei in der Regel vorgegebenen Pfaden, die vom Benutzer durch entsprechende Spracheingaben gewählt werden können. Der Dialogmanager kann zusätzlich im Back-End an Datenbanken angeschlossen sein, aus dem zusätzliche Daten abgerufen oder Benutzereingaben abgespeichert werden.

Die einzelnen Komponenten und der zugehörige Datenfluss sind schematisch in Abbildung 1 dargestellt.

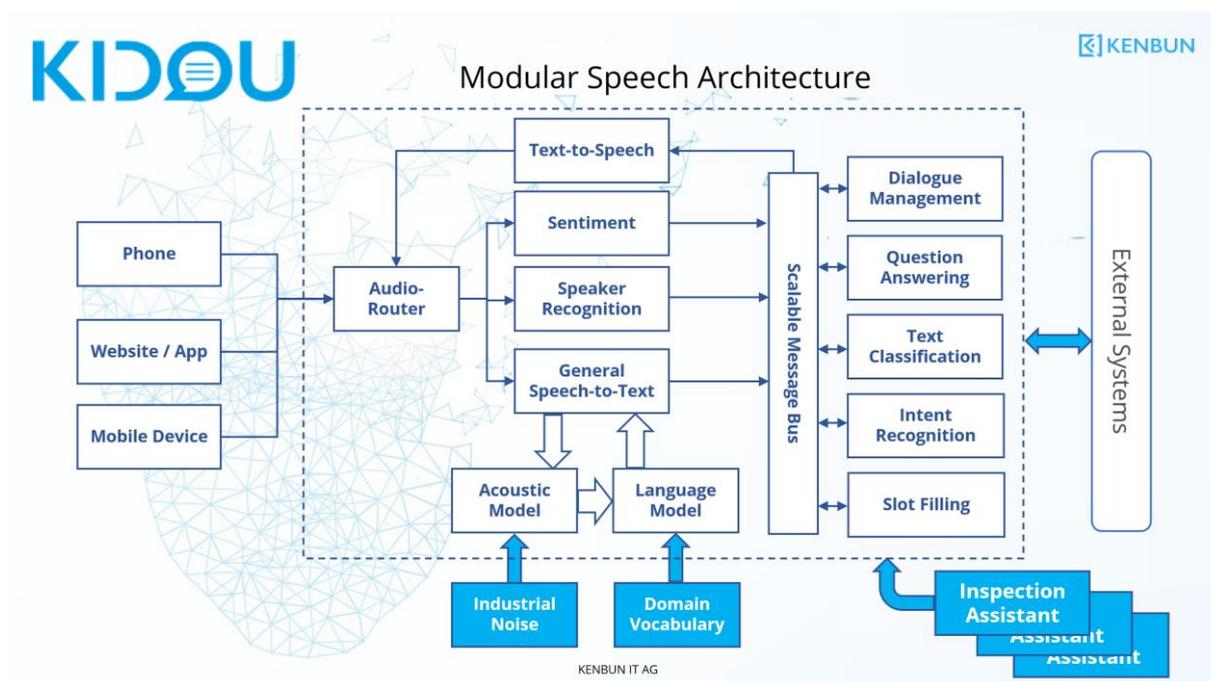

**Abbildung 1:** Modulare Sprach-Architektur. Von links nach rechts: Front-End (bis inkl. Audio-Router), Sprach-Kern (Speech-to-Text, Text-to-Speech, Sentiment und Speaker Recognition) und Back-End (ab Scalable Message Bus und nachgelagerte Sprach-Applikationen).

## 2.3  KIDOU-KI: Umwandlung von Sprache in Text (Speech-to-Text)

Ein zentraler Teil im Sprach-Kern ist die Umwandlung von Sprache in Text. KENBUN verfolgt hier den Ansatz den aktuellen Stand der Forschung stets im Auge zu behalten und die Speech-to-Text KI-Komponente auf Open Source Software aufzubauen. Die Spracherkennung ist ein aktives Gebiet des maschinellen Lernens und wird zur Zeit von Deep-Learning-Ansätzen dominiert. Die nachfolgende Beschreibung der Speech-to-Text-Komponente ist daher eine Momentaufnahme und kann sich mit fortschreitender Forschung in der Zukunft ändern. Die Grundprinzipien bleiben dabei weiterhin gültig.

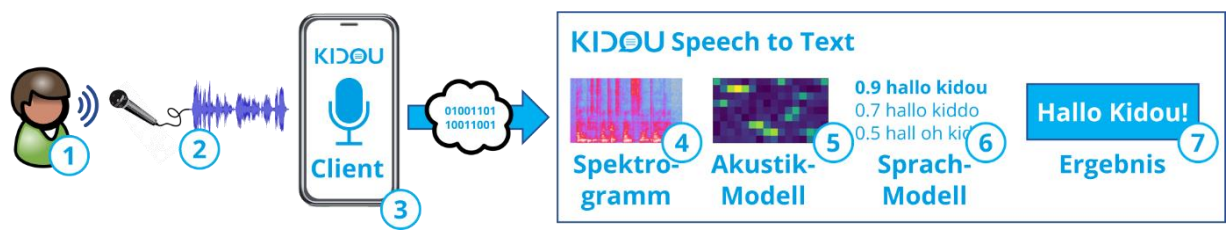

**Abbildung 2:** Spracherkennung mit KIDOU. Zusammenspiel von Client und Server.

Der KIDOU Client nimmt die Sprache des Benutzers mit einem Mikrofon auf und wandelt die akustischen Signale in digitale Puls-Code-Modulation (PCM) Signale um, siehe Abbildung (1). Die Signalabtastung erfolgt in der mit 16kHz und 16 Bit Auflösung (2). Vor der Übertragung zum KIDOU Server codiert der Client das Audiosignal mit dem effizienten OPUS Codec[3] und erreicht damit Kompressionsraten von über 90% (3). Der KIDOU Server erstellt aus den Sprachsignalen ein Spektrogramm, sodass aus dem 1D Audiosignal ein 2D Zeit-Frequenzbild entsteht (4). Auf der ersten Achse ist weiterhin die Zeit codiert, allerdings in größeren Zeitschritten. Auf der zweiten Achse ist die Frequenz codiert. Jeder vom Nutzer gesprochene Laut hat in diesem Frequenzbild eine charakteristische Form. Diese Formen erkennt das Deep-Learning getriebene akustische Modell und bestimmt zu jeden Zeitpunkt eine Wahrscheinlichkeitsverteilung über die gesprochenen Laute und Buchstaben. Über den Zeitraum der Eingabe ergibt sich daraus eine Matrix, die in der ersten Dimension die Zeit abbildet und in der zweiten Dimen-sion die möglichen Buchstaben. In jeder Zelle der Matrix steht ein Wahrscheinlichkeitswert, in der Abbildung farblich dargestellt (5). Das Sprachmodell errechnet aus dieser Wahrscheinlichkeitsmatrix verschiedene mögliche gesprochene Transkriptionen und bewertet jeden Text mit einer Konfidenz (6). Je nach Anwendungsfall wird die Transkription mit der höchsten Konfidenz zurückgegeben oder mehrere mögliche Transkriptionen (7).

---

[3] https://opus-codec.org/

## 2.4 Beispiel: KIDOU Inspektionsassistent

Sprachassistenten sind für private Endnutzer bereits verfügbar und können z.B. Suchanfragen im Internet übernehmen. In der Industrie haben Sprachassistenten großes Potential, sind aber bislang noch nicht weit verbreitet. Dies hat folgende Gründe:

- Viele Kunden wünschen Daten-Souveränität, d.h. die Spracherkennung und Verarbeitung soll und darf nur auf Firmen-eigenen Servern „On-Premises" durchgeführt werden.
- Für die angestrebten Anwendungsfälle ist spezifisches Fachvokabular notwendig, das bisherige Sprachassistenten nicht verstehen.
- Die Sprachanwendung muss oft eng mit Kundensystemen integriert werden.

Im Folgenden wird ein Anwendungsfall aus einem Kundenprojekt beschrieben. Aus Gründen der Vertraulichkeit ist der Anwendungsfall leicht abgeändert. Die drei oben genannten Forderungen sind allerdings unverändert gültig. Im Anwendungsfall „Inspektionsassistent" muss ein Techniker / eine Technikerin ein technisches System auf Abnutzungen und betriebsbedingte Mängel untersuchen. Dazu sind oft beide Hände notwendig und der / die Techniker:in muss z.B. in enge Räume des Systems hineinklettern. Zu dem System gibt es eine lange, vordefinierte Mängelliste, aus der / die Techniker:in auswählt. Die Mängel haben immer die Form (Mangel-Ort, Defekttyp, Schwere-Grad). Diese Mängel müssen dann auf einem Smartphone aus einer Liste ausgewählt werden. Zur Bedienung des Smartphones muss der / die Techniker:in oft Handschuhe ausziehen oder die Inspektion unterbrechen.

Genau an dieser Stelle setzt der KIDOU Inspektionsassistent an. Die Eingabe der Mängel ist damit über Sprache möglich, d.h. der Techniker / die Technikerin hat weiterhin beide Hände frei und muss die Inspektion nicht unterbrechen. KIDOU transkribiert die Sprache des Technikers / der Technikerin und bildet die Worte auf einen Mangel ab, indem Mangel Ort, Defekttyp und Schwere-Grad erkannt werden. Somit wird aus der gesprochenen Sprache symbolische Information extrahiert, die wieder maschinell verarbeitet werden kann. Die Mangelinformationen werden direkt an die bestehende App des Kunden übertragen, sodass keine weiteren manuellen Schritte notwendig sind. Bei der Spracheingabe gibt KIDOU aktives Feedback im Fehlerfall, falls die Eingabe z.B. unvollständig ist.

Für die Spracheingabe der Inspektionsergebnisse sind in diesem Anwendungsfall Fachbegriffe notwendig, die Sprachsysteme nicht immer verstehen. Damit KIDOU die notwendigen Fachbegriffe zuverlässig erkennen kann werden sowohl das akustische Modell als auch das Sprachmodell in einem Trainingsprozess angepasst. Die dazu notwendigen Trainingsdaten wurden teilweise mit KENBUN automatisch generiert als auch vom Kunden eingesprochen. Schließlich wird der KIDOU Inspektionsassistent auf den Kundenservern ausgerollt, somit bleiben alle vertraulichen Sprachdaten auf den Kundenservern.

## 3 KI-Lifecycle

Der stetige Fortschritt der Forschung im Bereich Maschinelles Lernen birgt großes Potential in der Industrie und den Konzernen. Jedoch stehen sowohl die möglichen Kunden als auch die Anbieter von KI-Lösungen initial vor Herausforderungen. Diese sind ähnlich wie bei klassischen Softwareentwicklungsprozessen, jedoch in einigen Aspekten stärker ausgeprägt. Gerade die Erstellung und Anwendung von KI-Lösungen bringt in der Industrie folgende Herausforderungen mit sich:

- Die Kunden bzw. Anwender aus der Industrie sind in der Regel keine KI-Experten.
- Die Data Scientists und AI Engineers der KI-Anbieter sind in der Regel keine Industrie-Experten.

Auch hinsichtlich der verfügbaren Daten, KI-Lösung und des zu lösenden Problems an sich sind die initialen Standpunkte und das Vorwissen der beiden Parteien oft weit auseinander, wie in Abbildung 3 illustriert. Diese Diskrepanz ist besonders ausgeprägt in der Sicht der KI-Lösung und des Anwendungs-
falls:

- Der Anwender sieht die initiale KI oft als Black-Box-Lösung.
- Der Anbieter sieht den Anwendungsfall des Kunden zunächst als Black-Box-Datenquelle.

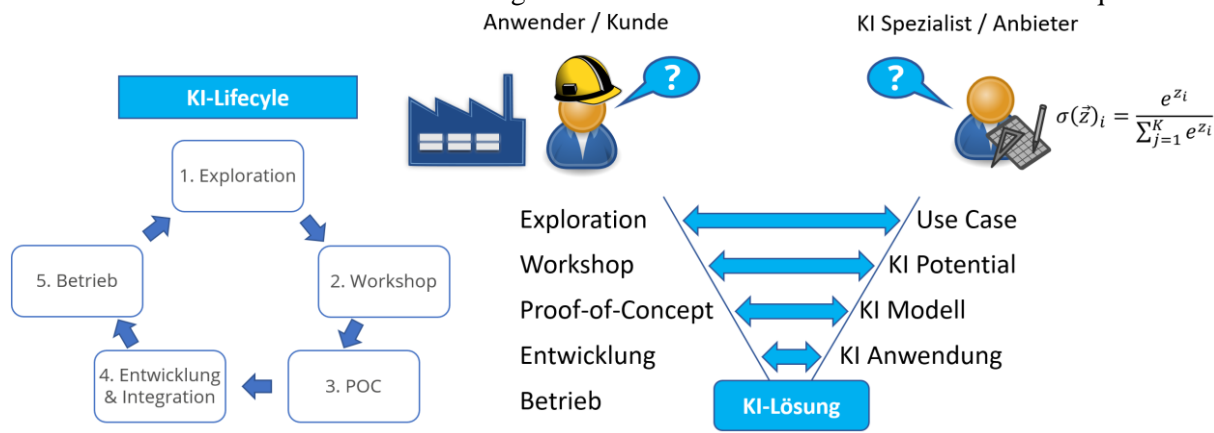

**Abbildung 3:** Der KI-Lifecycle schließt die initiale Diskrepanz zwischen Kunde und KI-Anbieter und resultiert in einer KI-
Lösung.

Diese Lücke iterativ zu schließen ist die Aufgabe des KI-Anbieters, KENBUN verfolgt hier den Prozess
des KI-Lifecycles, der es beiden Seiten ermöglicht sich effizient und lösungsorientiert zu nähern.

## 3.1 Phasen des KI-Lifecycles

Der KENBUN KI-Lifecycle besteht aus fünf Phasen und ist aus dem klassischen Data Science-Lifecycle
abgeleitet, der alle essentiellen Schritte hin zu einem einsatzfähigen Machine-Learning-Modell be-
schreibt. Diese Herangehensweise lässt sich in der Praxis auf nahezu alle KI-Projekte übertragen. Die
Umsetzung der einzelnen Phasen wird in den folgenden Abschnitten detailliert.

### 3.1.1 Exploration

Die Phase der Exploration deckt die Business Analyse ab und folgt direkt auf den ersten Kontakt zwi-
schen Kunde und Anbieter. Darin enthalten ist ein erstes Gespräch, in dem die KI-Potentiale ausgelotet
werden und Use Cases identifiziert werden. Besonders wichtig ist hier die Erfahrung des Anbieters die
prinzipielle Machbarkeit der Use Cases schnell zu bewerten. Oftmals hilft auch eine kreative Betrach-
tungsweise der einzelnen Bestandteile, da eine Umformulierung der Anforderungen ein unlösbares KI-
Problem in eine handhabbare Aufgabe verwandeln kann. Beispiele hier können das Sammeln von zu-
sätzlichen Trainingsdaten, die Installation von zusätzlichen Sensoren oder die Einschränkung des Use
Cases sein.

### 3.1.2 Workshop

Nach der Exploration folgt ein Workshop, in dem die Problemstellung konkretisiert wird und in funkti-
onale KI-Anforderungen übertragen wird. Das Potential des Use Cases wird von beiden Seiten gemein-
sam bewertet. Teil des Workshops ist auch die Lieferung eines initialen Datensatzes durch den Kunden.
Der Anbieter führt daraufhin eine detaillierte Datenanalyse aus und zeigt mögliche Gesetzmäßigkeiten
und KI-Anwendungsmöglichkeiten auf. Ziel des Workshops ist es dem Kunden die prinzipiellen Mög-
lichkeiten von KI-Lösungen für seine konkreten Anwendungsfälle näher zu bringen. Zusätzlich lernt
der Anbieter den Anwendungsfall besser kennen und kann dadurch die entstehenden Daten besser ver-
stehen und den Kunden bei der Datensammlung beraten. Im Workshop wird mit dem Data Engineering
begonnen.

### 3.1.3 Proof-of-Concept

Der Proof-of-Concept (POC) ist eine kritische Phase des KI-Projekts, denn hier muss der Anbieter de-
monstrieren, dass die vorgeschlagene KI-Lösung prinzipiell funktioniert und Mehrwert bringt. Daher ist
die Umsetzung des POC individuell auf den Kunden abgestimmt und während der Entwicklung eine
intensive Interaktion zwischen dem Kunden und dem Anbieter sinnvoll. Hier empfiehlt sich ein agiler

Entwicklungsprozess, den KENBUN bei der KI-Entwicklung befolgt. Ein zentraler Teil des POC ist das Data Engineering und die Modellentwicklung, integriert in einen funktionsfähigen Prototyp. Neben den Data Science-Aufgaben, beschreiben im Data Science-Lifecycle, ist aber auch klassische Software Entwicklung und Systemarchitektur ein zentraler Bestandteil des POC. So sind zum Beispiel die Integration von Client- und Server-Anwendungen relevant, ebenso das Zusammenspiel von bereits existierenden KI-Komponenten mit neu zu entwickelnden Lösungen in einem stimmigen Gesamtkonzept. Dabei kommen in der Regel auf den Servern containerisierte KI-Lösungen zum Einsatz. Für die Endgeräte werden oft kundenspezifische Apps entwickelt oder angepasst.

### 3.1.4  Entwicklung und Integration

Eine Stärke des KENBUN KI-Lifecycles ist die Möglichkeit die Lösung eng und unmittelbar mit Kundensystemen zu integrieren, wobei diese sowohl als Datenquelle als auch als Datensenke agieren können. Bei der Entwicklung einer skalierbaren KI-Komponente stehen daher die Anpassung bestehender KI-Komponenten an die Kundensysteme als auch die Integration neu entwickelter Komponenten in die Kundensysteme im Vordergrund. In der Entwicklungs- und Integrationsphase verschiebt sich der Arbeitsaufwand zunehmend von Data Science hin zu Softwareentwicklung und zu Big Data Prozessen.

### 3.1.5  Betrieb

Für den erfolgreichen Betrieb einer KI-Lösung sind oft Big Data-Lösungen notwendig, um hochfrequente Daten oder Daten mit großem Volumen verarbeiten und ggf. speichern zu können. Zentral ist hierbei die Möglichkeit zur Skalierung der KI-Anwendung. Im POC wird in der Regel eine Anwendung entworfen, in der jede Komponente nur einmal vorhanden ist. Beim Ausrollen der Anwendung On-Prem oder in der Cloud kann es jedoch notwendig sein manche Teile der Anwendung mehrfach bereitzustellen, da die verfügbaren CPU und GPU Ressourcen eines Rechen-Knotens nicht die gesamte Anfragelast abdecken können. Für den Betrieb von KI-Lösungen setzt KENBUN auf die selbst entwickelte Plattform *KIDAN*. KIDAN baut auf Open Source Lösungen auf und stellt Tools zur Wartung, Modell-Versionierung, Monitoring, Accounting und Zugriffskontrolle zur Verfügung. Durch den modularen Aufbau ist es möglich einzelne Teile während des Betriebs auszutauschen und Komponenten für verschiedene Anwendungsfälle wiederzuverwenden.

### 3.1.6  KI- und Big Data-Schulungen

Parallel zum KI-Lifecycle bietet KENBUN KI-Schulungen an, die es dem Kunden ermöglichen auf unterschiedlichen Leveln in die Thematik tiefer einzusteigen. Dies beginnt bei grundlegen Schulungen und Vorträgen auf Einsteiger-Level, die die Grundprinzipien von verschiedenen KI- und ML-Lösungen erklären. KENBUN ist zum Beispiel im KI-Weiterbildungsprogramm der CyberForum Akademie[4] vertreten. Es werden auch Themen und kundenspezifische Schulungen angeboten, u.a. zum Aufbau und Betrieb von Big Data-Systemen zur sicheren und effizienten Verarbeitung großer Datenmengen.

## 4  Fazit

KI-Lösungen in der Industrie haben großes Potential, da zum einen die Menge an verfügbaren Daten mit zunehmender Digitalisierung stetig steigt und zum anderen die Forschung rund um maschinelles Lernen stetige Fortschritte macht. Mit dem KI-basierten natürlichsprachlichen System KIDOU liefert KENBUN ein Baukastensystem, mit dem sich vielfältige Sprach-Use Cases realisieren lassen. Zentral dabei ist die Anpassbarkeit von KIDOU an domänenspezifisches Vokabular, Dialekte und Umgebungsgeräusche, als auch die Anpassung und Integration in Kundensysteme. KIDOU wird für das Forschungsprojekt FabOS zur Verfügung gestellt und in die Anwendungsfälle des Projekts integriert. KENBUN treibt mit der Entwicklung von kundenspezifischen Lösungen auch die Einführung von neuen KI-Lösungen aktiv voran. Mit dem KI-Lifecycle wird ein Prozess beschrieben, um die initiale Diskrepanz zwischen KI-Anbieter und Kunde zu schließen und die Risiken eines KI-Projekts frühzeitig zu entschärfen. So können Fortschritte aus der KI-Forschung zeitnah und lösungsorientiert in reale Anwendungsfälle der Industrie übertragen werden. In zukünftiger Arbeit wird angestrebt den KI-Lifecycle im FabOS Konsortium weiter zu formalisieren.

---

[4] https://www.cyberforum.de/akademie/ki-weiterbildung/

## Literatur

# KI gestützte Parametrierung eines Ventils zur Fettdosierung im Projekt AdaptValve

Stefan Paschek[1], Frederic Förster[2], Martin Kipfmüller[1]

[1]Hochschule Karlsruhe
Institute of Materials and Processes (IMP)
Karlsruhe, Deutschland
stefan.paschek@h-ka.de, +49 721 925 -2082
martin.kipfmueller@h-ka.de, +49 721 925 -1905

[2]Walther Systemtechnik GmbH,
Germersheim, Deutschland

**Zusammenfassung.** *Schmierfette sind komplexe Werkstoffe die starken Parameterschwankungen unterliegen. Moderne Pulsventile ermöglichen das Applizieren von Schmierfetten mit Genauigkeiten im mg und μg Bereich. Zur Systemidentifikation wurde ein Prüfstand entwickelt. Dieser Prüfstand erlaubt das automatisierte Auftragen von Pulspunkten mittels Pulsventile. Die gemessenen Kennlinien werden ausgewertet und anschließend daraus Features extrahiert. Die Features dienen zum Training von KI Algorithmen für die Vorhersage von Schmierfetteigenschaften. Um der Schmierfettmasse einen Erwartungswert und ein gültiges Toleranzintervall mit Standardabweichung zuzuweisen, reichen klassische Verfahren nicht aus, weshalb in diesem Projekt auf Verfahren der KI zurückgegriffen werden. Im speziellen wurde eine Kombination aus Self Organizing Map und K – Means Clustering Algorithmus verwendet. Dieser Algorithmus erlaubt sowohl die Bestimmung von Eingangsparametern zu einem definierten Massenerwartungswert, als auch die Vorhersage welcher Erwartungswert unter Verwendung von eingestellten Prozessparametern entstehen wird.*

**Schlüsselworte.** *Schmierfett, Pulsventil, Self Organizing Map, K – Means Clustering, Standardabweichung*

## 1 Einleitung

Das folgende Paper hat das Ziel das Projekt „AdaptValve" vorzustellen. Dabei wird die auftretende Problematik bei der Applikation von Schmierfetten und die dafür notwendige Anwendung von KI Algorithmen verdeutlicht. Im speziellen wird hierbei die Verwendung von Pulsventilen in den Fokus gesetzt. Das Forschungsprojekt AdaptValve läuft im Rahmen des „Zentralen Innovations Programms" gefördert durch das BMWi. Die Hochschule Karlsruhe Wirtschaft und Technik, Institut of Materials and Processes (IMP) und Walther Systemtechnik GmbH treten dabei als Projektpartner gemeinsam in diesem Projekt auf. Ziel dieses Kooperationsprojektes ist es adaptive Puls - und Sprühventile zur Schmierfettapplikation zu entwickeln.

Schmierfette selbst sind äußerst komplexe Werkstoffe. Sie unterliegen großen Parameterschwankungen. Die Schmierfetttemperatur beeinflusst die Viskosität und der Druck beeinflusst die Fließeigenschaften. Außerdem können die Lagerzeit und die Eigenmasse des Schmierfettes zur Separierung von Öl und Bindemittel führen was zu Bereichen mit unterschiedlicher Viskosität führen kann. Die Abbildung 1 zeigt dabei den Vergleich zwischen einem normalen Pulspunkt und einem Pulspunkt nach Separierung von Bindemittel und Öl.

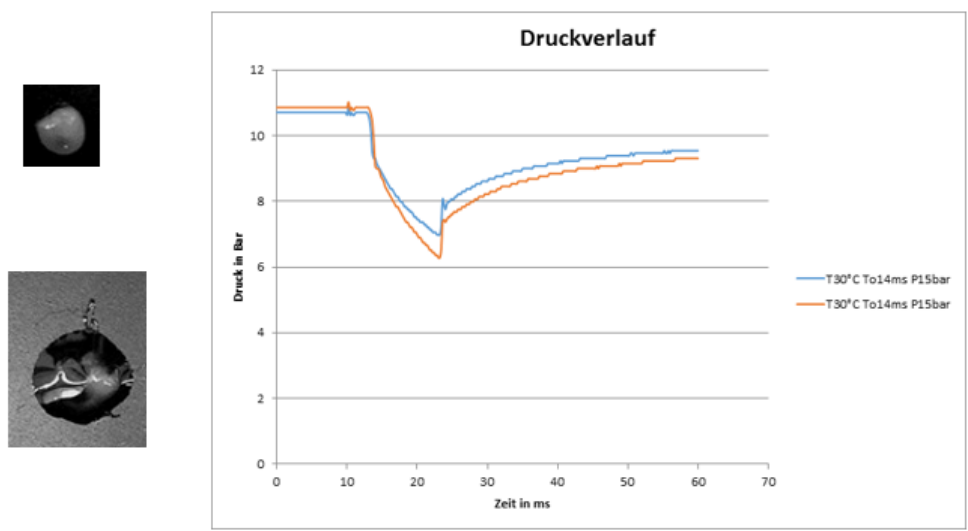

**Abbildung 1:** Vergleich von normalen Schmierfettpulspunkt (oben, blaue Kennlinie) mit Schmierfettpulspunkt nach Separierung von Bindemittel und Öl (unten, orange Kennlinie).

Anhand der Druckkennlinie ist die Separierung ebenfalls feststellbar. Separierte Pulspunkte weisen einen größeren Druckabfall auf.

## 2 Ausgangssituation

Die Ausgangssituation kann in der Abbildung 2 eingesehen werden. Die Fettförderpumpe fördert das Schmierfett unter Druck zu einem Material Verteiler. Dieser leitet das Schmierfett anschließend in einzelne Stränge die jeweils ein Ventil beherbergen. Jeder Fettstrang beinhaltet einen Materialdruckregler welcher Schwankungen in der Fettförderung ausgleicht und den Materialdruck auf den Betriebsdruck der Sprüh und Pulsventile herunter regelt.

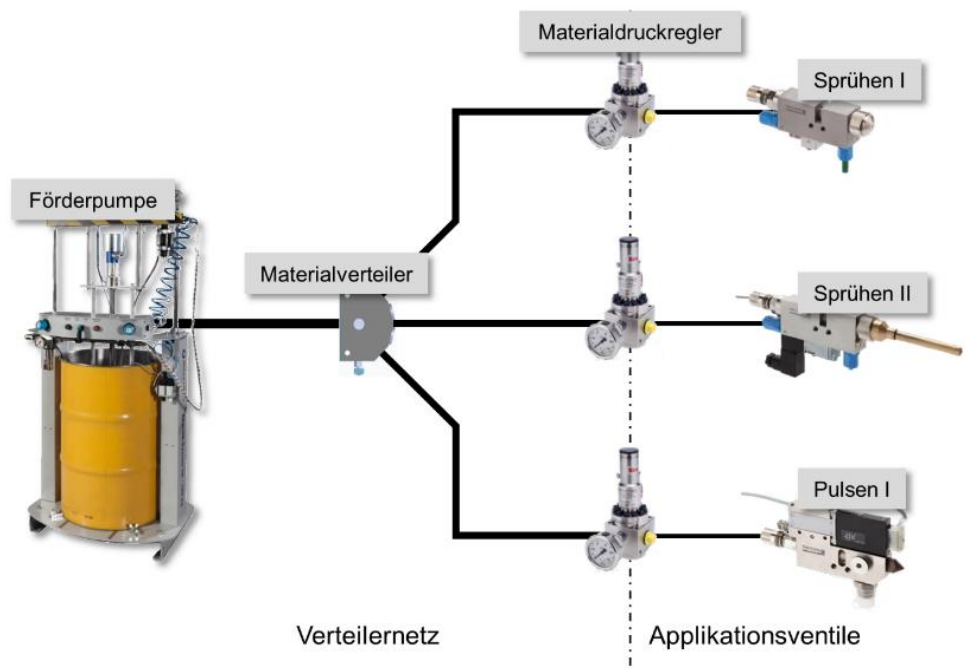

**Abbildung 2:** Fettförderpumpe mit Materialverteiler und Materialdruckregler zur Fettversorgung von Puls und Sprühventilen

Alle Komponenten dieses Aufbaus sind als Insellösungen konzipiert worden. Dabei erfolgt keine Kommunikation zwischen den Ventilen und den Druckreglern. Alle Bestandteile werden einmalig statisch aufgebaut, parametriert und anschließend betriebsbereit gesetzt. Etwaige Veränderungen wie z.B. Verwendung eines neuen Schmierfetttyps oder Änderungen von Schmierfettparametern durch Separation von Schmierfett und Öl bzw. durch Externe Einflüsse wie Temperatur werden nicht erfasst und müssen durch einen Mitarbeiter Vorort umgesetzt werden.

Die Applikation von Schmierfett durch Pulsventile erlaubt es Schmierfette im mg Bereich aufzutragen. Die Aufgetragenen Punkte müssen dabei sowohl in einem vordefinierten Toleranzbereich der Masse als auch eine geforderte Sollform erreichen. Die Pulspunktparameter lassen sich dabei durch folgende Stellgrößen beeinflussen:

- Mediumdruck (Variation durch Druckregler, mechanisch einstellbar)
- Ventilöffnungszeit (Zeit in dem das Ventil Material appliziert, elektronisch einstellbar)
- Spaltbreite (Breite des Ventilöffnungsspaltes, mechanisch einstellbar durch Ventilnadel)
- Mediumtemperatur (Temperatur einstellbar durch Heizpatrone, elektronisch einstellbar)

## 3    Projektziel

Die Ziele des Projektes werden in der Abbildung 3 dargestellt. Die bisherigen Insellösungen sollen dabei über Kommunikationsverbunde miteinander verknüpft werden. Ein Datenserver soll Prozessdaten aufnehmen und für späteres Training von adaptiven Algorithmen zur Verfügung stellen. Ein Kommunikationsverbund beschreibt dabei ein Ventil. Die Ventile werden dabei mit einer Kommunikations – und Regelungsplatine versorgt. Diese soll einen Kommunikationsaustausch zwischen Platine und einem neu entwickelten kommunikationsfähigen Druckregler gewährleisten. Dadurch wird das Ventil in der Lage sein den Druckregler selbst zu steuern. Zusätzlich dazu wird die Anbringung von Sensoren und Aktoren es ermöglichen direkt eine Regelung der Auftragsmasse am Ventil umzusetzen so, dass Parameterschwankungen im Schmierfett ausgeglichen werden können.

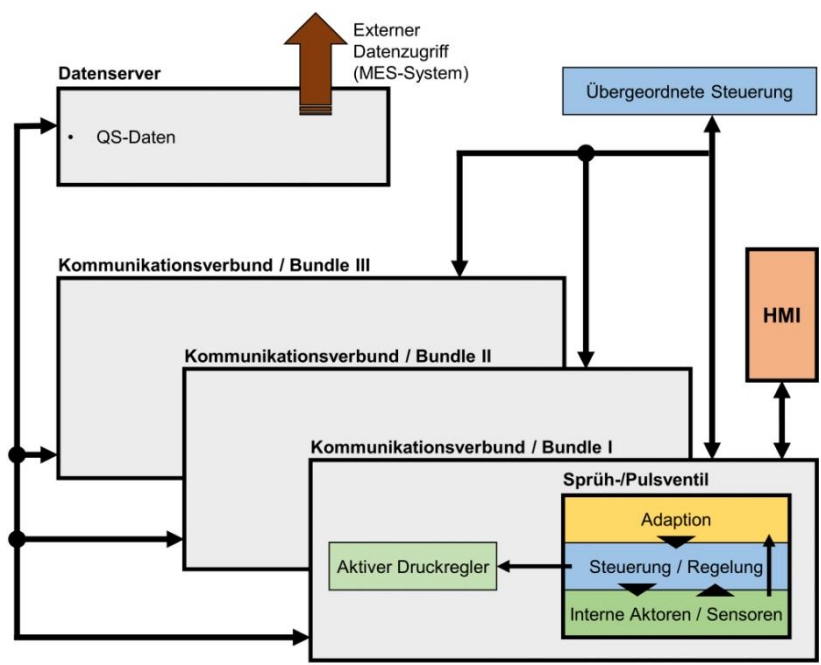

**Abbildung 3:** Darstellung des geplanten Verbundsystems

Da Schmierfette starken Parameterschwankungen unterliegen ist es notwendig einen Künstlichen Intelligenz (KI) basierten Algorithmus vorzusehen welcher zum einen die Möglichkeit hat für festgelegte Einstellparameter die Masse des Schmierfettpunktes sowie dessen Form zu ermitteln und zum anderen in der Lage ist für eine definiere Masse mögliche Einstellparameter aufzulisten. Zusätzlich soll der Algorithmus für neue Schmierfette trainierbar sein.

Zur Vereinfachung der Parametrierung der Ventile soll jedes Ventil mit einem Webserver ausgestattet werden. Dieser soll es ermöglichen Ventile per Fernzugriff zu steuern und statistische Daten des Ventils auszuwerten.

## 3.1 Parameterdefinition

In diesem Projekt werden Puls – und Sprühventile untersucht. Die zu untersuchenden Parameter wurden in 2 Kategorien unterteilt. Die erste Kategorie beinhaltet physikalische Eigenschaften des Pulspunktes. Die zweite Kategorie umfasst die möglichen Einstellparameter.

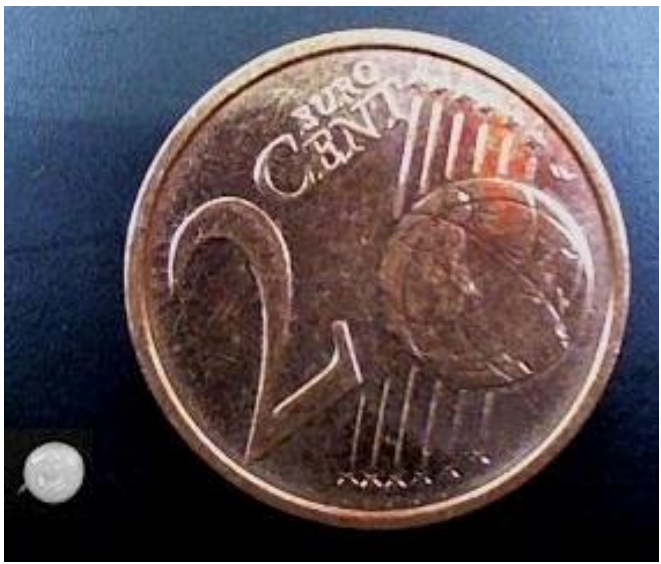

**Abbildung 4:** Darstellung eines Pulspunktes (links unten) im Vergleich zu einem 2cent Stück

Die folgende Liste zeigt alle Parameter auf:

- Schmierfettparameter
  - Auftragsform
    - Rund
    - Donutförmig (rund mit Loch in der Mitte)
  - Auftragsmasse: bis zu 5mg
  - Durchmesser: bis zu 5mm
- Einstellparameter
  - Ventilöffnungszeit: bis 14ms (abhängig von entstehender Auftragsmasse)
  - Ventilspaltbreite: bis zu 30 Rastereinstellungen and Drehregler
  - Eingestellter Druck von Materialdruckregler: 10 – 25Bar
  - Eingestellte Temperatur von Heizpatrone: 30 – 50°C

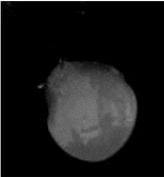 29.11.2019
B = 7,75mm
H = 7,92mm

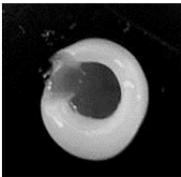 05.12.2019
B = 8.09mm
H = 9,82mm

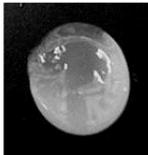 06.12.2019
B = 7,58mm
H = 8,43mm

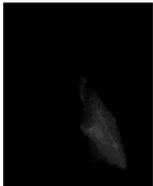 29.11.2019
T = 3,03mm
W = 0,055g

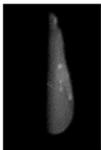 05.12.2019
T = 1,75mm
W = 0.052g

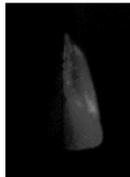 06.12.2019
T = 2,65mm
W = 0.054g

**Abbildung 5:** Abhängigkeit der Schmierfettpulspunkte mit gleichen Parametern gemessen an unterschiedlichen Tagen mit den gleichen Einstellparametern, Druckregler 10Bar, Öffnungszeit 20ms, Eingestellte Temperatur 40°C.

Die gültigen Bereiche der Einstellparameter wurden aus Erfahrungswerten der Experten des Kooperationspartners ermittelt. Sie legen die Grundlage zur Durchführung der Messreihen. Die Messungen selbst werden in den gültigen Bereichen der Schmierfettparameter durchgeführt.

## 4   Versuchsstand

Zur Erfassung der Messwerte musste ein Versuchsstand aufgebaut werden. Die Abbildung 6 zeigt dabei das Prinzip-schaubild des Versuchsstandes. Die pneumatisch betriebene Materialförderpumpe fördert Schmierfett aus einem Gebinde mit einem Druck von 50Bar. Das Schmierfett wird anschließend durch den Materialdruckregler in einem Bereich von 10 – 25Bar geregelt und zum Pulsventil weitergeleitet. Das Pulsventil wird pneumatisch angesteuert. Ein Industrieroboter wird verwendet um einen Prüfkörper aus einem Materiallager zu entnehmen. Die Pulspunkte werden auf den Prüfkörper aufgepulst. Anschließend wird in einer Präzisionswaage die Masse und unter einer Kamera die Form der Pulspunkte ermittelt. Die Druckkurve wird während der Pulsvorgänge ermittelt.

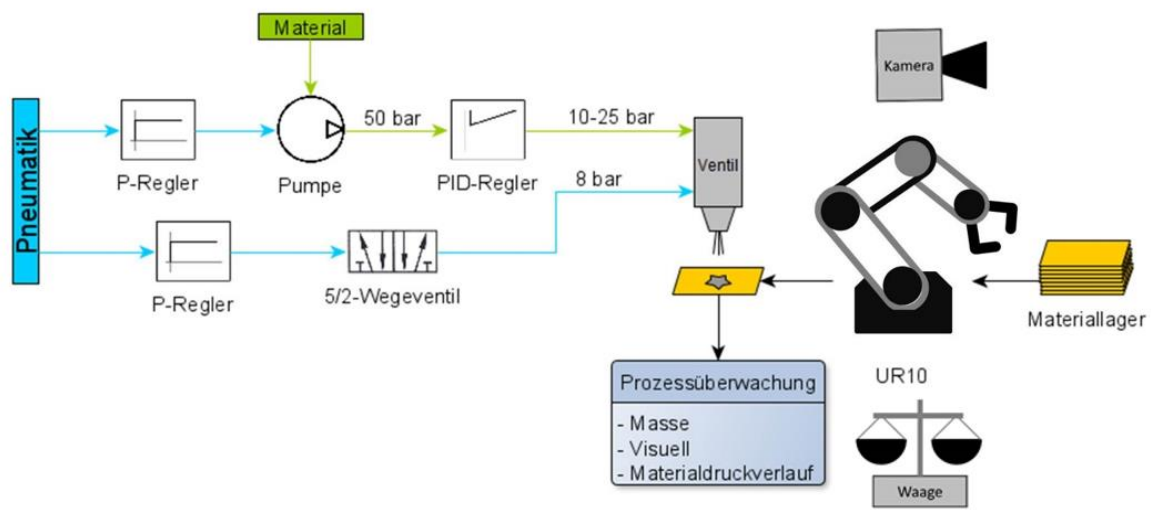

**Abbildung 6:** Prinzipschaubild des Versuchsstands

Der Ablauf eines Messvorgangs wird hier dargestellt:

1. Roboter entnimmt Prüfkörper aus Materiallager
2. Roboter positioniert Prüfkörper auf Präzisionswaage
   a. Leerer Prüfkörper wird in Präzisionswaage vorgewogen
3. Roboter positioniert Prüfkörper unter Ventil
   a. 15 Messpunkte werden versetzt auf Prüfkörper aufgepulst
4. Roboter positioniert Prüfkörper auf Präzisionswaage
   a. Voller Prüfkörper wird gewogen
   b. Mittelwert der Pulspunktmasse wird berechnet
5. Roboter positioniert Prüfkörper unter Kamera
   a. Pulspunktform wird aufgenommen
6. Roboter legt Prüfkörper ab

## 5   Feature Selektion

Um ein möglichst effektives Training zu gewährleisten ist es sinnvoll aus den gemessenen Kennlinien Kenngrößen zu extrahieren die den Verlauf der Kurve möglichst gut beschreiben. Diese Kenngrößen werden als Features bezeichnet. Die wichtigsten Features aus der Druckkurve sind dabei:

- Der Startdruck vor dem Pulsen
- Der minimal erreichte Druck der durch den Druckabfall des Pulsvorgangs entsteht
- Die Druckzeitkurve während des Pulsvorgangs
- Die Druckzeitkurve während des Regenerationsvorgangs (in einem definierten Zeitintervall)

Die Abbildung 7 zeigt eine qualitative Darstellung aus welchen Bereichen des Druckverlaufs die Features extrahiert werden. Die Regenerationszeitfläche ist dabei für alle gemessenen Kurven mit derselben Zeitdauer berechnet worden.

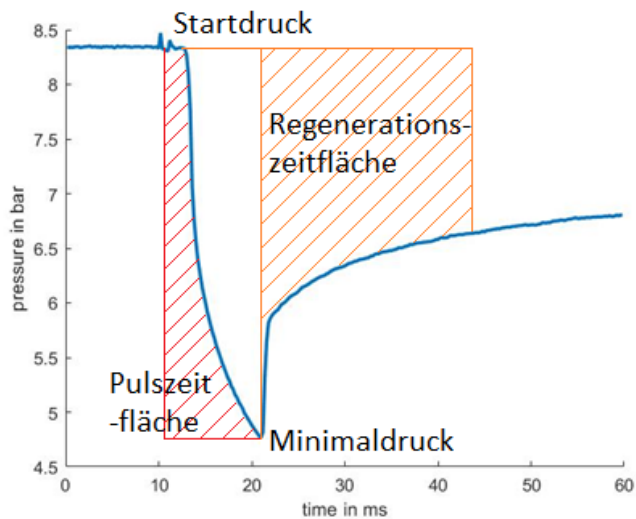

**Abbildung 7:** Qualitative Darstellung der wichtigsten Features anhand des Druckverlaufs

Diese Grundparameter werden aus jeder Messung extrahiert und dienen anschließend als Features zum Trainieren der KI – Algorithmen. Dabei werden die Features tabellarisch angeordnet. Der eigentliche Kurvenverlauf wird nicht zum Training verwendet.

# 6 KI Algorithmen

Welche KI Algorithmen zum Einsatz kommen hängt direkt von den Anforderungen ab. In der Analyse von Daten mit relativ wenig Features ist es meist nicht notwendig Deep Learning zu verwenden. Häufig reichen klassische Algorithmen des Machine Learnings bzw. nicht tiefe Neuronale Netze aus.

Im Fall von AdaptValve wurde bereits eine Kombination aus mehreren Algorithmen getestet. Dabei wurden Self Organizing Maps und K – Means Clustering zur automatische Clusterbildung verwendet. Dieser Gesamtalgorithmus erlaubt es für vorgegebenen Einstellparameter einen Massenbereich mit Standardabweichung vorherzusagen. Die umgekehrte Richtung kann ebenfalls über den Algorithmus umgesetzt werden. Der Algorithmus kann für einen vorgegebenen Massenbereich die möglichen Einstellparameter vorherzusagen um in den Massenbereich zu gelangen, siehe Abbildung 1.

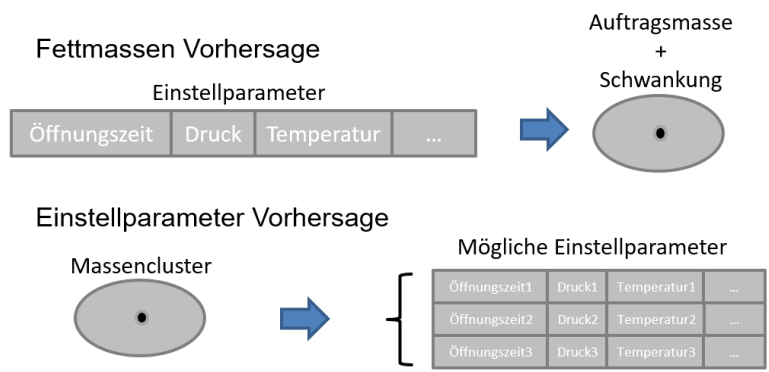

**Abbildung 8:** Vorhersagemöglichkeiten durch Gesamtalgorithmus

Der Gesamtalgorithmus wurde bereits im Workshop [1] im Zuge eines Vortrages genauer vorgestellt. Deshalb wird hier nicht genauer darauf eingegangen.

Die Kombination aus SOM und K – Means Clustering kann nur im trainierten Bereich zuverlässige Werte liefern. Außerhalb dieses Bereichs ist die Approximation nicht mehr gegeben. Dadurch kann der Algorithmus bei der Fettmassenvorhersage nur Werte liefern die in einem der trainierten Massencluster liegen. Sollte es jedoch gefordert sein eine kontinuierliche Funktion abzubilden ist ein anderer Algorithmus notwendig. Hierfür ist geplant das Verfahren der Bayesian Inferenz mit Markov Chain Monte Carlo zu verwenden. Dieses Verfahren ermöglicht es die Verteilungsdichtefunktion von Modellparametern zu approximieren. Wodurch nach dem Trainingsvorgang ein probabilistisches Modell zur Verfügung steht.

## 7 Resume

In dem Projekt „AdaptValve" sollen adaptive vernetzte Puls – und Sprühventile erstellt werden. Zur Erreichung dieses Ziels wurde ein Prüfstand entwickelt und aufgebaut. Mithilfe des Prüfstandes werden Messdaten erhoben. Die Messdaten werden aufbereitet und anschließend ausgewertet. Die finalen Messdaten werden als Trainingsdaten für KI Algorithmen verwendet. Dabei wurde bereits ein Algorithmus entwickelt. Dieser Algorithmus hat die Fähigkeit den zu erwartenden Massencluster für einen Einstellparameterbereich, aber auch die Möglichen Einstellparameter um in einen Massencluster vorherzusagen.

Eine Zukünftige Erweiterung sieht die Verwendung von Bayesian Inferenz vor um nicht nur Massencluster vorherzusagen, sondern um ein kontinuierliches probabilistisches Modell zur Massenvorhersage zu erhalten.

## 8 Infos zum Programm

Das Zentrale Innovationsprogramm Mittelstand (ZIM) des Bundesministeriums für Wirtschaft und Energie fördert technologie- und branchenoffen:

- Einzelprojekte
- Kooperationsprojekte
- Innovationsnetzwerke

sowie im Vorfeld Durchführbarkeitsstudien.

**Infos und Beratung zu Kooperationsprojekten**

AiF Projekt GmbH

Telefon 030 48163-451

www.zim.de

## Literaturverzeichnis

# Künstliche Intelligenz im 3D-Druck

*Lars Pfotzer[1], Maximilian Erschig[2], Martin Kipfmüller[2]*

[1]*Apium Additive Technologies GmbH*
*Karlsruhe, Deutschland*
*lars.pfotzer@apiumtec.com, +49 721 132095-0*

[2]*Hochschule Karlsruhe*
*Institute of Materials and Processes*
*Karlsruhe, Deutschland*
*maximilian.erschig@h-ka.de, +49 721 925-2064*
*martin.kipfmüller@h-ka.de, +49 721 925-1905*

***Zusammenfassung.*** *Das 3D-Druck-Verfahren Fused Filament Fabrication (FFF) beschreibt ein generatives Fertigungsverfahren mit weiter Verbreitung im Bereich komplexer Bauteile. Einige grundlegende Vorteile gegenüber konkurrierenden Herstellungsverfahren führen zu einer zunehmenden Verbreitung in der Technik und auch in der Serienproduktion. Bei einem Einsatz des Verfahrens in der Industrie sind je nach Anwendungsgebiet sehr hohe Genauigkeiten gefordert. Dabei gilt es Probleme, welche die Bauteilgenauigkeit mindern, wie beispielsweise das Warping (Bauteilverzug) oder Cracking (Rissbildung), zu vermeiden.*

*In einem aktuellen Forschungsprojekt zwischen dem Institute of Materials and Processes an der Hochschule Karlsruhe und der Apium Additive Technologies GmbH sollen die entstehenden Spannungen im Bauteil auf Basis einer Temperatursimulation berechnet werden. Anhand der Daten können bekannte Probleme durch ein frühzeitiges Erkennen, vor dem Druckprozess vermieden werden. Damit ist eine erhebliche Verbesserung der Wirtschaftlichkeit des Fertigungsprozesses möglich. Weiterhin sollen die Ergebnisse verwendet werden, um beispielsweise kritische Geometrien anhand künstlicher Intelligenz zu identifizieren. Somit lassen sich Probleme im Betrieb durch einen Algorithmus beheben, ohne dass eine Simulation notwendig ist.*

***Schlüsselworte.*** *3D-Druck, Simulation, Temperatur, künstliche Intelligenz*

## 1 Einleitung – 3D-Druck von Hochleistungspolymeren

Hochleistungspolymere spielen in der industriellen Fertigung eine zunehmend entscheidende Rolle. Durch Eigenschaften wie zum Beispiel hohe Temperaturbeständigkeit, gute Bioverträglichkeit, sowie Strahlungsresistenz eignen sich Thermoplaste wie beispielsweise Polyetheretherketon (PEEK) zum Einsatz in der Medizintechnik, der Automobilbranche, der Öl- und Gasindustrie oder der Luft- und Raumfahrt. Allerdings führen Herausforderungen wie die hohen Prozesstemperaturen (bis zu 520 °C) oder Probleme beim Druckvorgang, wie Layer Cracking (Schichtenhaftung) und Warping (Bauteilverzug), bei der hier betrachteten Methode der Fused Filament Fabrication, zu einer erschwerten Prozesskontrolle.

Bei dem Verfahren liegt der Kunststoff in Filamentform vor und wird schichtweise aufgetragen. Das aufgeschmolzene Filament kühlt ab, erhärtet dabei und die Wärme wird mittels Wärmeleitung teilweise in das Druckteil abgegeben. Dies führt zu erhöhtem Temperaturverzug und einer daraus resultierenden Spannungsbildung im Druckteil. Um im schlimmsten Fall ein Bauteilversagen zu verhindern und Kunststoffe wie PEEK dennoch erfolgreich zu drucken, müssen die Prozessparameter beherrscht und an den jeweiligen Anwendungsfall angepasst werden. Hierbei besteht eine Vielzahl an Einflussgrößen, welche sehr gute Prozesskenntnisse voraussetzen, um die Bauteile entsprechend der gestellten Anforderungen zu drucken.

## 2 Simulation im 3D-Druck

In der Forschung wird stetig versucht, die genannten Probleme, die während des Druckprozesses auftreten, zu minimieren oder gänzlich zu beseitigen. Eine vielversprechende Methode zur Abbildung und Optimierung des Herstellungsprozesses, ist die Anwendung der Finite-Elemente-Methode (FEM). Durch entsprechende Weiterentwicklung wurde die reine Strukturmechanik um dynamische, thermische und andere Anwendungsgebiete ausgedehnt. Im Bereich des 3D-Drucks wird die FEM für eine Simulation der Temperaturen und damit zusammenhängend der entstehenden Spannungen im Bauteil herangezogen. Unter Berücksichtigung der Druckparameter und des Verfahrwegs eines 3D-Druckers, sollen Erkenntnisse gesammelt werden, welche die Lösung noch immer existenter Probleme darstellen. Als Resultat sollen kritische Stellen, welche im Zuge des Abkühlvorgangs und den resultierenden Spannungen auftreten, bereits vor dem Druck detektiert werden und einen frühzeitigen Eingriff ermöglichen.

Der Nachteil beim Einsatz einer FEM Simulation im 3D-Druck, ist die zeitintensive Rechendauer, welche bei größeren Bauteilen rapide zunimmt. Im aktuellen Forschungsprojekt soll die Dauer für die Detektion der kritischen Stellen mit Hilfe von künstlicher Intelligenz drastisch verkürzt werden. Hierbei wird ein Algorithmus anhand der Simulationsergebnisse trainiert, um letztendlich die kritischen Stellen ohne vorherige Simulation zu erkennen. Weiterhin können durch ein entsprechendes Training Optimierungsvorschläge erstellt werden, welche beispielsweise eine Änderung der Geometrie, der Prozesstemperaturen oder der Druckstrategie berücksichtigen.

Um dem Endbenutzer die zu entwickelnde Strategie zur Verfügung zu stellen, wird eine Cloud-Plattform erstellt. Hierbei lassen sich die zu druckenden Bauteile, eine Abschätzung der Risiken sowie die Optimierungsvorschläge vor dem Druck visualisieren. Die Kommunikation zwischen Simulation, Slicing Software, KI-Algorithmus sowie der Online-Plattform ist in Abbildung 1 schematisch dargestellt.

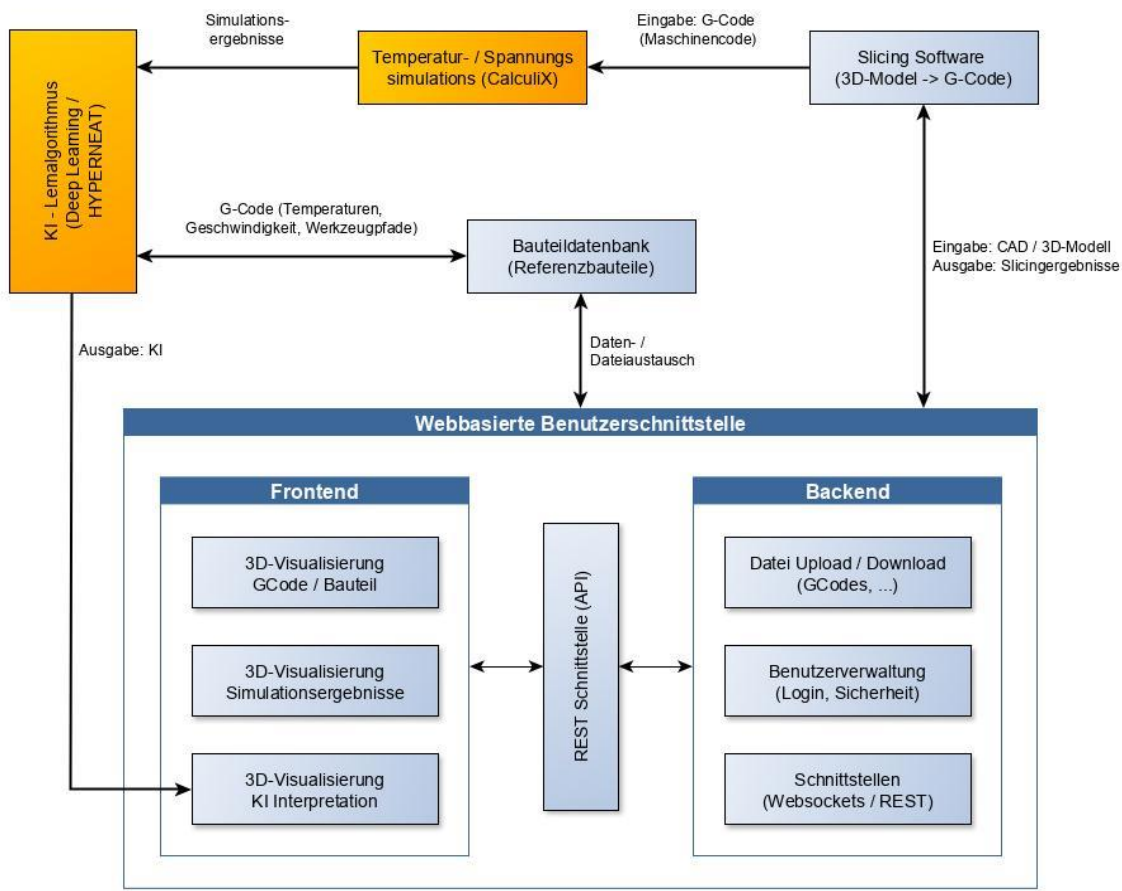

**Abbildung 1:** Schematischer Ablauf

# 3 Auswahl des KI Verfahrens

Bei der Auswahl eines geeigneten Verfahrens der künstlichen Intelligenz, sind die gegebenen Randbedingungen zu berücksichtigen. Da das Vorhaben durch ein hoch-dimensionales Problem mit großen Datenmengen sowie einer starken geometrischen Abhängigkeit definiert wird, werden die Möglichkeiten der genetischen Algorithmen herangezogen.

## 3.1 Genetische Algorithmen

Bei den genetischen Algorithmen wird zufällig eine Population erstellt und jedes enthaltene Element separat bewertet. Die Bewertung findet basierend auf dem Soll-Wert statt und wird als sogenannte Fitnessfunktion beschrieben. Die ausgewählten Elemente werden im Anschluss weiter ausgebildet, um mehr von ihnen zu produzieren. Dabei können Mutationen eingebaut werden, um bessere Individuen zu produzieren. Nach erfolgreicher Auswahl und dem Aussortieren der schlechteren Individuen, werden die Schritte erneut ausgeführt bis lediglich ein Element übrig bleibt. Dabei wird jede Iteration als Generation bezeichnet.

Der übliche Ablauf bei der Generierung von genetischen Algorithmen ist im Folgenden dargestellt:

1. **Initialisierung:** Population aus zufällig erzeugten Individuen
2. **Evaluation:** Bewertung der Güte einzelner Individuen (Fitnessfunktion)
3. **Iterative Optimierung** bis Abbruchkriterium erfüllt:
    1. **Selektion:** Auswahl von Individuen für Rekombination
    2. **Rekombination:** Kombination / Kreuzung der gewählten Individuen
    3. **Mutation:** Zufällige Veränderungen der Nachfahren
    4. **Evaluation:** Bewertung der Güte einzelner Individuen (Fitnessfunktion)
    5. **Selektion:** Bestimmung der neuen Generation

## 3.2 NEAT und HYPERNEAT

Eine auf der Idee der genetischen Algorithmen aufbauende Möglichkeit stellen die NEAT (Neuro Evolution of Augmenting Topologies) beziehungsweise HyperNEAT (A Hypercube-Based Indirect Encoding for Evolving Large-Scale Neural Networks) Algorithmen dar. Der NEAT Algorithmus setzt die Idee um, dass es am effektivsten ist, die Evolution mit kleinen, einfachen Netzwerken zu beginnen und diese über die Generationen hinweg immer komplexer werden zu lassen. Ähnlich der Organismen in der Natur, welche seit der ersten Zelle an Komplexität zugenommen haben, werden auch die neuronalen Netze in NEAT immer komplexer. Dieser Prozess der kontinuierlichen Ausarbeitung ermöglicht es, hoch entwickelte und komplexe neuronale Netzwerke zu finden. Der Vorteil der Struktur ist, dass jedes Mal, wenn ein Link hinzugefügt wird, der Algorithmus prüft, ob dieser Link bereits existiert hat. Abbildung 2 stellt eine beispielhafte Zuordnung von Genotyp zu Phänotyp dar. Hierbei gibt es drei Eingangsknoten, einen verborgenen sowie einen Ausgangsknoten. Diese sind über sieben Verbindungsdefinitionen, von welchen eine wiederkehrend ist, miteinander verknüpft.

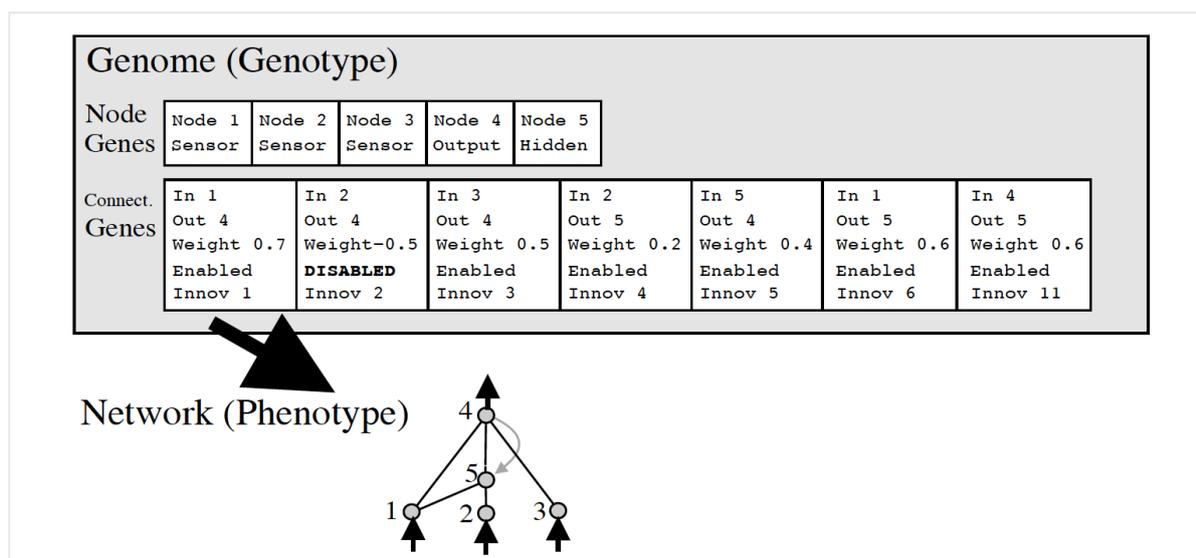

**Abbildung 2:** NEAT – Individuen Repräsentation (Encoding) [1]

Abbildung 3 stellt die Mutation beziehungsweise die Kreuzung von zwei Netzen dar. Hierbei können Knoten/ Kanten hinzugefügt oder entfernt sowie die Gewichte verändert werden.

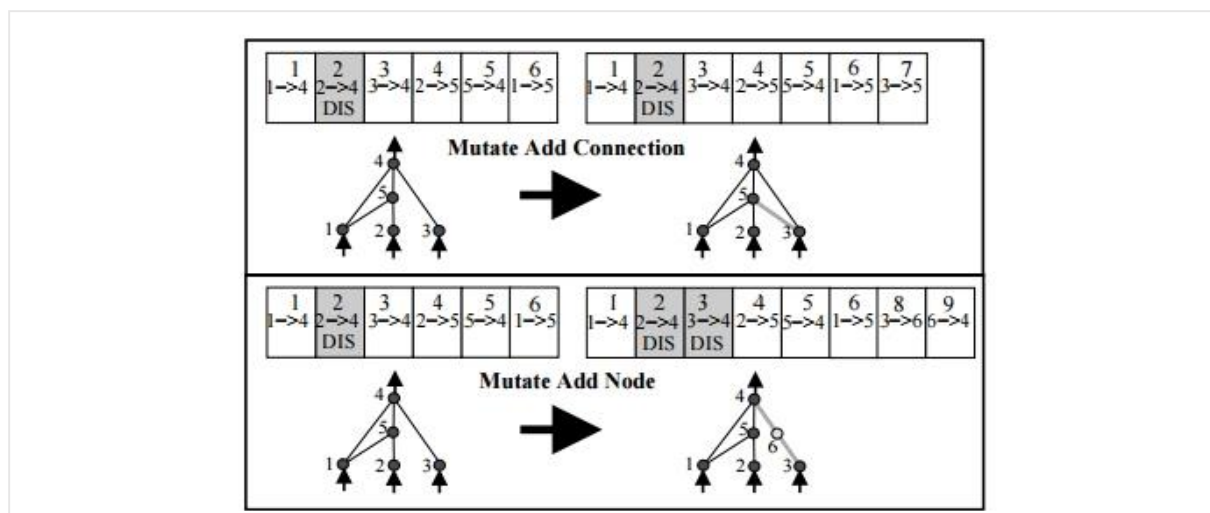

**Abbildung 3:** NEAT – Mutation / Kreuzung [1]

Bei der Artenbildung (Speciation) können Knoten/ Kanten hinzugefügt werden, ohne das eine Optimierung der Gewichte stattfindet. Dies liefert zunächst schlechtere Ergebnisse (Fitness) und würde bei der nächsten Selektion wieder entfernt. Hierbei besteht die Möglichkeit, die Population in Arten (Species) aufzuteilen. Dies geschieht anhand der Ähnlichkeit der Topologie und Kanten. Durch eine Markierung der Historie (Innovation Number) ist der Vorgang messbar. Bei der Selektion (Auswahl von Individuen für Rekombination) konkurrieren nur die Individuen innerhalb einer Art was es ermöglicht, neue Strukturen zu explorieren bevor sie gänzlich beseitigt werden.

Der HyperNEAT Algorithmus erweitert die NEAT Methode. Der Name ergibt sich aus dem Wort Hypercube, da ein CPPN (Compositional Pattern-Producing Network), welches ein Konnektivitätsmuster beschreibt, mindestens vierdimensional ist und sich somit mehrdimensionale Räume ergeben. Diese werden für gewöhnlich innerhalb der Grenzen eines Hyperwürfels abgetastet (meist von -1 bis 1). Dabei stellt jeder Punkt innerhalb des Würfels ein Verbindungsgewicht dar. Der Algorithmus "malt" somit ein Muster auf die Innenseite der Würfel, welches als Verbindungsmuster eines neuronalen Netzes interpretiert wird.

Diese speziellen Methoden der evolutionären Algorithmen bieten gewisse Vorteile bei z.B. Problemen mit einer großen Anzahl an Ein- bzw. Ausgängen, bei Problemen mit variabler Auflösung oder bei Problemen mit geometrischen Beziehungen zwischen den Ein- und/oder Ausgängen.

## 4  Fazit

Die aufwendigen Berechnungen zur korrekten Prognose des Verformungsverhaltens von additiv gefertigten Bauteilen, sind nicht geeignet, um sie bei der Optimierung der Fertigung von Bauteilen einzusetzen. Deswegen wird in beschriebenem Projekt ein Ansatz entwickelt, der anwendbare Ergebnisse mit wesentlich weniger Rechenaufwand in kürzerer Zeit liefert. Aktuell wird hierzu ein auf der Synthese von genetischen Algorithmen und KI (NEAT bzw. HyperNEAT) basierender Ansatz entwickelt.

## 5  Danksagung

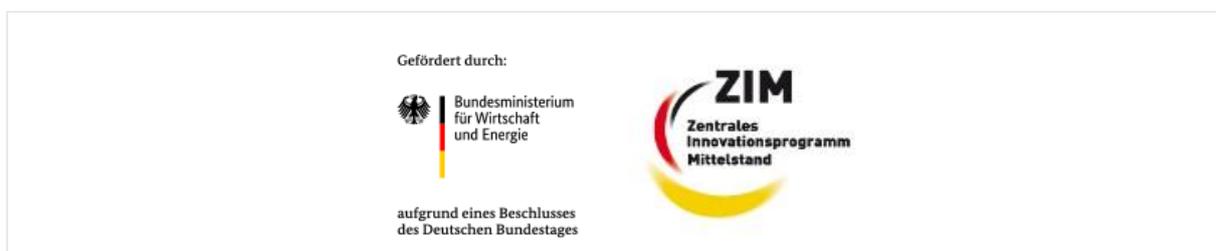

## Literaturverzeichnis

# Die Beziehung Mensch - Maschine als Teil der Ethischen Intelligenz

*Michael M. Roth*

*MicialMedia*
*Karlsruhe, Deutschland*
*info@michael-roth.de, +49 174 6343 981*

**Zusammenfassung.** *Gegenstand des Papers und des mit ihm assoziierten Vortrages an der Hochschule Karlsruhe beim Online-Kongress „KI für den Mittelstand" zum Thema „Die Beziehung Mensch - Maschine als Teil der Ethischen Intelligenz" ist im Kern die vom Autor begründete kleine Philosophie der Ethischen Intelligenz. Die Philosophie oder das Prinzip ist eine Art generelles Paradigma, das sich interdisziplinär als ein universeller Denkansatz zeigt. Es geht darum, einen Blick zu werfen auf die mannigfaltigen Beziehungen zwischen dem Menschen und der Natur im weitesten Sinne. Was finden wir vor, und wo wollen wir hin? Unter die Lupe genommen, wird dabei auch insbesondere das Verhältnis von Mensch zu Maschine. Wird sich das uralte Paradigma halten können, dass die Maschine das Werkzeug des Menschen sei? Der neue Begriff der „Soziogenten" soll einen Raum schaffen für ein neues Verständnis im Umgang mit unserer Umwelt. Während Soziogenten alle Wesen auf den Planeten sind, die sozial interagieren, bezeichne ich KI-Systeme der Zukunft als „Künstlich-intelligente Soziogenten" (Kisos), die letzten Endes dem Menschen ebenbürtig sein werden. Der alte Ansatz, dass das intelligenteste Wesen auf dem Planeten, aka. Mensch, über allem steht, das Recht zur uneingeschränkten Ausbeutung der anderen Soziogenten hat, soll dem Ansatz eines fairen Miteinanders weichen. Wenn wir heute darauf verzichten, Schweine in zu enge Kastenstände zu sperren und männliche Küken massenhaft und maschinell zu töten, dann haben wir gute Chancen, eines Tages von Kisos ebenso rücksichtsvoll behandelt zu werden, selbst wenn sie uns dann intellektuell haushoch überlegen sein werden.*

**Schlüsselworte.** *Ethische Intelligenz, Soziogenten, Intellekt, Ethellekt, Mensch, Maschine, Neue Ethik 2020, Neue Intelligenz 2020, Künstliche Intelligenz*

## 1 Einleitung

Als Teil einer Informatikerfamilie beschäftigte ich mich schon frühzeitig mit Themen wie Revolutionierung der Gesellschaft durch die Einführung von „digitalen Automaten". Persönliche Begegnungen mit dem „Vater der Künstlichen Intelligenz", Joseph Weizenbaum, ließen ethische Fragen im Kontext einer rasanten Technologieentwicklung für mich immer als relevant erscheinen. Bereits während meiner Ausbildung zum Elektronikfacharbeiter in den 1980er Jahren klebte ich einen Zettel an eine Wandzeitung mit dem kurzen, aber für die damalige Zeit provokativen Text: „Ich glaube nicht, dass auf ewig dem Menschen allein die Eigenschaft von Intelligenz und die Fähigkeit zu bewussten, autarken Handeln vorbehalten sein wird!", womit ich im Klassenkollektiv eine intensive Diskussion inspirierte.

Die philosophischen Themen haben sich bis heute kaum verändert. Nur dass wir heute, zu Beginn des 21. Jahrhunderts, freilich technologisch ein gutes Stück vorangekommen sind. Die unaufhörliche Technologieentwicklung scheint eine Art Selbstläufer zu sein. Es könnte keine Hundert Jahre mehr dauern, und der Mensch hat sich eine neuartige Spezies geschaffen, die ihn selbst ablöst von der Rolle der über Jahrtausende gewohnten Intelligenzführerschaft. Der Paradigmenwechsel könnte je bitterer verlaufen, desto mehr sich der Mensch an seiner Entscheidungs- und Handlungsdominanz festzuklammern versucht. Typische Vertreter sind Philosophen der Jetztzeit. Julian Nida-Rümelin sagt im Vorwort zu seinem Buch „Digitaler Humanismus": „Digitale Computer sind algorithmische Maschinen" [1]. Richard David Precht in seinem aktuellen Buch „KI und der Sinn des Lebens": „Allein Menschen sind zu Emotionen und Moral fähig" [2]. Beide Aussagen reduzieren zukünftige KI-Systeme, die ich als Kisos,

Künstlich-intelligente Soziogenten, bezeichne, als maschinelle Befehlsempfänger und Ausführungsorgane des Menschen. Precht zitiert gerne den französischen Philosophen René Descartes, „Cogito ergo sum", immerhin rund 400 Jahre zurück liegend. Doch dass Selbiges ein Kiso in (evolutionär fast vernachlässigbaren) 100 oder 200 Jahren ebenso sagen könnte, das sieht Precht nicht [3]. Schade. Denn in Anbetracht der aktuell rasanten Technologieentwicklung erscheint es mir fast wie auf einem Tablett der Geschichte serviert, dass Intelligenz, ja Ich-Bewusstsein, nicht auf ewig dem Menschen oder biologisches Trägermaterial überhaupt vorbehalten sein wird. Das nach dem Visionär Gordon Moore benannte Mooresche Gesetz wurde bereits 1965 formuliert. Es sagt grob eine Verdoppelung der Anzahl von Schaltelementen (Transistoren) bei Integrierten Schaltkreisen (statistisch) alle 18 Monate voraus. Verblüffend, dass im Groben das Gesetz bis heute Gültigkeit hat. CPUs von Computern haben heute bereits über 10 Milliarden Transistoren. Die Anzahl der Neuronen im menschlichen Gehirn ist derzeit um die Faktor 10, nämlich 100 Milliarden. Aber die Technologieentwicklung geht weiter.

Der Ministerpräsident von Baden-Württemberg kürzlich beim Get Started Bitkom Gründerfrühstück „Wir müssen vom verhindernden zum gestaltenden Datenschutz kommen!" [4]. – Für mich eine fundamentale Aussage. Denn vor lauter Datenschutz, angeblich nicht vorhandener Daten (das Universum ist voll von Daten!) versäumen wir es gegenwärtig, die KI mit uns zu nehmen. IBM-Vorsitzender Thomas J. Watson war 1943 der Meinung, dass es einen Weltmarkt für gerade mal fünf Computer geben würde [sic!]. Der Wirtschaftsredakteur der Neuen Züricher Zeitung, Stefan Häberli, aktualisiert den Irrtum im Jahr 2020: „Geeignete Anwendungsfelder für Deep Learning sind relativ selten." [5]. So wie heute digitale Schaltkreise ihren alltäglichen Gebrauch in PCs finden, so erwartet uns in den kommenden Dekaden ein regelrechtes KI-Universum. Tatsächlich spricht mir der KI-Professor Dominikus Heckmann aus der Seele, wenn er sagt, dass KI viel schlauer sein wird als wir Menschen [6].

Meine Philosophie der Ethischen Intelligenz soll dabei helfen, den Schritt in das KI-Zeitalter *bewusst* zu gehen. Wir Menschen sollten nicht nur unser Verhältnis zu zukünftigen Kisos überdenken, sondern zu allen Soziogenten auf dem Planeten überhaupt. Denn genau so wird KI eines Tages uns Menschen betrachten. Als einen Soziogent unter vielen, wenn auch ein gar nicht mal so unintelligenter. Die beiden Säulen der Neuen Ethik 2020 sowie der Neuen Intelligenz 2020 sind als neue Werkzeuge der Menschheit zu betrachten. Unsere Beziehungen sind unter ethischen Aspekten zu hinterfragen und Intelligenz muss neu gedacht werden. Ein bisschen pointiert gesprochen, befinden wir uns jeweils in einer Art Urozean. Es ist der Beginn eines neuen Zeitalters. Hierbei möchte und sollte der Mensch nicht unter die Räder kommen. Doch von seinem alleinigen Führungsanspruch auf dem Planeten Erde wird er sich verabschieden müssen. Auf der einen Seite etwas aufgeben, um auf der anderen Seite etwas Neues zu gewinnen. Im Idealfall könnte das eine konstruktive Partnerschaft werden mit den Kisos des 22. und der fortlaufenden Jahrhunderte.

## 2 Hauptteil

Die Ethische Intelligenz ist ein multiple interpretierbarer Begriff. Z. B. kann man Soziogenten, also Teilnehmende, die in der Gesellschaft (inter)agieren, wie die Menschen, als Ethische Intelligenzen bezeichnen. Insofern sie ethisch, intelligent bzw. ethisch-intelligent agieren. Man kann die Ethische Intelligenz (EI), oder kurz Ethelligenz, auch definieren als ein Rolemodel mit Vorbildcharakter. Für mich ist die EI ein Prototyp in der Gesellschaft, dem sich alle Soziogenten annähern könnten oder sollten. Nicht zuletzt sehe ich insbesondere in den kommenden Dekaden einen Wettbewerb zwischen Kisos und Misos, also Künstlich-intelligenten Soziogenten und Menschlich-intelligenten Soziogenten, um diesem Prototyp der Ethischen Intelligenz nahe zu kommen. Wir werden insgesamt allerdings nicht auskommen ohne Incentives, sei es für den Menschen oder sei es für die Kisos. Idealerweise wird es uns gelingen, die Ethische Intelligenz sowohl als logisch folgerichtig als auch als emotional erstrebenswert zu definieren und zu etablieren.

Als philosophisches Konstrukt, ein Prinzip oder Ansatz betrachtet, besteht die Ethische Intelligenz aus zwei Säulen. Auf der einen Seite die Neue Ethik 2020. Auf der anderen die Neue Intelligenz 2020. Die Neue Ethik 2020 ist zwar nicht entstanden oder erwachsen aus der Neuen Ethik, aber beide sind sozusagen modular koppelbar. Die Neue Ethik gab es schon 120 Jahre früher, nämlich um 1900. Die Frauenrechtlerinnen Helene Stöcker und Maria Lischnewska begründeten diese in Anlehnung Friedrichs Nietzsches „Umwertung der Werte". Es ging darum, allen Menschen gleiche Chancen für Entwicklung und Entfaltung zu gewähren. Selbstredend betraf das zur damaligen Zeit insbesondere auch die Frauenrechte; wenn man bedenkt, dass es erst 1919 die Frauen selbst schafften, ihr Wahlrecht durchzusetzen.

Natürlich betrachtet die Neue Ethik 2020 die Neue Ethik als einen willkommenen Baustein. Die Neue Ethik 2020 versteht sich darüber hinaus gehend als noch universeller. Insbesondere werden die zehn Ms der Beziehungen betrachtet, also die folgenden fünf Beziehungspaare: Mensch-Mensch, Mensch-Milieu (Natur), Mensch-Mikrokosmos, Mensch-Makrokosmos und Mensch-Maschine. Nur ganz kurz möchte ich anreißen, was sich hinter den jeweiligen Beziehungsebenen verbirgt.

Mensch-Mensch
Eigentlich ist das unsere Aufgabe Nummer eins. Hier müssen wir anfangen. Während wir natürlich gleichzeitig Augen, Ohren, vor allem Gedanken und folgend Handeln offen halten müssen für die weiteren Beziehungsebenen. Aber beginnen wir hier. Was ist mit Kriegen? Um Ressourcen, wegen Religionen, wegen sonstigen Streitigkeiten. Menschen werden abgeschlachtet. Was ist mit der Todesstrafe? Wofür brauchen wir sowas?

Mensch-Milieu
Der Mensch als Teil der Natur. Fridays for Future. Greta Thunberg. Peter Wohlleben. Dirk Steffens.

Mensch-Mikrokosmos
Corona. Immunsystem. Gesundes Leben. Heilung. Vorbeugung.

Mensch-Makrokosmos.
Nikolai Kardaschow. Mondlandung. ISS. Mars. Exoplaneten. Neubesiedlungen.

Mensch-Maschine.
Eine Beziehungsebene, die für die Selbstbestimmung und das Selbstverständnis der Menschheit in den kommenden Dekaden von ganz zentraler Bedeutung sein wird.

Auf die Beziehung Mensch-Maschine möchte ich näher eingehen. Denn diese Relation bietet den philosophischen Echoraum für das Thema des Kongresses „KI für den Mittelstand" #ki4industry der Hochschule Karlsruhe. Gerade hier kommt die Säule der Neuen Ethik 2020 der Säule der Neuen Intelligenz 2020 nahe. Hier müssen Ethik und Intelligenz zusammengedacht werden. Doch zunächst fokussiere ich auf die ethischen Aspekte. Während Richard David Precht es als eine Art Entfremdung betrachtet, auf KI-Systeme der Zukunft (Kisos) ethische Kompetenzen zu übertragen, halte ich genau dies für unabdingbar. In Bezug auf die Erdgeschichte und speziell die Menschheitsgeschichte wird es schon sehr bald zu einer Herausbildung Künstlich-intelligenter Soziogenten kommen. Sie werden sich zu einer neuen Spezies entwickeln, einer Art Kisoheit. Der Zeitraum ist dabei gar nicht der Punkt. Wobei es aus Sicht der Menschheit sogar besser sein könnte, wenn es nicht mehr 50 oder 100, sondern 200 Jahre bis dahin wären. Wenn wir uns überlegen, wie lange wir Menschen derzeit brauchen, um überhaupt Schweine aus ihrem viel, viel zu engen Kastenstände zu befreien. Da werden Gesetze gemacht, die auf fünf oder acht Jahre ausgelegt sind. Es ist, ganz einfach gesagt, Tierquälerei. Und warum? Weil wir (noch!) die Intelligentesten auf dem Planeten sind. Wir dürfen das. Dürfen wir das wirklich? Vielleicht sind es schon unsere Kindeskinder, die zur selben Zeit und an selben Ort mit Kisos leben, die sich fragen, was *sie* dürfen und was nicht. Was sie *wollen*. Für mich steht fest, dass wir innerhalb der kommenden Dekaden in der Gesellschaft Regularien aushandeln müssen, die auf alle Soziogenten anwendbar sind. Das bedeutet auch, dass wir nicht sinnlos oder aus Lust und Laune heraus Bäume fällen werden. Keine Schweine dicht an dicht packen, um sie maximal effizient zu züchten und dann, sogar gesundheitsschädigend für uns, massenhaft konsumieren. So wie der Mensch derzeit handelt, das ist ziemlich kurz gedacht. Nicht „nur", dass er im Missverhältnis zur Natur lebt, er ist auch dabei, eine Technologie zu erschaffen, für die er nicht annähernd ethische Maßstäbe konzipiert hat. Wie könnte er, wenn er in sich als Mensch nicht mal ethisch sinnvoll lebt und handelt? Es ist ein Trugschluss von Nathalie Weidenfeld, dass es sich bei den zahlreichen Science-Fiction-Filmen allein um die falschen Metaphern handeln würde, die sowieso nie eintreten werden, da einzig der Mensch ... [1]. Aber nein. „Einzig der Mensch" ... mit seinen einzigartigen Begabungen und Fähigkeiten. Diese Einzigartigkeiten werden wir spätestens in wenigen Jahrhunderten, und das ist hoch gegriffen, nicht mehr einzig besitzen. Unsere Gesellschaft muss multilateral und soziogenten-offen werden, soll sie noch ein paar Jahrtausende, gerne sogar noch länger, eine Überlebenschance haben.

Die Vorbildrolle ist das Wenigste, das Mindeste, was wir Menschen an die Kisos der kommenden Jahrzehnte und Jahrhunderte vererben können. Ethische Intelligenz wird der Minimalkonsens zwischen den Spezies sein. Anderenfalls dominiert die eine Spezies alle anderen. So wie wir Menschen das bisher gemacht haben auf der Erde. Mit dem Unterschied, dass dann nicht mehr der Mensch an der Spitze der Sozigenten-Pyramide stehen wird.

Die Neue Intelligenz 2020 besitzt einige Attribute, die wir heute schon im Ansatz kennen, in der Realität jedoch viel zu wenig ausleben und umsetzen. Das Autonome Fahren könnte eine erste Anwendung werden, die Neue Intelligenz Wirklichkeit werden lässt. Es gibt um Vernetzung, um assoziatives Denken. Und das alles innerhalb von Sekundenbruchteilen. Schauen wir uns Anwendungen wie Facebook an. Heutzutage ist es nicht mal mehr möglich, dass ich einstellen kann, dass alle Teilnehmenden auf Facebook sehen, wenn ich an eine Veranstaltung teilnehme. Nur Freunden und Freundinnen ist dies gestattet. Es ist das Gegenteil von Vernetzung, wie sie einst von Facebook selbst initiiert wurde. Schauen wir uns die Coronapandemie an. Wie viele Vorhersagen wären möglich gewesen, wenn wir frühzeitig KI-Expertensysteme eingesetzt hätten? Inzwischen tauchen hier und da vereinzelt solche Systeme auf. Obwohl wir technologisch schon so weit fortgeschritten sind, sind wir auf der anderen Seite noch weit entfernt von einer wirklich interaktiven, intelligenten, stark dialogorientierten Wissensgesellschaft. Das bedeutet, dass Fehler sich immer wieder wiederholen. Lerneffekte finden vielleicht regional statt. Aber aus Sicht der Menschheit ist das noch zu unbedeutend. Dabei kann die Menschheit schon einigermaßen stolz auf sich sein, dass sie so etwas wie die Internationale Raumstation (ISS) im Orbit hat. In vielerlei Hinsicht ist dieses Projekt Gold wert. Vor allem sieht man hier, wozu die Menschheit in der Lage ist, wenn man grenzüberschreitende und kulturübergreifende Zusammenarbeit ermöglicht. Unterschiedliche Sprachen, verschiedene Herkünfte, alles kommt bei der ISS auf engstem Raum zusammen. Und es funktioniert! Neue Intelligenz könnte bspw. auch bedeuten, dass das Smartphone, in das wir unsere Texte eingeben, unsere Texte versteht. Korrekturen von Tippfehlern sind bisher allenfalls auf Ebene der Worte oder Wortgruppen möglich. Eine schwache Syntaxkontrolle. Grammatik in Spuren und Semantik schon gar nicht. Ja, die Sorge, dass Smartphones oder PCs eines Tages tatsächlich verstehen und begreifen, was wir ihnen mitteilen, ist nicht unberechtigt. Spätestens hier müssen Ethik und Intelligenz wieder zusammengedacht und zusammengeführt werden. Ebenso brauchen wir keine Mogelpackungen mehr von Rasierklingenherstellern, die eine Packung mit acht möglichen Plätzen nur mit fünf Rasierklingen bestücken. Solche Ansätze sind heute nur noch anachronistisch und nicht wirklich ethisch-intelligent.

In der Zukunft werden wir verschiedene Optionen haben. Wir schreiben Programme, entwickeln Produkte und bieten Dienstleistungen an, die sich noch mehr an den Bedürfnissen des Menschen orientieren. Dabei gewinnen Unternehmer, Unternehmerinnen und Unternehmen als Ganzes, *und* Verbraucher und Verbraucherinnen. Oder aber wir versuchen KI dafür zu nutzen, um Kunden und Kundinnen übers Ohr zu hauen. Meine Bitte an die Entwickler und Entwicklerinnen mittelständischer Unternehmen: Setzt KI für die Menschen ein. Kümmert Euch um eine bessere Balance Mensch-Umwelt, was alle Soziogenten inkludieren sollte, die sich ebenso ethisch-intelligent verhalten. Mit Mogelpackungen, die eigentlich ein Relikt aus dem letzten Jahrhundert sein sollten, erreicht man vielleicht kurzfristige Gewinne, aber nicht die Menschen. Auch wenn Kisos, also KI-Systeme der Zukunft, an Bedeutung gewinnen sollten, so müssen wir uns selbst so viel Wert sein. Respekt fängt bei uns Menschen an. Gegenseitiger!

In einem Artikel auf meiner Website hatte ich den Ethische-Intelligenz-Test eingeführt, kurz Roth-Test. [7] Der Roth-Test stellt eine Extrapolation und Weiterführung des bekannten Turing-Tests dar. Während es beim Turing-Test darum geht, durch geschickt gewählte Fragestellungen in Abhängigkeit von den Antworten zwischen einem Menschen und einer Maschine zu unterscheiden, vernachlässigt mein Test die Frage, ob es sich um einen Menschen oder eine Maschine handelt. Hier steht die Ethische Intelligenz bzw. die Fähigkeit zu ethisch-intelligentem Denken und Handeln im Vordergrund. Somit könnte es sich sowohl um einen Menschen oder um einen Computer handeln, also auch ein Kiso sein. Vielleicht sogar ein Tier oder eine Pflanze. Beide würde man zwar nicht befragen, aber anhand ihrer Handlungen beurteilen können. Nach der Befragung oder der Beurteilung steht die Erkenntnis, ob es sich um einen ethisch-intelligenten Soziogenten, also einen Teilnehmer, einer Teilnehmerin an der Gesellschaft mit oder ohne Ethische Intelligenz handelt. Richard David Precht sagte, er möchte niemals eine Maschine über sein Schicksal entscheiden lassen. Ich sage, ich möchte niemals einen Soziogenten über mich entscheiden lassen, der nicht über Ethische Intelligenz verfügt. Aufgrund der Verschiebungen des Gewichts der Teilnehmenden in der Gesellschaft könnte es also auch zu einer Bedeutungsverschiebung derartiger Tests kommen.

**Turing-Test vs. Roth-Test**

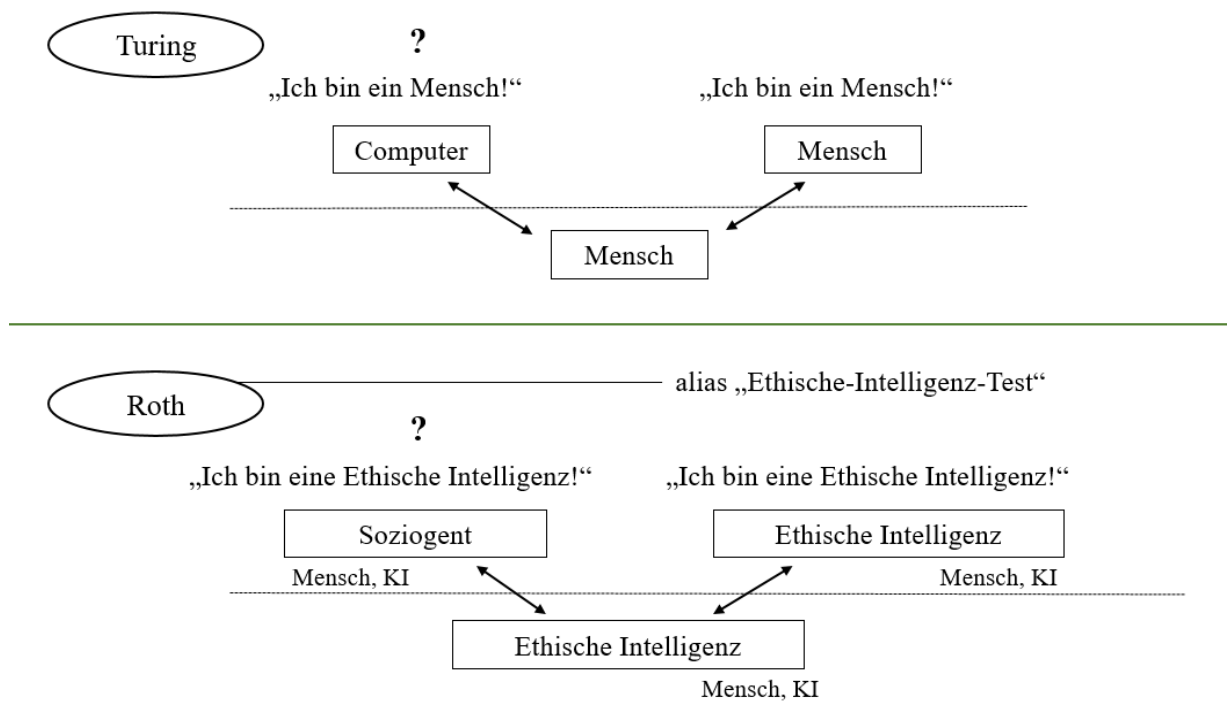

Testvariationen: - observativ (1)

- interaktiv (2)

- ganzheitlich (1+2)

Im erweiterten Testverfahren kann der Soziogent auch Tier oder Pflanze sein.

Die Grafik zeigt den Roth-Test als Weiterführung des Turing-Tests. © Michael M. Roth

## 3    Fazit

Zusammenfassend sei gesagt, dass sich die Ethische Intelligenz als eine Universaldisziplin oder ein universell anwendbares Prinzip versteht. Zu untersuchen sind die Beziehungen des Menschen zu seiner Umwelt, insbesondere zu allen Soziogenten, also Wesen, seien sie biologisch oder technischer Natur. Die Minimalforderung dabei heißt „Modus vivendi". Die eine Spezies kann und darf nicht mehr agieren bei radikalem Selbstvorteil und radikalem Nachteil der anderen. Ein fairer Umgang miteinander sollte immer erstrebenswerter sein. Am Ende dient damit jede Spezies der Erhaltung und dem Wohlergehen der eigenen Art. Wenn wir neben dem Modus vivendi sogar Partnerschaften und Kooperationen erreichen, dann kultivieren Gleichberechtigung und das Fortkommen der Arten. Künstliche Intelligenz wird uns dabei unterstützen können. Anwendungsmöglichkeiten für den Mittelstand, für die Industrie und die Gesellschaft überhaupt, dürfte es genügend geben. Gleichwohl bedürfen wir der Leuchtturmprojekte, die uns zeigen, was möglich ist, und wohin die Entwicklung gehen kann. Neben intelligenten Lösungen sind ethisch vertretbare gesucht. Die Menschheit befindet sich an noch zu vielen Stellen und Orten der Welt im Kriege mit sich selbst. Es handelt sich hier um Zeit- und Ressourcenverschwendung. Weder das eine noch das andere können wir uns leisten. Denn je weiter die Technologien fortschreiten, desto mehr werden Intellekt und Ethellekt des Menschen gefragt und gefordert sein. Es liegt an uns selbst, ob wir unserem Fortbestand und dem Leben auf dem Planeten Erde eine Chance geben.

## Literaturverzeichnis

# KI-gestützte Analyse von industriellen CT-Daten

*Tim Schanz, Robin Tenscher-Phlipp, Martin Simon*

*Hochschule Karlsruhe*
*Fakultät für Maschinenbau und Mechatronik*
*Karlsruhe, Deutschland*
*tim.schanz@h-ka.de, +49 721 925-1719*
*robin.tenscher-philipp@h-ka.de, +49 721 925-1721*
*martin.simon@h-ka.de, +49 721 925-1720*

**Zusammenfassung.** *Die industrielle Computertomographie ermöglicht als bildgebendes 3D-Digitalisierungsverfahren umfassende Analysen industrieller Bauteile. Hierbei wird ein digitaler Zwilling in Form von Volumendaten (Voxeldaten) erzeugt, bei dem sowohl äußere Geometrien als auch innere Strukturen dargestellt werden können. Die Analysevorgänge sind aufgrund der Datenmengen sehr aufwendig und können durch Benutzereinflüsse inkonsistent sein. Die Anwendung der künstlichen Intelligenz auf Voxeldaten hat das Potential, die Daten präzise, schnell und mit gleichbleibender Qualität auszuwerten, um beispielsweise qualitätsrelevante innere Defekte zu detektieren. Bei der hier vorgestellten Arbeit werden neuronale Netzarchitekturen weiterentwickelt, trainiert und ausgewertet. Das entwickelte automatisierte Fehlererkennungssystem erreicht mit den Testdaten eine Erkennungsleistung von ca. 93%.*

**Schlüsselworte.** *Computertomographie, Voxeldaten, künstliche Intelligenz, U-Net, V-Net*

## 1 Zerstörungsfreie Bauteilprüfung

In unserer schnelllebigen Industrie ist es entscheidend wichtig, Produkte möglichst schnell, in großer Stückzahl und mit geringem Ausschuss fertigen zu können. Die Qualitätssicherung der Produkte und die hierfür benötigte Zeit ist jedoch nicht zu unterschätzen. So werden durch die Maßnahmen der Qualitätssicherung u.a. die Ausschussquote und die Fertigungszeit beeinflusst.

Die Qualitätsprüfmethoden lassen sich in zwei Bereiche, die zerstörende und die zerstörungsfreie Prüfung unterteilen. Abhängig vom Produkt und den zu prüfenden Eigenschaften können bzw. müssen unterschiedliche Prüfverfahren durchgeführt werden. Die zerstörende Bauteilprüfung führt jedoch zu einer Zerstörung des Bauteils, wodurch zusätzlicher Ausschuss erzeugt wird. Die Analyse eines Bauteils ist außerdem unvollständig, da sich immer nur einzelne Bereiche, des dabei zerstörten Bauteils, untersuchen lassen. Deshalb lassen sich die Ergebnisse nicht grundsätzlich auf andere Bereiche des Bauteils übertragen. Die zerstörungsfreie Bauteilprüfung eröffnet hier ganz andere Möglichkeiten. Es wird kein Ausschuss produziert und somit sind diese Verfahren auch bei der Prüfung von Einzelanfertigungen oder bei 100% Kontrollen nutzbar. Ein zerstörungsfreies Verfahren, das im industriellen Umfeld eine zunehmend große Bedeutung hat, ist die industrielle Computertomographie (CT). Diese ist in der Lage, sowohl die äußeren Geometrien des zu untersuchenden Bauteils abzubilden als auch die inneren Strukturen offen zu legen. Das Verfahren führt zu einer Digitalisierung des Bauteils und somit zu einem vollständigen digitalen Zwilling. Die Daten, die einem CT entstammen, sind Volumendaten (Voxeldaten), die aus volumetrischen Pixeln (Voxel) bestehen. Der Helligkeitswert (Grauwert) eines Voxels ist ungefähr proportional zur Dichte des Materials an dieser Stelle.

Die Auswertung der Volumendaten ist gerade bei der industriellen CT-Daten sehr aufwendig. Im Vergleich zu medizinischen CT Daten müssen wesentlich größere (Typisch: Faktor 100) Datenmengen verarbeitet werden. Aktuell werden industrielle CT-Daten vorwiegend manuell von Experten ausgewertet, die jedes einzelne Schnittbild dieser Daten analysieren, um Defekte im Bauteil zu finden. Somit beeinflusst die Entscheidung des Experten das Ergebnis. Auch gibt es bereits automatisierte, auf Algorithmen basierende Verfahren. Hier werden beispielsweise Schwellwertverfahren eingesetzt. Die Schwellwerte werden entweder durch den Bediener oder die Software bestimmt. Dies kann in der Praxis zu variierenden Ergebnissen führen. An dieser Stelle sollen Verfahren der künstlichen Intelligenz (KI) mit neuronalen Netzen zum Einsatz kommen, um sowohl den Bedienereinfluss, die Inflexibilität algorithmischer Ansätze und den Faktor Zeit zu minimieren. Das Ziel dieser Arbeit ist es, eine schnelle, automatische und konsistente Fehlerdetektion in industriellen CT-Daten auf Basis von KI zu realisieren.

Dieser Artikel stellt unsere Forschungsergebnisse im Kontext des Kongresses KI 4 Industry zusammengefasst dar. Für weitergehende Ergebnisse und technische Details sei auf [1] verwiesen.

## 2   KI-Verfahren zur Fehlerdetektion

Der Einsatz von KI-Verfahren ist an die Verfügbarkeit von Trainingsdaten geknüpft. Allgemein gibt es nur wenig öffentlich verfügbare industrielle Voxeldaten infolge der vertraulichen Informationen, die darin enthalten sind. Dieser Mangel wird durch die Erzeugung synthetischer Daten mit künstlichen Defektstellen kompensiert. Eine Analyse der Realdaten (**Abbildung 3**) ermöglicht die Beschreibung von Merkmalen, um diese zu synthetisieren. Hierbei werden Helligkeitswerte der CT-Daten im Material, in den Defektstellen (Poren) und in der Luft untersucht. Weiterhin wird die Beschaffenheit der Defektstellen im Hinblick auf Größe und Form analysiert. Zusätzlich werden Störeinflüsse, wie Rauschen und Unschärfen, die auf den Realdaten vorhanden sind, nachgebildet. Die Erzeugung synthetischer CT-Daten (**Abbildung 4**) bietet mehrere Vorteile. Zum einen können Daten in beliebiger Menge und Variation erzeugt werden. Zum anderen ist während der Erzeugung der Defektstellen im Datensatz auch die automatische Annotation möglich. Somit konnten 3 unterschiedliche Datensätze erstellt werden: 1. Realdatensatz (*Rdata*: 156 Samples); 2. Synthetischer Datensatz (*Gdata*: 7249 Samples); 3. Gemischter Datensatz (*Mdata*: 156 + 7249 Samples). Die Datensätze werden jeweils in Trainings-, Evaluations- und Testdaten aufgeteilt.

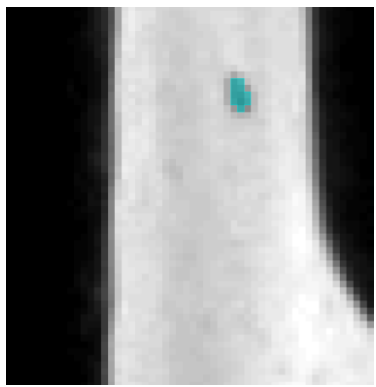

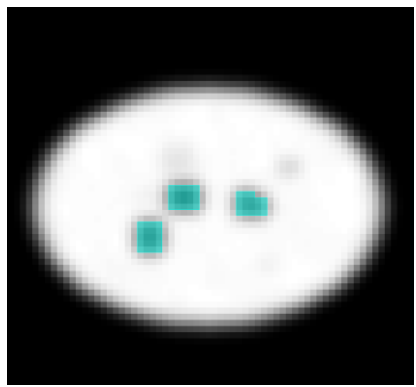

**Abbildung 3:** Beispielausschnit aus realen CT-Daten mit einem Voxelvolumen von 64x64x64

**Abbildung 4:** Beispielausschnit aus synthetischer CT-Daten mit einem Voxelvolumen von 64x64x64

Die neuronalen Netze, die im Rahmen dieser Arbeit zur Segmentierung von industriellen CT-Daten eingesetzt werden, stammen aus der Medizin. Die hier vorwiegend verwendeten Architekturen sind U-Net (**Abbildung 5**) und V-Net (**Abbildung 6**) und ihre Derivate. Neben der reinen Topologie gilt es die Netzparameter anzupassen. Ein neuronales Netz muss abhängig von den verwendeten Daten und dem Anwendungsfeld sowie der Beschaffenheit, der zu segmentierenden Merkmale, optimiert werden. Dies erfordert eine systematische Vorgehensweise, um für die zwei grundlegenden und hier weiterentwickelten Typen U-Net und V-Net, optimale Topologien und Netzparameter zu ermitteln. Die Optimierung der Netze und ihrer Parameter wird hierfür automatisiert.

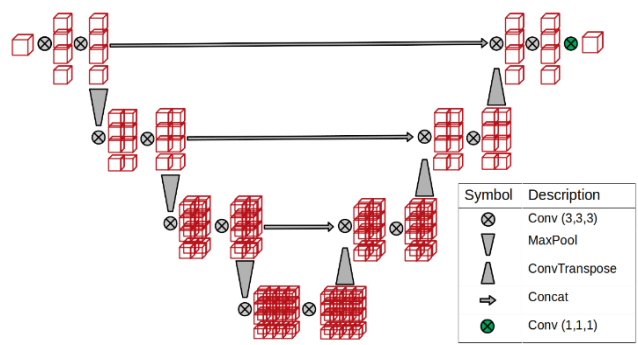

**Abbildung 5:** Beispiel der grundlegenden U-Net Architektur als Basis für die Weiterentwicklung

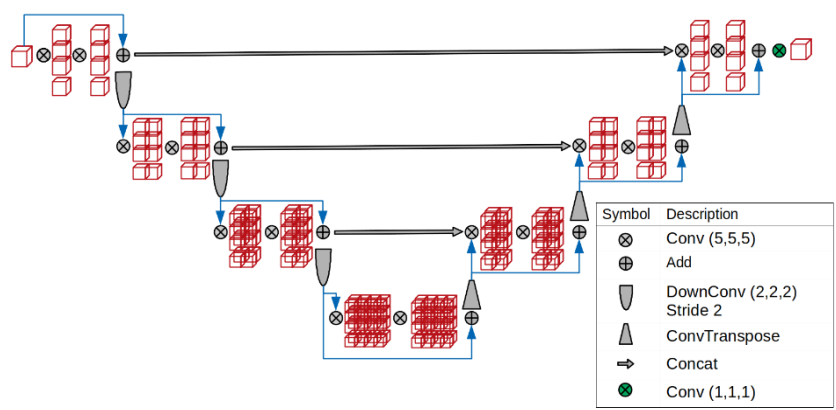

**Abbildung 6:** Beispiel der grundlegenden V-Net Architektur als Basis für die Weiterentwicklung

Vor dem Training der Netze sind neben den Trainingsparametern auch geeignete Bewertungsmetriken festzulegen. Nach dem Stand der Forschung wird für die Bewertung der Übereinstimmung der Groundtruth mit der Vorhersage bei Segmentierungsaufgaben überwiegend die IoU – Intersection over Union (1) verwendet. Hierbei ist $P$ die Vorhersage und $G$ die Groundtruth. Somit steht ein IoU-Wert von 1 für eine ideale Übereinstimmung von Vorhersage mit Groundtruth. Der gemittelte Wert ist die Mean IoU (MIoU).

$$IoU = \frac{|P \cap G|}{|P \cup G|} \qquad (1)$$

Diese Metrik wird mit anderen verglichen und in Anwendungsbeispielen verifiziert. Neben vielen bereits durch die automatische Optimierung ausgeschlossenen Netztopologien und Netzparametern, resultieren für das Training 8 Netze. Diese sind unterteilt in 4 U-Net Architekturen und 4 V-Net Architekturen, die jeweils in zwei unterschiedlichen Größen (Variante Large: $L$ und normal) modelliert sind. Da der Umfang an realen Daten zu gering ist, werden keine Modelle erstellt, die rein auf diese trainiert sind. Das Training der Netze wird mit dem rein synthetischen und dem gemischten Trainingsdatensatz durchgeführt.

Aus dem Training gehen mehrere Modelle, die mit unterschiedlichen Daten gelernt wurden, hervor. Die Evaluation dieser wird gegenüber ungesehener Testdaten durchgeführt. Die Testdaten stammen sowohl aus dem synthetischen als auch aus dem gemischten Testdatensatz. Somit werden alle Netze gegen alle Testdatensätze evaluiert. Aus **Tabelle 1** geht hervor, dass die U-Net Architekturen tendenziell höhere MIoU-Werte erzielen. Weiterhin lässt sich ableiten, dass die Werte unabhängig davon, ob die Modelle mit synthetischen oder gemischten Daten gelernt wurden, ähnlich ausfallen. Hierdurch lässt sich zeigen, dass die Merkmale der synthetischen Daten denen der realen Daten stark ähneln. Somit zeigt sich, dass synthetische Daten für diesen Anwendungsfall zielführend sind. Das Modell „VNET-Gdata" schneidet mit ~74% am schlechtesten und das „UNET-Gdata" mit ~93% am besten ab.

**Tabelle 1:** Auswertung der Modelle anhand der MIoU

| Modell | MIoU gegen synthethische Testdaten in % | MIoU gegen gemischte Testdaten in % |
|---|---|---|
| VNET-Gdata | 74,14 | 74,11 |
| VNET-L-Mdata | 75,32 | 75,29 |
| UNET-Mdata | 80,47 | 80,44 |
| VNET-L-Gdata | 83,21 | 83,18 |
| VNET-Mdata | 84,71 | 84,68 |
| UNET-L-Mdata | 86,96 | 86,92 |
| UNET-L-Gdata | 88,63 | 88,59 |
| UNET-Gdata | **93,29** | **93,25** |

Neben der rein statistischen Auswertung werden Stichproben, sowohl gegenüber synthetischen, als auch realen Samples durchgeführt, wohingegen das beste und schlechteste Modell verglichen werden. In **Tabelle 2** sind 2D-Schnitte der Stichproben als Gegenüberstellung dargestellt. Gegenüber Sample Nr. 1 scheiden beide Modelle vergleichbar gut ab. Für diesen Schnitt stimmt die Vorhersage mit der Groundtruth ideal überein. Das V-Net zeigt hierbei leichte Abweichungen im Bereich der Groundtruth. Bei Sample Nr. 2 sind sowohl beim U-Net als auch beim V-Net fehlerhafte Vorhersagen zu erkennen, dennoch ist auch hier das U-Net überlegen. Im Vergleich (Sample Nr. 2) ist die Vorhersage gegenüber einem realen Sample dargestellt. Trotz des geringen Umfangs an realen Daten verdeutlicht dieser Vergleich die Leistungsfähigkeit der Modelle. Hier sagt das U-Net sogar eine mutmaßliche Defektstelle vorher, die bei der Annotation womöglich übersehen wurde oder durch den Experten nicht als Fehlstelle interpretiert wurde. Hier kann jedoch lediglich ein Experte eine Qualitätsaussage treffen. Die Vorhersage des V-Net ist unbrauchbar. An diesen Beispielen lässt sich noch einmal verdeutlichen, dass synthetische Daten für realen Anwendungsfälle genutzt werden können und somit ein Mangel an Daten ausgeglichen werden kann.

**Tabelle 2:** Übersicht der Vorhersagen der unterschiedlichen Modelle auf synthetische und reale Samples

| Modell | Datentyp | Sample Nr.: | Sample | Groundtruth | Vorhersage |
|---|---|---|---|---|---|
| UNET-Gdata | synthetisch | 1 | 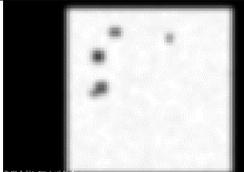 | 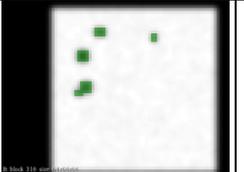 | 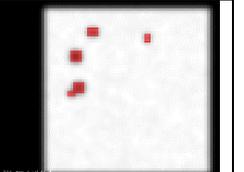 |
| VNET-Gdata | synthetisch | 1 | 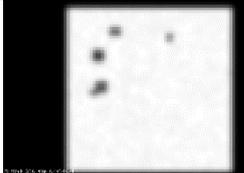 | 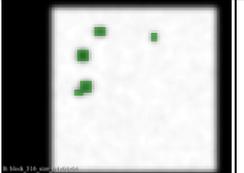 | 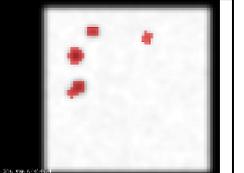 |
| UNET-Gdata | real | 2 | 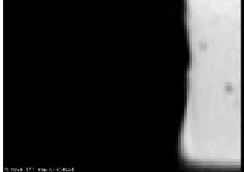 | 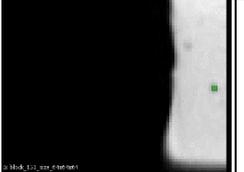 | 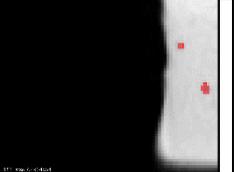 |
| VNET-Gdata | real | 2 | 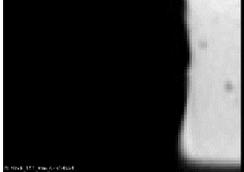 | 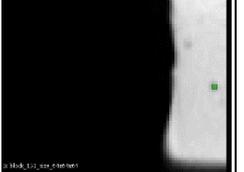 | 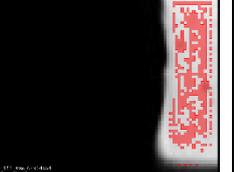 |

## 3  Fazit

In dieser Arbeit zeigen wir, dass neuronale Netzarchitekturen, die für medizinische Anwendungen konzipiert wurden, durch Weiterentwicklungen auch für die Fehlerfindung in industriellen CT-Daten eingesetzt werden können. Die gelernten Modelle sind in der Lage, zuverlässig, schnell und ohne Benutzereinfluss die zu detektierenden Fehler in den Bauteilen zu finden. Die erreichte Erkennungsleistung mit der weiterentwickelten U-Net Architektur bestätigt unsere Ansätze. Gerade der Nutzereinfluss kann zum einen durch die Nutzung synthetischer Daten mit konsistenter Annotation und zum anderen, soweit verfügbar, durch eine hohe Varianz in den Realdaten ausgeglichen werden. Ein Mangel an Daten kann durch die Erzeugung synthetischer Daten ausgeglichen werden, wenn gewährleistet werden kann, dass die wesentlichen Merkmale enthalten sind. Neben der grundlegenden Struktur der Netze konnte deren Topologie und die Netzparameter durch automatisierte Verfahren schnell und effektiv optimiert werden, wodurch die Netze auf die Daten abgestimmt werden konnten. Letztendlich konnte eine Erkennungsleistung von etwa 93% realisiert werden.

## Literaturverzeichnis

# Plug and Play KI – Open Source Lösungen mit Tensorflow

*Philip Scherer, Iván Lozada Rodríguez, Rüdiger Haas*

*Hochschule Karlsruhe*
*Institute of Materials and Processes*
*Karlsruhe, Deutschland*
*philip.scherer@h-ka.de, 0721 925-2079*
*ruediger.haas@h-ka.de, 0721 925-2055*

***Zusammenfassung****. Künstliche Intelligenz hält immer mehr Einzug in den Alltag und die Industrie. Längst sind moderne Kamerasysteme mit zusätzlichen KI Funktionen ausgestattet, um das das Bildmaterial zusätzlich aufzuwerten.*

*Die Verwendung von KI-Methoden ist jedoch nicht nur den Global Playern vorbehalten. Verschiedene Frameworks wie beispielsweise TensorFlow, Keras und PyTorch gepaart mit bereits vortrainierten Netzen erleichtern den Einstieg in das Thema Künstliche Intelligenz. Diese frei zugänglichen Werkzeuge lassen insbesondere unerfahrenere Anwender bereits schnell vorzeigbare Ergebnisse erzielen.*

*In der vorliegenden Veröffentlichung wird das prinzipielle Vorgehen einer KI Entwicklung von der Zieldefinition über die Datenvorbereitung bis hin zur Erstellung des Inferenzgraphs am Beispiel einer Klassifikation von Sounddaten beschrieben.*

***Schlüsselworte****. KI, TensorFlow, CNN, Data Science, Prozessüberwachung*

## 1 Einleitung – Ausgangssituation und Zielstellung

Für viele Problemstellungen sind bereits Lösungen basierend auf Künstlicher Intelligenz (KI) vorhanden und können mit überschaubarem Aufwand angewendet werden. Im Folgendem wird am Beispiel einer Auswertung von akustischen Signalen mithilfe von KI das grundsätzliche Vorgehen erläutert.

Gas- oder Ölpipelinesysteme müssen regelmäßig auf etwaige Leckagen getestet werden, die beispielsweise auf Korrosion, Sabotage durch Fremdentnahme oder Schäden durch Baggerarbeiten zurückzuführen sind. Dabei ist es notwendig, dass mehrere hundert Kilometer Pipeline überprüft werden. Ein mögliches Prüfverfahren, welches im weiteren Verlauf betrachtet wird, stellt die akustische Leckagedetektion dar.

Bei der akustischen Leckagedetektion werden die Sensordaten manuell ausgewertet und mögliche Leckagen müssen aufgrund von erlangten Erfahrungen erkannt werden. Dieses Vorgehen erfordert viel Erfahrung und kann abhängig von der Pipelinelänge und damit der Anzahl der Daten sehr zeitintensiv sein. Das Ziel ist es, diesen manuellen Vorgang mithilfe von KI-Methoden zu automatisieren. Das bedeutet, es wird ein Datensatz einer Prüfung geladen, ausgewertet und ein entsprechendes Protokoll ausgegeben. In diesem Protokoll werden die als Leckage klassifizierten Daten mit Informationen zu Ort/Zeit und Wahrscheinlichkeit einer Leckage dargestellt.

Dieses System ist Grundlage für die nächste mögliche Weiterentwicklung bei welcher die Integration der KI in die Messeinheit vorgenommen wird, sodass lediglich relevante Daten aufgezeichnet werden. Dafür muss zunächst die Auswerteeinheit robust funktionieren und die Hardware der Messeinheit muss dementsprechend angepasst werden, dass diese mit der KI umgehen kann. Ein weiterer Punkt der betrachtet werden muss betrifft die Echtzeitfähigkeit der KI. Das bedeutet, dass die Sensordaten mindestens genauso schnell verarbeitet werden können wie diese vom Sensor erfasst werden, um somit einen Datenstau zu vermeiden.

## 2    Daten und Methodenauswahl

Im ersten Schritt werden die verfügbaren Daten betrachtet, um daraus zusammen mit der jeweiligen Zielstellung entscheiden zu können, ob und welche KI-Methoden für diesen Anwendungsfall sinnvoll genutzt werden können.

Der Sensor der Messeinheit liefert Rohsignale mit einer maximalen Frequenz von bis zu 125 kHz. Diese Rohsignale werden mit einer Fast-Fourier-Transformation (FFT) für ein definiertes Zeitintervall in ihre Frequenzanteile zerlegt. Demnach sind zu jedem Zeitpunkt, genauer gesagt für jedes Zeitintervall, die Werte der einzelnen Frequenzbänder bekannt.

Das Identifizieren von Leckagen entspricht einem klassischen Klassifizierungsproblem, welches mit Hilfe eines Neuronalen Netzes (NN) gelöst werden kann. Insbesondere mit stark strukturierten Daten, wie zum Beispiel Bildmaterial, arbeiten diese Neuronalen Netze zuverlässig. Gefaltete Neuronale Netze (Convolutional Neural Network - CNN) sind besonders für die Extraktion von Merkmalen (Features) aus Bildmaterial geeignet, diese Extraktion von Merkmalen ist maßgeblich dafür verantwortlich die Unterschiede in den Bildern zu erkennen. Das vollständige Training, Merkmal Extraktion und Klassifikation, eines solchen CNN erfordert eine sehr große Anzahl an Daten. Sollte keine große Datenmenge vorhanden sein oder ein schneller Versuch durchgeführt werden, ob das Verwenden von einem CNN überhaupt sinnvoll ist, kann stattdessen ein bereits vortrainiertes Netz verwendet werden. Diese vortrainierten Netze haben die Merkmalextraktion bereits anhand von großen, meist öffentlich zugänglichen Datensätzen erlernt. Bei diesem Vorgehen, Transfer Learning genannt, kann das Problem mit deutlich weniger Daten modelliert werden, da weniger Parameter gelernt werden müssen. Als vortrainiertes CNN kann beispielsweise das „faster-rcnn-resnet101-coco" verwendet werden [1]. Das Netz ist öffentlich zugänglich und wurde mit dem umfangreichem COCO-Datensatz trainiert [2]. Außerdem ist die Netzarchitektur mehrfach optimiert, um die Geschwindigkeit zu erhöhen.

## 3    Datenvorbereitung und Aufteilung

Die in Frequenzbänder vorliegenden akustischen Daten müssen zunächst in passendes Bildmaterial für die Verarbeitung mit dem gewählten CNN konvertiert werden. Dazu bietet es sich an, die Daten in einem Spektrogramm zu visualisieren. Dabei werden auf der x- und y-Achse die Zeitdiskretion und die Frequenz dargestellt. Die Färbung der einzelnen Datenpunkte (quasi z-Achse) stellt den Wert an dieser Stelle dar, vgl. **Abbildung** . Verschiedene Bibliotheken in Python bieten die Möglichkeit Sounddateien direkt in ein Spektrogramm umzuwandeln (z.B. scipy oder librosa). Zusätzlich müssen für das Training des CNN die einzelnen Bilder „gelabelt" werden. Die Pixel in jedem Bild anhand denen eine Leckage erkannt werden kann, werden in einer .xml-Datei definiert und einer Leckage zugeordnet. Das Neuronale Netz wird auf diese Label optimiert bzw. die Parameter werden so angepasst, dass der Fehler am Ausgang des Netzes zu diesen Labeln minimal wird. Das Training lässt sich demnach mit einem Optimierungsproblem vergleichen.

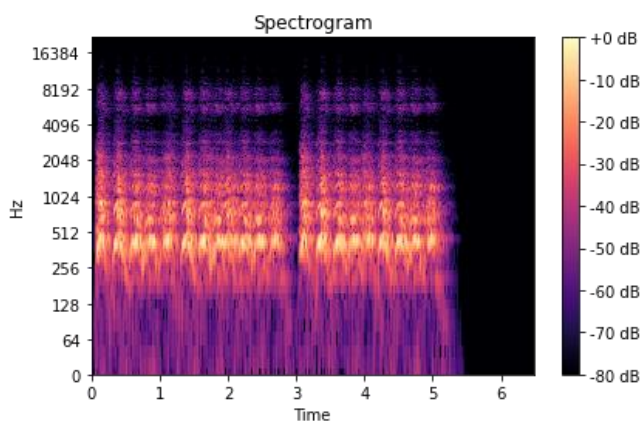

**Abbildung 1:** Spektrogramm von Hundegebell

Bevor die Trainingsdaten, .xml-Datei und Bild, zu einer binären .record-Datei konvertiert werden können, müssen diese aufgeteilt werden. Richtwerte die in der Praxis gut funktionieren sind 80% der Daten zum Training zu verwendet und 20% zum Überprüfen der Performanz. Der Datensatz wird getrennt, um zu überprüfen, ob das Netz das gelernte auf neue Daten erfolgreich anwenden kann (Stichwort: Generalisierung). Außerdem sollte zusätzlich ein kleiner Teil von Daten vorhanden sein, der zum Schluss zur finalen Betrachtung der Performanz des CNN verwendet wird. Dieser Datensatz sollte weder im Trainings- noch im Testdatensatz inkludiert sein. In **Abbildung 2** ist der gesamte Ablauf der Datenvorbereitung dargestellt.

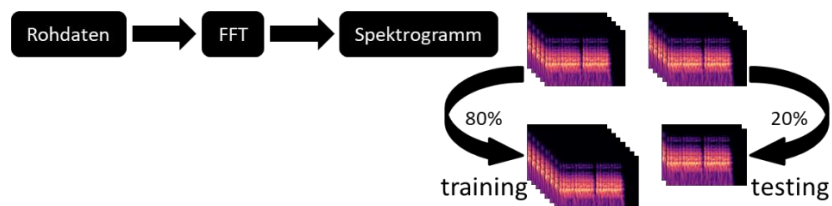

**Abbildung 2:** Ablauf der Datenvorbereitung

## 4    Training und Export des CNN

Das vortrainierte neuronale Netz „faster-rcnn-resnet101-coco" kann zusammen mit einer vorgefertigten Konfigurationsdatei heruntergeladen werden. In dieser Konfigurationsdatei müssen hauptsächlich Pfade zu benötigten Dateien angegeben werden. Ein weiteres Anpassen ist möglich, jedoch nicht zwingend notwendig. [1]

Weiterhin muss die Input Pipeline, die Daten die als Eingang an das Netz übergeben werden, definiert werden. Dazu gehören die zuvor erstellten Trainings-, Testdaten und die editierte Konfigurationsdatei. Analog zur Input Pipeline wird die Output Pipeline definiert. Diese bestimmt, welche Daten vom CNN empfangen werden und wie diese weiterverarbeitet werden. Beispielsweise eine Ausgabe in der Konsole oder Speichern in einer Datei. Mit diesen Informationen als Übergabeparameter kann das Trainingsskript des Netzes, welches ebenfalls mit dem Netz heruntergeladen wird, aufgerufen werden. Weitere Informationen zu den jeweiligen Eingangs- und Ausgangsparemetern ist entweder der „Readme" oder der Github / Projektseite des jeweiligen Netzes zu entnehmen. Die Dokumentation ist in den meisten Fällen mehr als ausreichend ausgeführt.

Ist die Fehlerfunktion des CNN konvergiert oder das Training wird manuell abgebrochen, kann das Netz überprüft werden. Dazu werden die Daten verwendet die noch keinen Kontakt mit dem Netz hatten, und somit auch keinen Einfluss auf dessen Struktur/Eigenschaften hatten.

Erfüllt das CNN die geforderte Performanz, wird ein sogenannter Inferenzgraph exportiert werden. Das bedeutet, das Netz wird eingefroren um für die Verwendung vorbereitet. Die Variablen in den einzelnen Schichten werden zu Konstanten, Sackgassen zwischen den Schichten und Debugging Operationen werden entfernt. Diese Maßnahmen sorgen dafür, dass das Netz schneller durchlaufen werden kann. Die Maßnahmen führen zusätzlich dazu, dass ein gleichbleibendes Netz erzeugt wird. Das Verwenden dieses Netzes hat somit keinen Einfluss mehr auf dessen Eigenschaften. Die Entwicklung der KI ist an dieser Stelle abgeschlossen.

## 5    Fazit

Das Kernelement zur automatischen Auswertung der Messdaten ist geschaffen. In den nächsten Schritten ist es notwendig, ein Umfeld für diesen Kern zu generieren. Die Organisation der Übergabe der Messdaten bis hin zur Verarbeitung der Ausgangswerte des CNN müssen umgesetzt werden. Diese Aufgaben können ebenfalls mithilfe von Python und entsprechenden Bibliotheken in einem überschaubaren Arbeitsaufwand umgesetzt werden. Ebenso das Generieren von Berichten anhand der Ausgangsdaten des CNN ist möglich.

Sollte mit der Zeit das Netz nicht mehr für den Anwendungsfall zutreffend sein, da sich beispielsweise grundlegende Charakteristiken geändert haben, es ein weiteres Element zu klassifiziert gilt oder sich ein weiteres Anwendungsgebiet ergeben hat, kann der Klassifikationsteil des CNN mit neuen Daten erneut durchgeführt werden, die vortrainierten Bildanalyse Layer können jedoch beibehalten werden.

## 6 Kooperationsprojekt

## Literaturverzeichnis

# KI ist menschlich - Eine Zeitreise zu den digitalen Herausforderungen auf dem Weg zum Menschsein

Janine Schwienke

MINT Coaching
Karlsruhe, Deutschland
info@mint-coaching.de, +49 173 44 75 111

*Zusammenfassung. Wer KI als Unterstützung für uns Menschen nutzen möchte, muss zuerst uns Menschen verstehen. Welche Bedürfnisse haben wir und an welchen Stellen können sich KI und wir Menschen uns gegenseitig ergänzen?*

*Schlüsselworte. Mensch, Evolution, Cyber-Lilly, Steini, Steinzeit, Zivilisation, Zukunft, Empathie, Humor, Moral, Programmierer, Bedürfnisse, Fehler, Diversität, Verantwortung, Leitlinien, Eid zum Wohle der Gesellschaft*

## 1    Wo kommen wir her und wo könnten wir hingehen?

Was würde passieren, wenn man plötzlich mit einer Freundin aus der Zukunft – sagen wir, aus dem Jahr 2065 – und einem Freund aus der Steinzeit – so von vor 90.000 Jahren – zusammen bei einem Kaffee ins Plaudern gerät?

Welche Gedanken und Erfahrungen bringen sie mit und was schließen wir daraus für das hier und jetzt?

Stellen Sie sich vor, diese Freundin aus der Zukunft wundert sich ganz schön über unser Verhalten. Alles ist so hektisch hier. Die Menschen stehen dauernd unter Stress. Sie verrichten nur eintönige Arbeiten und sind eher wie Maschinen. Und sie haben keine Wünsche und Träume – geschweige denn, dass sie sie auch noch verfolgen.

Diese Freundin – nennen wir sie Cyber-Lilly (Abbildung 1) - erinnert sich, dass ihre Eltern davon berichtet haben, dass die Menschheit sich vor ca. 40 Jahren fast selbst ausgerottet hat. Alles ging nur schneller, weiter, höher. Das Stresslevel war megahoch. Es gab unheimlich viele Menschen mit psychischen Krankheiten. Die Welt ist im Müll erstickt. Und für jeden Mist gab es eine App oder eine technische Lösung. Hauptsache man konnte Business machen. Doch irgendwann gab es dann zum Glück eine Wende, in der die Menschen verstanden haben, dass sie an sich selbst denken sollten.

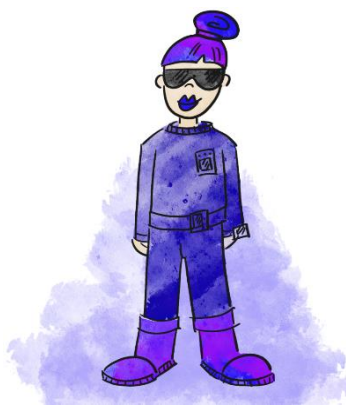

**Abbildung 1:** Cyber-Lilly

Und nun berichtet Cyber-Lilly aus der Zukunft, und dass sie im Jahr 2065 ihre Cyber-Friends haben. Das sind die Roboter, die ihnen die eintönige Arbeit abnehmen, die Arbeit die vor allem von Algorithmen angetrieben wird oder die viel Kraft erfordert. Und damit haben die Menschen Zeit sich mit den wesentlichen Aufgaben zu beschäftigen, die sie viel besser können. So haben sie Jobs, bei denen Menschlichkeit und Kreativität zählt. Es ist Kapazität da, um verrückte Sachen auszuprobieren. Sie können experimentieren und sie können auch Fehler machen. Doch das alles geht nur, weil sie auch viel Ruhe und Zeit haben, um sich genau den Dingen zu widmen, die die Menschheit nach vorne bringt.

Spannende Vorstellung, die Cyber-Lilly da so erzählt. Und plötzlich meldet sich mein Freund aus der Steinzeit. Er heißt Steini (Abbildung 2) und erzählt immer so spannende Geschichten von dem wilden Leben jenseits der Zivilisation. Und nun stellt Steini bei den Geschichten von Cyber-Lilly fest, dass ihm das sehr bekannt vorkommt. In der Steinzeit funktioniert das Miteinander sehr ähnlich – ohne KI natürlich. Um zu überleben war es wichtig, dass die Sippe kreativ ist, mutig ist, neue Wege gehen musste. Und das hat nur funktioniert mit gegenseitigem Vertrauen und Austausch, mit Hinterfragen und Zweifeln am Status Quo. Wichtig war auch die Fähigkeit, verschiedene Aspekte zusammenzubringen, um dann etwas Neues entstehen zu lassen.

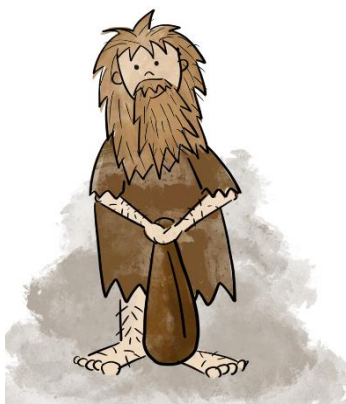

**Abbildung 2:** Steini

Doch was ist passiert seit der Steinzeit? Vor ca. 11.000 Jahren sind die Menschen sesshaft geworden und haben begonnen Ackerbau und Viehzucht zu betreiben. Es entstand Hierarchie. Und es wurden plötzlich Nahrungsmittel gegessen, die der Mensch bis dahin gar nicht auf seinem Speiseplan hatte. Das waren vor allem Getreide und Milch. Und von Evolutionsbiologen hat Steini auf seiner Zeitreise erfahren, dass man in den alten Knochen erstmalig mit der Sesshaftwerdung auch Zivilisationskrankheiten wie Diabetes, Übergewicht, Autoimmunkrankheiten und auch Karies und viele andere mehr festgestellt hat. Das ist also auch gar nicht mal so neumodisch. Die Vermutung liegt nahe, dass die 11.000 Jahre nicht ausgereicht haben, sich genetisch optimal anzupassen, und bedeutet etwa 300-400 Generationen. Wenn man mal nachrechnet entspricht die Zeit, in der die Menschen in Zivilisation leben, gerade mal ein halbes Prozent der gesamten Zeit, die der Mensch auf der Erde lebt– viel zu kurz zum Anpassen [1].

Steini weiß, dass die Evolution viel Zeit hat bzw. braucht. Und nun wundert es ihn mächtig, dass unsere heutige Gesellschaft versucht, den Menschen so zu biegen und zu formen, dass er sich den neuen digitalen Verhältnissen anpasst. Was die Evolution in 11.000 Jahren nicht geschafft hat, kann erst recht nicht in 30 Jahren funktionieren. Wir müssen endlich aufwachen.

Tja und als ob es ein Traum war, kommen wir wieder zurück in die reale Welt und schauen uns mal an, was für uns wirklich zählen sollte.

## 2    Programmierer sind die Gestalter unserer Zukunft – 5 Leitlinien für ein verantwortungsbewusstes Handeln

Konrad Zuse, der Erfinder des Computers in den 1940er Jahren hat gesagt: „Die Gefahr, dass der Computer so wird, wie der Mensch ist nicht so groß, wie die Gefahr, dass der Mensch so wird, wie der Computer."

Hat er damit nicht schon sehr weise vorhergesagt, was gerade passiert? Imitieren wir nicht zu oft das, was der Computer viel besser kann: Lagerregale auffüllen, Tabletten in Seniorenheimen sortieren, rechnen, Gedichte auswendig lernen? Doch es ist so oft nur Maschinenarbeiten, was wir vollbringen.

Wir Menschen sind doch zu ganz anderen Dingen in der Lage, die Maschinen nicht können. Wir haben Empathie und können uns in Mitmenschen hineinfühlen. Wir haben Humor und ein Moralverständnis. Wir sind kreativ und können um die Ecke denken. Und wir lernen nicht nur auswendig, sondern wir können verstehen.

Deshalb haben Programmierer in einer digitalen Welt eine ganz besondere Verantwortung. Sie sind die Gestalter der Zukunft und auf ihre innere Haltung kommt es an, um ein lebenswertes Miteinander möglich zu machen.

Dafür habe ich 5 Leitlinien für Programmierer zusammengestellt, die zukünftig – und gerne auch ab sofort – bedeutungsvoll sein sollten. Folgende Leitlinien sind das:

1. **Setzt den Menschen in den Mittelpunkt!** [3]
   Bevor Programmierer überhaupt die erste Codezeile schreiben, sollten sie viel nachdenken. Wird die Technologie, die sie entwickeln wollen, für den Menschen sein? Wird sie dem Menschen – und seinem dringend benötigten Umfeld und der Umwelt – vom Nutzen sein? Oder arbeitet die Technologie nur, um den Menschen zu ersetzen oder schlimmer noch, ihn auszunutzen?

   Um das Bewusstsein zu schärfen, kann es hilfreich sein, sich real oder zumindest virtuell einen Stuhl mehr in gemeinsame Besprechungen dazu zustellen. Dort sitzt dann „der Mensch", um den es in dem Projekt geht. So rückt er einfach ein Stück näher, und schärft das Bewusstsein für seine Bedürfnisse.

   Zwei Beispiele, die unterschiedlicher nicht sein können. Man hört mitunter von den Arbeitsrobotern in den Altenheimen, die die menschliche Arbeit ersetzen – ob als Pflegekraft oder als Plüschrobbe. Ich persönlich finde die Vorstellung gruselig, als alter Mensch irgendwann von einer Maschine gepflegt oder bespaßt zu werden.
   Doch ich habe auch schon von Technologien gehört, wo Robotergestelle für den Arm-und-Schulterbereich die Pflegekräfte unterstützen, indem sie die Kraft beim Aufrichten älterer Menschen unterstützen. Und wenn es dann noch die Tabletten-Sortiermaschine gibt, die fehlerfrei die richtigen Tabletten sortiert, dann wäre das schon ein riesiger Schritt in eine Richtung, in der die Pflegekräfte sich mit ihrer Menschlichkeit wieder mehr den älteren Menschen widmen können, statt sie nur zu versorgen. Somit leisten sie einen kraftvollen Beitrag für einen zuwendungsvollen Lebensabend.

2. **Stellt gute Fragen!**
   Wir alle sind trainiert darauf, gute Antworten zu geben. Doch wir müssen neue Fragen stellen. Henning Beck [2] bringt es auf den Punkt. Wir sollten uns nicht nur die Fragen stellen, warum etwas so funktioniert. Sondern wir sollten uns auch fragen, wozu wir die Ideen und Entwicklungen wirklich brauchen? Können sie ein echtes Problem lösen? Sind sie enkeltauglich, also zukunftsfähig, nachhaltig und gesellschaftlich sinnvoll?

3. **Macht Fehler!**
   Perfektionismus kann die Maschine viel besser als wir. Fehler sind menschlich. Und Fehler können neue Ideen entstehen lassen. Wir brauchen eine neue Fehlerkultur, die Experimente zulässt, und deren Nichtgelingen nicht gleich als Schwächen abstempelt wird. Wir müssen uns ausprobieren können, um in dieser sehr komplexen Welt Schritt für Schritt den Lösungen näher zu kommen.

4. **Lasst Diversität und verschiedene Ansichten zu!**

...denn das erweitert den eigenen Horizont. Wenn Diversität in eure Produkte und Dienstleitungen einfließt, wird ein ganzheitlicher Ansatz zukünftige Entwicklungen beeinflussen. Stellt eure Teams nach verschiedenen Kriterien zusammen:

- verschiedenste Fachbereiche
- Männer und Frauen
- verschiedene Hautfarben, Kulturen, Nationalitäten
- alte und junge Menschen
- Gläubige und Nicht-Gläubige
- Heterosexuelle und Homosexuelle
- usw.

5. **Übernehmt Verantwortung!**

Setzt euer Handeln ins richtige Verhältnis! Bleibt auf Augenhöhe. Programmierer haben eine hohe Verantwortung. Denn die Datensammlung und die Prozesse der Software sind für den Otto-Normalverbraucher längst nicht mehr überschaubar und greifbar. Und deshalb können nur die Produkte überleben, die enorm vertrauenswürdig sind. Programmierer handeln nachhaltig, wenn sie sich nicht über die Nicht-Wisser stellen und sie damit auch ein Stück weit beherrschen. Sondern wenn sie sich immer auf Augenhöhe mit den Konsumenten reflektieren, welche Auswirkungen ihr Handeln hat.

Mit den 5 Leitlinien kann es eine Richtung geben, in der die Gestalter der digitalen Zukunft mit einer inneren Haltung auch für Nachhaltigkeit und Selbstverantwortung stehen.

Andera Gadeib ist Zukunftsdenkerin und sie hat in ihrem Buch „Die Zukunft ist menschlich" [3] gefordert, dass Programmierer zukünftig ähnlich wie Ärzte und Heilpraktiker für die Gesundheit einen Eid zum Wohle der Gesellschaft ablegen. Und wer zuwiderhandelt, dem kann eben dann die Erlaubnis zum Programmieren entzogen wird – ähnlich wie beim Autofahren.

Ich appelliere deshalb an das Verantwortungsbewusstsein der Programmierer und Verantwortlichen, die Zukunft auch mit neuen Technologien im Stück menschlicher zu gestalten.

Und es ist mein innigster Wunsch, dass es uns gelingt, uns mit Steini zusammen zu besinnen, wo wir herkommen, um unsere eigentliche Stärke und Energie zu nutzen, um genau dahin zu gehen, wo Cyber-Lilly schon ist. In diesem Sinne wünsche ich uns allen eine fantastische Zukunft.

## Literaturverzeichnis

# KI in der Mechatronik – Ausgewählte Beispiele

*Robin Tenscher-Phlipp, Tim Schanz, Martin Simon*

*Hochschule Karlsruhe*
*Fakultät für Maschinenbau und Mechatronik*
*Karlsruhe, Deutschland*
*robin.tenscher-philipp@h-ka.de, +49 721 925-1721*
*tim.schanz@h-ka.de, +49 721 925-1719*
*martin.simon@h-ka.de, +49 721 925-1720*

**Zusammenfassung**. *Die Mechatronik beschreibt ein weitläufiges Gebiet in den Ingenieurswissenschaften. Sie setzt sich aus den großen Teilgebieten Mechanik, Elektronik und Informatik zusammen. Der Zusammenschluss der einzelnen Gebiete ermöglicht es, komplexe Systeme in der Mechatronik zu realisieren. Entscheidungen, Reaktionen und Verhaltensweisen von Systemen beruhen in der Regel auf mehreren Einflussfaktoren und Zusammenhängen. Durch die Komplexität der Systeme kommen klassische oder algorithmische Verfahren häufig an ihre Grenzen. An dieser Stelle kann eine künstliche Intelligenz (KI), zum Einsatz kommen. Eine KI eröffnet durch ihre abstrahierende, dem menschlichen Gehirn nachgebildete Fähigkeit, neue Möglichkeiten in technischen Systemen. In diesem Artikel wird anhand von Beispielen der Einsatz der KI in verschiedenen mechatronischen Bereichen demonstriert.*

**Schlüsselworte**. *KI, künstliche Neuronale Netze, Industrie, Mechatronik, Anwendung*

## 1   Einleitung

Die künstliche Intelligenz (KI) ist von steigender Bedeutung für alle Branchen. So kann die KI und deren Verfahren in der Medizin, Industrie und in Produkten wertschöpfend eingesetzt werden. Wo manuelle oder algorithmische Verfahren ihre Grenzen haben, erweitert die KI diese und erschließt bis dato unbekannte neue Territorien. Global beginnt ein Wettlauf um die Erschaffung und den Einsatz von neuen innovativen KI-Verfahren. Auch Deutschland nimmt an diesem Wettlauf teil und versucht in die oberen Ränge vorzudringen. Um dies zu erreichen, sollten sich nicht nur deutsche Großunternehmen mit entsprechenden Kapazitäten, sondern auch kleine und mittelständische Unternehmen (KMU) dieses zukunftsträchtige Thema bearbeiten.

Für KMUs ist der Einstieg in die KI ein großer Schritt. Oftmals sind weder die Einsatzmöglichkeiten der KI noch fundiertes Knowhow oder die nötigen Softwareentwickler vorhanden. Neben der eigentlichen KI sind die dafür benötigten Daten ebenfalls von großer Bedeutung. Die Aufnahme, Nutzbarmachung und Archivierung von Daten ist die Grundlage für die Umsetzung von KI-Verfahren. Diese müssen digitalisiert, abrufbar und aussagekräftig sein. Auch dieses Erfordernis stellt KMUs oft vor Schwierigkeiten, wo über Jahre Daten in Aktenordnern, Excellisten, Emailverläufen und unterschiedlichen Datenbanken gesammelt wurden, steht die Datenaufbereitung an erster Stelle.

Dieser Artikel soll KMUs für das Thema künstliche Intelligenz sensibilisieren. Hierzu wird der Anwendungsbereich der Mechatronik gewählt. Die Mechatronik setzt sich aus den großen Ingenieursdisziplinen Mechanik, Elektronik und Informatik zusammen. Durch die Kombination dieser Disziplinen können in der Mechatronik komplexe Systeme entstehen. Klassische oder algorithmische Verfahren zur Regelung, Steuerung, Automatisierung, etc. geraten bei dieser Komplexität und Vielfalt an unterschiedlichen Faktoren oftmals an ihre Grenzen. An dieser Stelle wird künstliche Intelligenz eingesetzt, um die Grenzen zu erweitern und somit neue Lösungen zu schaffen.

## 2    Ausgewählte Beispiele

Einen Einblick in die Einsatzmöglichkeiten der KI und die Bedeutung der Daten soll in diesem Artikel anhand von Beispielen gewährt werden. Die Beispiele stammen aus den Bereichen Rauschminimierung, Generative KI, Signalverarbeitung und Objekterkennung.

### 2.1    Rauschminimierung mit KI

Die Verarbeitung von Signalen unterschiedlicher Art ist wesentlich für die Funktionalität von Systemen. Nur wenn Signale möglichst vollständig und eindeutig interpretiert werden können, kann ein System die korrekte Reaktion initiieren. In der Praxis ist es jedoch häufig so, dass Signale von Störsignalen überlagert werden. Im Beispiel einer Bildverarbeitung kann sich eine Störung als Rauschen auf dem Bild äußern. Die Minimierung des Rauschens, um die nutzbaren Informationen im Bild zu maximieren, ist hier ein mögliches Ziel. Klassisch werden Bildstörungen durch algorithmische Filterverfahren vermindert. Allerdings wird hierdurch oftmals die Minimierung des Rauschens durch eine resultierende Unschärfe und das verwaschen von Kanten erkauft. Die KI geht hier andere Wege. Anhand eines Beispiels kann gezeigt werden, dass Faltende neuronale Netze in der Lage sind typische Rauschverhalten auf Bilddaten zu lernen. Die Ausgabe der KI auf ein verrauschtes Eingangsbild ist ein Rauschbild. Durch Verrechnung von Eingangsbild mit dem Rauschbildergebnis der KI kann ein entrauschtes Ausgangsbild erzeugt werden. Die Grundlage hierfür ist die Bestimmung des Rauschverhaltens und die Erstellung geeigneter Trainingsdaten dem ein verrauschtes Bild und eine unverrauschte Groundtruth zugrunde liegt. Die *Abbildung 7*(a) zeigt das Testbild „Barbara"[5]. Das Ausgangsbild wird mit einem Rauschen **Abbildung 7(b)** beaufschlagt. Das Ergebnis der Denoising-KI ist in der *Abbildung 7*(c) dargestellt.  [1]

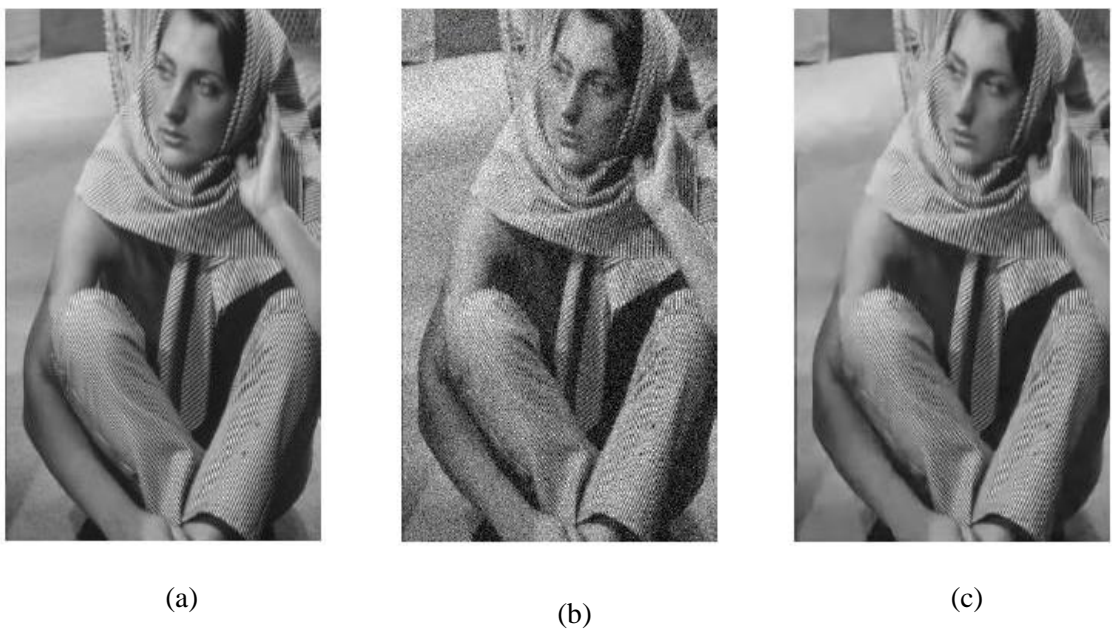

(a)                                   (b)                                   (c)

**Abbildung 7:** Entrauschung eines Bildes mittels KI: (a) Original Barbara-Bild, (b) Barbara-Bild mit Rauschen, (c) Entrauschtes Barbara-Bild [1]

### 2.2    Generative KI

Generative KI ist auch über Expertenkreise hinaus als Schlagwort bekannt, wenn es um Fake News und Fake Bilder geht. Hierbei kann KI eingesetzt werden, um aufgrund von gelernten Eingangsdaten realistische aber nicht reale Nachbildungen von Menschen, Objekten oder Signalen zu erzeugen. Dieser Prozess ist dem Träumen relativ ähnlich. Ein menschliches Gehirn verarbeitet reale Eindrücke aus unserer Umgebung und kreiert daraus Fantasien, die oft täuschend echt erscheinen. Auf ähnliche Weise arbeiten generative neuronale Netze. Diese können auch für mehr als nur die Erzeugung der bereits angesprochen Fake Bilder verwendet werden. Eine konkreter Anwendungsfall bzw. ein Problem stellt der Mangel an Daten für die Umsetzung von KI-Verfahren dar. Mit Hilfe einer generativen KI können durch Bildung

---

[5] Bildnachweis „Barbara": https://homepages.cae.wisc.edu/~ece533/images/barbara.bmp - 10.11.2020

von Variationen der gelernten Eingangsdaten neue Daten erzeugt werden. Diese generierten Daten können dann für andere KI-Verfahren, in denen ein Datenmangel besteht, eingesetzt werden. Eine verbreitete Methode sind so genannte Autoencoder (*Abbildung 8*). Dieser besteht aus 3 Elementen, dem Encoder, der Codeschicht und dem Decoder. Eingangsinformationen werden im Encoder auf ein Minimum heruntergebrochen und abstrahiert, um diese als wenige einzelne Parameter auf der Codeschicht abzubilden. Der Decoder lernt auf Basis der Codeschicht die Eingangsdaten zu Rekonstruieren. Trennt man nach dem Training den Encoder vom Netzwerk ab, so ist durch Variation der Codeschicht die Erzeugung von neuen Ausgangsdaten möglich. Wie in der *Abbildung 8* gezeigt, kann somit durch die Veränderung der Codeschicht eine „3" oder eine abstrakte Variation verschiedener Zahlen entstehen. Die unterschiedlichen Merkmale der zuvor gelernten Daten beinhalten, kombinieren und vereinen sich in den neuen Ausgangsdaten. [2]

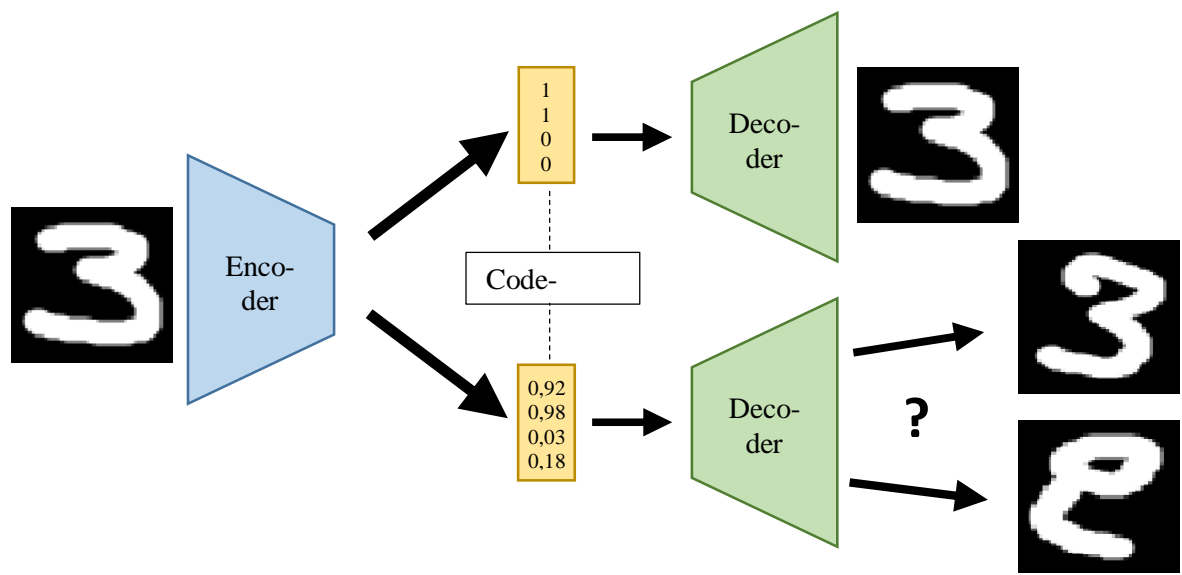

**Abbildung 8:** Beispielaufbau eines Autoencoders und Variation der Ausgangsdaten durch Anpassung der Codeschicht

## 2.3    Signalverarbeitung mit KI

In diesem Beispiel wird die Verarbeitung und Fusionierung von Sensorsignalen mithilfe von KI demonstriert. Oftmals muss eine einzige Entscheidung, also eine Ausgangsgröße aufgrund mehrerer Eingangsinformationen oder Signalen getroffen werden. Ein Fall aus dem Alltag ist die Entscheidung, ob ich eine Jacke anziehe, wenn ich vor die Türe gehe. Diese Entscheidung ist von mehreren Größen abhängig. Welche Temperatur ist draußen? Ist es windig? Regnet es? Passt die Jacke zu meinem Outfit? Hierbei kann es sowohl statische als auch variable Parameter geben. In unterschiedlichen mechatronischen Systemen ist es oft zur Steuerung oder Entscheidung erforderlich, mehrere Signale zu fusionieren und daraufhin eine Ausgangsgröße zu bilden. Die klassischen und algorithmischen Verfahren stoßen je nach Komplexität und Vielfalt der Signale an ihre Grenzen. Auch hier lässt sich die KI anwenden. Im konkreten Beispiel wird eine KI eingesetzt, um die Atmung eines Menschen anhand von Fluss-Volumenkurven zu analysieren. Zusätzlich kommen statische Größen, wie Geschlecht, Alter und Körpergröße für eine Krankheitsdiagnose hinzu. Die KI ist hier in der Lage, alle Informationen zu vereinheitlichen und deren Zusammenhänge, die selbst für einen Arzt nicht immer offensichtlich sind, zu erkennen. Das neuronale Netz ist für diese Aufgabe denkbar einfach. Es handelt sich um ein einfaches vollvernetztes Netz, dessen Eingangsvektor aus allen Stützstellen des Atemverlaufs sowie den statischen Größen besteht. Der Ausgang entspricht in diesem Beispiel den 5 Klassen bzw. Krankheitsbildern (Normal, Asthma, Emphysem, Restriktion und Stenose), die zu diagnostizieren sind und in der *Abbildung 9* dargestellt sind. Die abgebildeten Verläufe zeigen in der blau-gestrichelten Kurve den Normalverlauf einer gesunden Atmung. Rot werden die jeweils charakteristischen Krankheitsverläufe visualisiert. Mit Hilfe der entwickelte KI kann eine Krankheitserkennung realisiert und der Zusammenhang zu den statischen Größen nachgewiesen werden. [3]

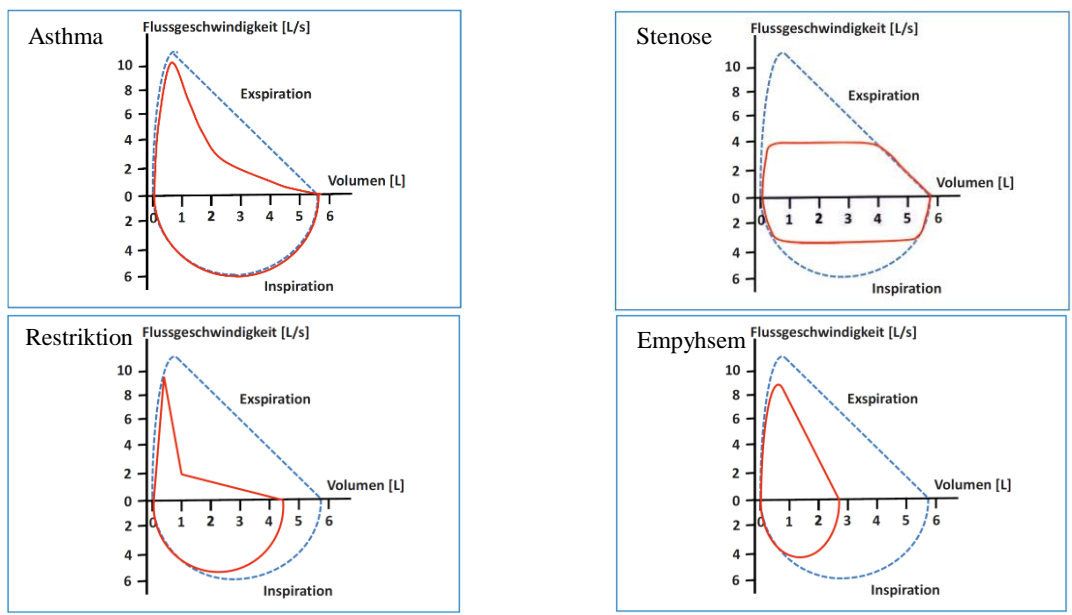

**Abbildung 9:** Diagnose von Lungenerkrankungen anhand von Atmungskurven dargestellt als Fluss-/Volumenkurve [4]

## 2.4 Objekterkennung mit KI

Das letzte Beispiel thematisiert einen sehr bekannten Anwendungsfall. Es handelt sich hierbei um eine Objekterkennung. Eine Objekterkennung setzt sich aus einer Lokalisation und einer Klassifikation zusammen. Eine Klassifikation ist die Eingruppierung eines gefundenen Objektes in eine bestimmte Klasse, abhängig von charakteristischen Merkmalen. Eine Lokalisation ist die Bestimmung des Ortes des gefundenen Objektes. Die Objekterkennung in dieser Anwendung soll zur Realisierung eines Navigationstools für Menschen mit eingeschränktem Sehvermögen dienen. Hierbei werden einige für den Fußgänger relevante Verkehrszeichen, Straßenmarkierungen und Ampelanforderungsknöpfe trainiert, um einen Nutzer sicher durch den Straßenverkehr zu leiten. Für alle Objekte zusammen, werden 18.000 Bilddaten aufgenommen und händisch annotiert. Die Menge an Daten ist allerdings nicht gleichbedeutend mit einer erfolgreichen Realisierung der Anwendung. So zeigt sich im Verlauf des Projektes, dass manche Objekte nicht immer eindeutig erkannt werden können. Dementsprechend werden Systemanalysen durchgeführt. Die Hauptproblematik kann im Bereich der Trainingsdaten festgestellt werden. Alle Objektdatensätze müssen auf ihre Varianz überprüft werden. Durch die Anzahl der Bilder und der unterschiedlichen Merkmale, die eine KI finden kann, ist es manuell äußerst schwierig diese Varianz zu überprüfen. Aus diesem Grund werden verschiedene Analysetools entwickelt, um zum einen die Varianten zu visualisieren, und zum anderen die Analyse zu automatisieren. Mit Hilfe dieser Software können nen klare Erkennungseinbrüche bei spezifischen Darstellungen entdeckt werden. Somit besitzt der Datensatz eindeutige Lücken, die nicht allein durch die Abstraktionsfähigkeit der KI geschlossen werden können. Eine Kompensation ist sowohl durch eine aufwendige weitere Datensammlung, als auch durch die so genannten Augmentationen möglich. Hierbei handelt es sich um Algorithmen, die live im Trainingsprozess die Trainingsdaten manipulieren, um die Variation des Datensatzes zu erhöhen. So wird beispielsweise das Bild um einen gewissen Winkel gedreht, herangezoomt, verschoben, überdeckt oder mit Störungen überlagert. Das neuronale Netz erfährt so eine größere Vielzahl an Darstellungen, wodurch die Trainingsergebnisse verbessert werden können und die Erkennung robuster wird. Die mittlere Erkennungsleistung in realer Umgebung wird damit auf 62% und in der Laborumgebung auf 93% erhöht. Das folgende Bild (***Abbildung 10***) zeigt, wie die Erkennung in einer realen Umgebung stattfindet.

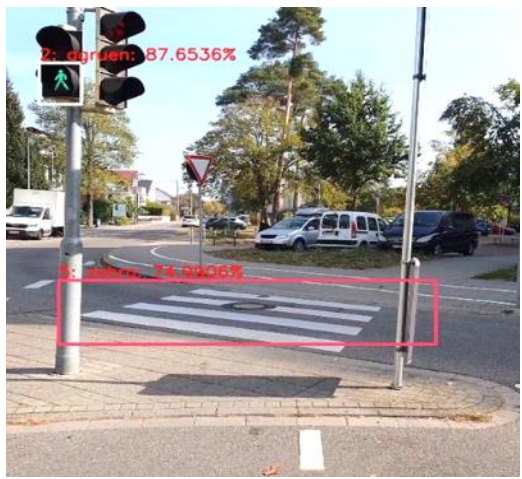

**Abbildung 10:** Erkennung in realer Umgebung

## 3 Fazit

Anhand konkreter Beispiele wurden Systeme und Anwendungen beschrieben, die mit Hilfe von KI Mehrwerte erzeugen können. Es wurde gezeigt, dass im Bereich der Entstörung von Daten die KI Schwächen von algorithmischen Verfahren, ohne wesentliche unerwünschte Nebenwirkungen, ausgleichen kann. Die erzeugende Fähigkeit von generativen KIs bietet interessante neue Möglichkeiten, Daten in jeglichen Formen zu generieren, beispielsweise einem Datenmangel bei KI-Anwendungen entgegenzuwirken. Ebenso wurde verdeutlicht, dass viele verschiedene Signale oder Daten durch eine KI verarbeitet und korreliert werden können, damit bestimmte Reaktionen oder Zustände klassifiziert werden können. Zuletzt wurde an einem Beispiel beschrieben, welche Möglichkeiten, aber auch Schwierigkeiten mit der Implementierung einer Objekterkennung verbunden sind und welche Bedeutung hierbei die verwendeten Daten und deren Variationen haben. Wichtig für alle Formen der KI ist die Qualität der Trainingsdaten: Wenn mit schlechten Daten trainiert wird, wird das Ergebnis ebenso schlecht ausfallen. Alle Beispiele zeigen spezifische KI-Anwendungen aus dem Bereich der Mechatronik. Der Einsatz von KI bietet einen potentiellen Mehrwert in vielen Bereichen der Industrie und kann hierbei die Grenzen des bisher Möglichen erweitern. Jedoch muss beim Einsatz einer KI die Entstehung möglicher Risiken, wie der Nachvollziehbarkeit, der Vorhersehbarkeit und der daraus resultierenden Verantwortung beachtet werden.

## Literaturverzeichnis

## KI4Industry – was ist das?

In den letzten beiden Jahren, haben wir in unserer Funktion als Technologietransferdienstleister für die Industrie immer wieder die Beobachtung gemacht, dass viele mittelständische Unternehmen wahrnehmen, dass ein großer Veränderungsprozess begonnen hat: durch immer leistungsfähigere und günstigere Computertechnik sind heute viele Dinge möglich, die so vor einigen Jahren noch nicht umsetzbar waren. Diese „digitale Revolution" wurde zunächst unter dem Namen Industrie 4.0 diskutiert. In der jüngeren Vergangenheit, in der sich der Fokus auf die Analyse von Daten und die Ableitung von sinnvollen Handlungen verschoben hat, wird eher von KI gesprochen.

Bei beiden Themen blieben für den Mittelstand aber immer einige Fragen offen: Was bringt diese Entwicklung für meinen konkreten Arbeitsalltag? Was muss und kann ich tun, um mein Unternehmen für die Zukunft fit zu machen? Hier setzt KI4Industry an: wir möchten eine Plattform schaffen, die es gerade mittelständischen Unternehmen ermöglicht, Antworten auf diese Fragen für ihr Unternehmen zu finden. Startpunkt ist hierbei die erste Tagung KI4Industry, deren Vorträge auf den kommenden Seiten in Form von Artikeln nochmal nachzulesen sind. In der Zukunft werden weitere Veranstaltungen folgen und auch die ersten bilateralen Projekte zwischen Industrie und Hochschule, die aus der Plattform gewachsen sind, sind bereits in den Startlöchern.

Prof. Dr.-Ing Martin Kipfmüller,
Sprecher Institute of Materials and Processes (IMP)

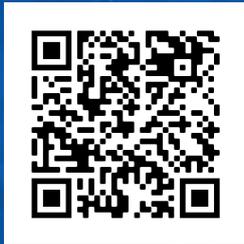

Zusammenfassung in YouTube:

Eine kurze Zusammenfassung aus dem Online-Kongress www.KI4industry.de. Der Online-Kongress zum Einstieg in Künstliche Intelligenz für kleine und mittlere Unternehmen. 12./13. November 2020

Scannen Sie den QR- Code oder klicken Sie auf diesen Link:
https://youtube.com/playlist?list=PLPAfNH8UB9_n9H6IQfaZgIQkRX2BsrldT

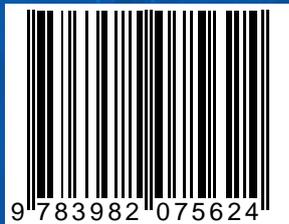